\documentclass[journal]{IEEEtran}
\usepackage{amsmath,amsfonts}
\usepackage{algorithmic}
\usepackage{algorithm}
\usepackage{array}
\usepackage{textcomp}
\usepackage{stfloats}
\usepackage{url}
\usepackage{verbatim}
\usepackage{graphicx}
\usepackage{cite}

\usepackage[utf8]{inputenc}
\usepackage[T1]{fontenc}
\usepackage{babel}
\usepackage[babel=true]{csquotes}
\usepackage{amsthm} 
\usepackage{booktabs}
\usepackage{multirow} 
\usepackage{amssymb} 
\usepackage{float}
\usepackage{hyperref} 
\usepackage[english]{cleveref} 
\usepackage{soul}

\usepackage{color}
\usepackage[dvipsnames,table]{xcolor}
\usepackage{pifont}
\usepackage{svg}
\usepackage[inline]{enumitem}
\usepackage{longtable}
\usepackage{tabularx}
\usepackage{makecell}
\usepackage{lscape}
\usepackage{threeparttable}
\usepackage{xurl}
\usepackage{wasysym}

\usepackage{silence}
\renewcommand{\tablename}{Table}
\usepackage{caption}
\usepackage[caption=false,font=normalsize,labelfont=normalsize,textfont=normalsize]{subfig}

\newcommand\sbullet[1][.5]{\mathbin{\vcenter{\hbox{\scalebox{#1}{$\bullet$}}}}}

\newcolumntype{P}[1]{>{\centering\arraybackslash}p{#1}}
\newcolumntype{C}[1]{>{\centering\arraybackslash}m{#1}}

\setstcolor{blue}

\newtoggle{finalPaper}
\toggletrue{finalPaper} 

\iftoggle{finalPaper} {
    \newcommand{\addtxt}[1]{#1}

    \newcommand{\rmvtxt}[1]{}
} {
    \newcommand{\addtxt}[1]{\textcolor{blue}{#1}}

    \newcommand{\rmvtxt}[1]{\st{#1}}
}

\definecolor{ao}{rgb}{0.0, 0.5, 0.0}
\definecolor{amber}{rgb}{1.0, 0.49, 0.0}
\newcommand{\yestick}{{\color{ao}\ding{51}}}
\newcommand{\notick}{{\color{red}\ding{55}}}
\newcommand{\partialtick}{{\textbf{\color{amber}$\mathord{?}$}}}

\definecolor{gray45}{gray}{.45}
\definecolor{gray75}{gray}{.75}
\definecolor{orange-fig}{HTML}{C55A11}

\begin{document}
\bstctlcite{IEEEexample:BSTcontrol}

\title{Decentralized Federated Learning: Fundamentals, State of the Art, Frameworks, Trends, and Challenges}

\author{Enrique Tomás Martínez Beltrán, Mario Quiles Pérez, Pedro Miguel Sánchez Sánchez, Sergio López Bernal, Gérôme Bovet, Manuel Gil Pérez, Gregorio Martínez Pérez, and Alberto Huertas Celdrán,~\IEEEmembership{Member, IEEE}
\thanks{This work has been partially supported by \textit{(a)} 21629/FPI/21, Fundación Séneca, Región de Murcia (Spain), \textit{(b)} TED2021-129300B-I00, by MCIN/AEI/10.13039/501100011033, NextGenerationEU/PRTR, UE, \textit{(c)} PID2021-122466OB-I00, by MCIN/AEI/10.13039/501100011033/FEDER, UE, \textit{(d)} the Swiss Federal Office for Defense Procurement (armasuisse) with the DEFENDIS and CyberForce projects, and \textit{(e)} the University of Zürich UZH. (Corresponding author: Enrique Tomás Martínez Beltrán.)}%
\thanks{Enrique Tomás Martínez Beltrán, Mario Quiles Pérez, Pedro Miguel Sánchez Sánchez, Sergio López Bernal, Manuel Gil Pérez, and Gregorio Martínez Pérez are with the Department of Information and Communications Engineering, University of Murcia, 30100 Murcia, Spain (e-mail:enriquetomas@um.es; mqp@um.es; pedromiguel.sanchez@um.es; slopez@um.es; mgilperez@um.es; gregorio@um.es).}
\thanks{Alberto Huertas Celdrán is with the Communication Systems Group, Department of Informatics (IFI), University of Zurich, 8050 Zürich, Switzerland (e-mail: huertas@ifi.uzh.ch).}
\thanks{Gérôme Bovet is with the Cyber-Defence Campus, Armasuisse Science and Technology, 3602 Thun, Switzerland (e-mail: gerome.bovet@armasuisse.ch).}}

\markboth{IEEE COMMUNICATIONS SURVEYS AND TUTORIALS}%
{Shell \MakeLowercase{\textit{et al.}}: A Sample Article Using IEEEtran.cls for IEEE Journals}


\maketitle

\begin{abstract}
In recent years, Federated Learning (FL) has gained relevance in training collaborative models without sharing sensitive data. Since its birth, Centralized FL (CFL) has been the most common approach in the literature, where a central entity creates a global model. However, a centralized approach leads to increased latency due to bottlenecks, heightened vulnerability to system failures, and trustworthiness concerns affecting the entity responsible for the global model creation. Decentralized Federated Learning (DFL) emerged to address these concerns by promoting decentralized model aggregation and minimizing reliance on centralized architectures. However, despite the work done in DFL, the literature has not (i) studied the main aspects differentiating DFL and CFL; (ii) analyzed DFL frameworks to create and evaluate new solutions; and (iii) reviewed application scenarios using DFL. Thus, this article identifies and analyzes the main fundamentals of DFL in terms of federation architectures, topologies, communication mechanisms, security approaches, and key performance indicators. Additionally, the paper at hand explores existing mechanisms to optimize critical DFL fundamentals. Then, the most relevant features of the current DFL frameworks are reviewed and compared. After that, it analyzes the most used DFL application scenarios, identifying solutions based on the fundamentals and frameworks previously defined. Finally, the evolution of existing DFL solutions is studied to provide a list of trends, lessons learned, and open challenges.
\end{abstract}


\section{Introduction}
\label{sec:introduction}

\begin{figure}[!ht]
\centering
\includegraphics[width=\columnwidth]{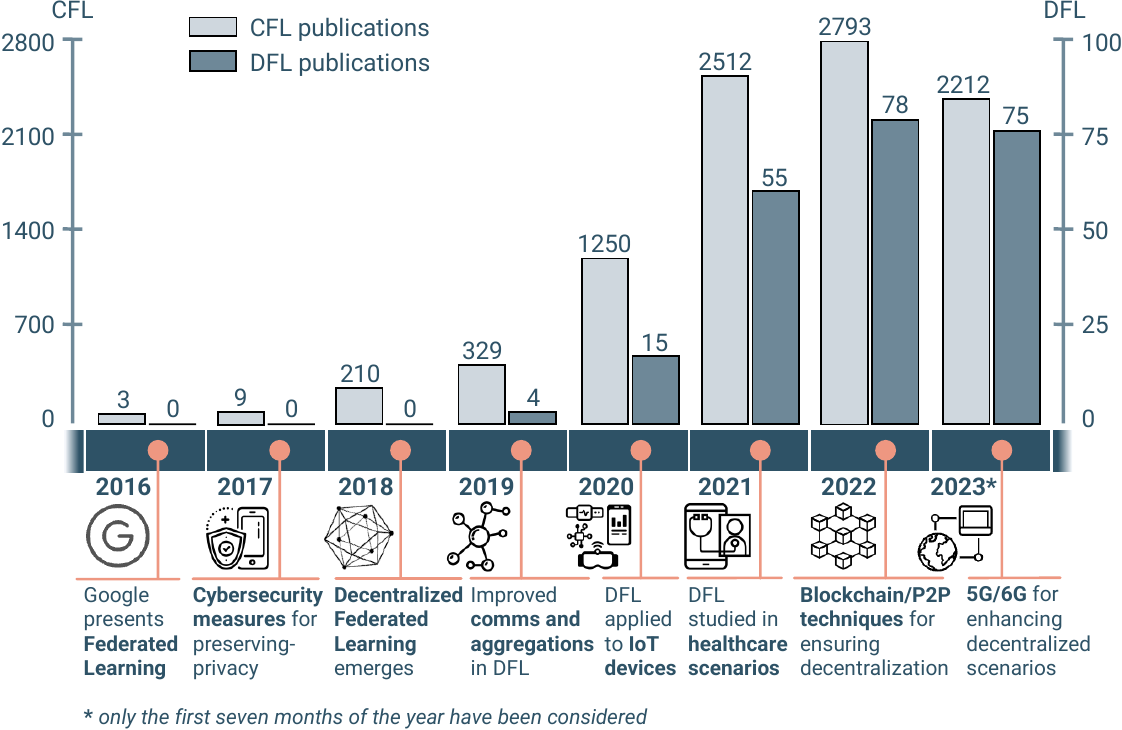}
\caption{\addtxt{DFL Timeline (source: Web of Science).}}
\label{fig:timeline-DFL}
\end{figure}

\IEEEPARstart{A}{rtificial} Intelligence (AI) and, in particular, Machine Learning (ML), will be one of the techniques that benefit from the vast amount of data expected in the coming years. However, data originating from a massive number of Internet of Things (IoT) devices are commonly stored in a distributed manner for a wide range of scenarios such as smart grids \cite{Liu:voltage_control:2022}, remote health monitoring \cite{Lian:efficient_privacy_healthcare:2022}, or Internet of Vehicles (IoV) \cite{Barbieri:vehicles:2022}. In such cases, data collection in central entities, as traditionally done in ML, is often infeasible or impractical due to limited communication resources, data privacy concerns, or country regulations. In 2016, Federated Learning (FL) \cite{McMahan:communication_efficient:2016} solved this issue by allowing entities (also known as participants, clients, or nodes of a federation) to train collaborative models without sharing training data. FL can be classified into two categories according to how federated models are created: centralized, known as Centralized Federated Learning (CFL), and decentralized, known as Decentralized Federated Learning (DFL). Nowadays, CFL is the predominant FL approach. It considers a central orchestration server to create and distribute a global model to the rest of the participants or clients. More in detail, clients train their models with local data, then send the local model parameters to a central server, where a global model is created by aggregating and combining the parameters of the individual model. \figurename~\ref{fig:timeline-DFL} shows the timeline since Google introduced FL and how new features and approaches for decentralized scenarios have emerged. As can be seen, the growth for CFL and DFL is remarkable.

DFL, also known as Serverless or Distributed FL, emerged in 2018 to distribute the aggregation of model parameters between the neighboring participants \cite{He:fully_dfl_algorithm:2018}. The operation of DFL focuses on transmitting fast updates computed locally by each node (e.g., model parameters or gradients) and metadata (e.g., activations functions in neural networks) to the rest of the federation nodes. Therefore, compared to CFL, DFL improves the limitations of having a single point of failure, trust dependencies, and bottlenecks at the server node \cite{Hard:google_centralized_to_dfl:2021}. Firstly, DFL improves fault tolerance because nodes constantly update their knowledge about available nodes or those that have ceased communicating \cite{Wang:edge_communication_optimization:2021}. This enhances the robustness of the network and mitigates the risk of a single point of failure, as it happens in CFL. Secondly, DFL improves trust issues by distributing trust between the federation nodes, which determines the trust of the entire federation, drastically reducing the single point of attack. Thirdly, DFL reduces the network bottleneck issue by allowing more evenly distributed communication and workload among the nodes, thus reducing the chances of congestion or delays in the overall network performance. Furthermore, DFL provides self-scaling federations, as new nodes can join the network and start communicating with other nodes, considering factors such as device throughput or computational capacity. Although the number of connections may decrease due to the limited number of nodes each node can share data with, the network remains tightly connected \cite{McMahan:communication_efficient:2016}. Finally, DFL enables more efficient use of resources by distributing the computing power required to aggregate the model parameters among all participating nodes, rather than relying on a single parameter server under centralized control \cite{Savazzi:cooperative_fl_robots_drones:2021}.

Despite the benefits provided by DFL, it also introduces new challenges in areas such as communication overhead, training optimization, and trustworthy AI. Depending on how the model aggregation is distributed across the network, some DFL topologies may face increased communication overhead \cite{Bellet:topology_data_heterogeneity:2021}. In these cases, careful design and optimization of the communication protocols, client selection strategies, and trust mechanisms can help address these limitations. Trust plays a critical role in these tasks, as it influences the decision of which clients to share and aggregate model parameters with. Additionally, the significant amount of model parameters exchanged from the network edge to the data centers carries the risk of saturating the backbone. As a result, using DFL necessitates reevaluating the infrastructure from a new perspective, as it reduces the need for centralized infrastructure while aiming to improve network performance and trustworthiness. In this context, trust becomes essential in addressing the emerging challenges of DFL, such as client selection and parameter sharing. These challenges warrant further exploration and research, as the paper at hand does.

\begin{figure}[!ht]
\centering
\includegraphics[width=0.85\columnwidth]{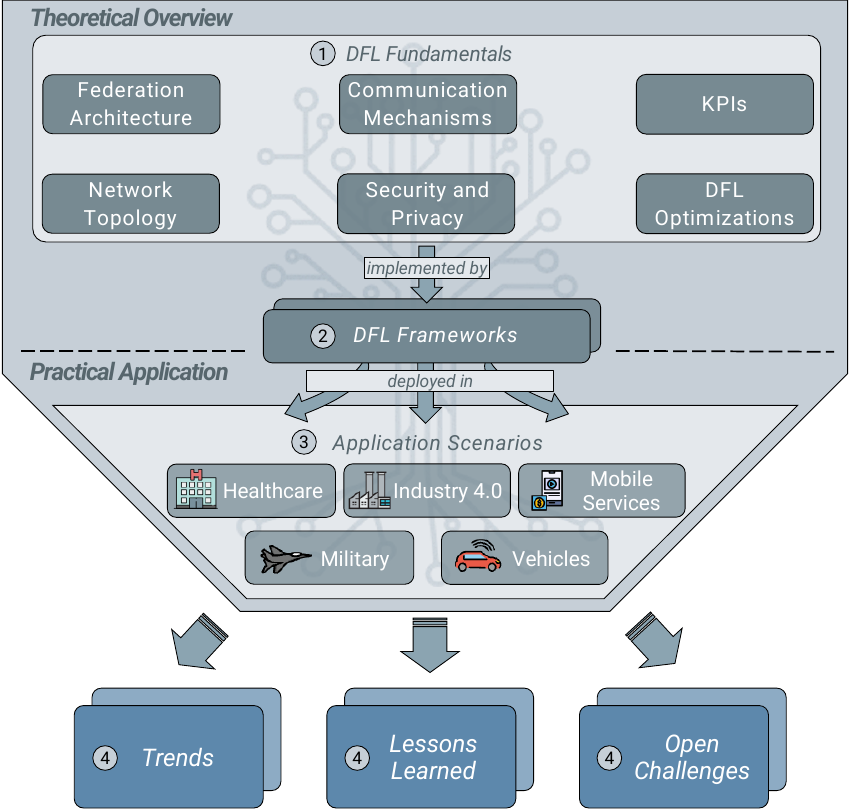}
\caption{Methodological steps and storyline of this work.}
\label{fig:storyline}
\end{figure}

To explore the research field of DFL in an organized fashion, the paper follows the methodological steps indicated in \figurename~\ref{fig:storyline}. First, it is essential to identify and study the key aspects that differentiate DFL from CFL (step 1 in \figurename~\ref{fig:storyline}). For this purpose, the literature has highlighted some elements such as (i) federation architecture, in charge of modeling the decentralized scenario based on participants, roles, decentralization of the nodes, and distribution of data features; (ii) network topology, defining the associations between the nodes of the federation \cite{Georgatos:efficient_adaptive_local_links:2022}; (iii) communication mechanisms, orchestrating the exchange of model parameters within the federation \cite{Bonawitz:google_sync:2019}; and (iv) security and privacy, studying possible cyberattacks and the countermeasures to preserve data privacy and models robustness \cite{Qu:privacy_framework_iot:2022}. In addition to the above fundamentals, identifying Key Performance Indicators (KPIs) is important for assessing DFL performance \cite{Zhao:network_system:2022}. Three perspectives can be explored in this context: nodes, communications, and models. The first focuses on evaluating the heterogeneity and dynamism of nodes in DFL, the second on the efficiency of inter-node communications for data exchange, and the third on the performance of ML/DL models for solving collaborative tasks. Finally, mechanisms to optimize KPIs are also essential to ensure the efficiency and multi-scenario adaptability of the approach \cite{Wang:dfl_select:2022}. In this context, this article expands upon the existing literature by comprehensively identifying and describing the DFL fundamentals, KPIs, and optimization mechanisms.

The aforementioned fundamentals are crucial for developing and comparing frameworks capable of training DFL-based models (step 2 in \figurename~\ref{fig:storyline}). These fundamentals are particularly important because they guide frameworks on managing model parameter exchanges efficiently, establishing federations with network participants, implementing effective aggregation mechanisms for local models, and fostering more reliable and trustworthy models. By considering these fundamentals, DFL frameworks can ensure better performance, security, scalability, and trustworthiness across a wide range of use cases \cite{epfl:disco_framework:2022}. Therefore, the article at hand also aims to support researchers working in decentralized systems using federated approaches with a review of the most relevant features of existing frameworks.

Then, application scenarios of DFL should be analyzed according to the fundamentals and frameworks they implement (step 3 in \figurename~\ref{fig:storyline}). At this point, it is important to identify strengths and weaknesses, and enables more informed decisions for selecting and deploying suitable approaches in real-world scenarios. Nowadays, the most used scenarios are: (i) healthcare, favoring the decentralization of clinical records and collaborative diagnosis \cite{Nguyen:healthcare_nature:2022}; (ii) Industry 4.0, improving the efficiency of automated industrial systems \cite{Kang:bl_metaverse_industrial:2022}; (iii) mobile services, decreasing response times and increasing the bandwidth of constraints devices \cite{Wang:dfl_edge:2022}; (iv) military, creating robust networks between Unmanned Aerial Vehicles (UAVs) and protecting them with cyber defense mechanisms \cite{Wang:uav_military:2021}; and (v) vehicles, ensuring high mobility and local storage management \cite{Yu:p2p_vehicles:2020}. Subsequently, each scenario is explored by studying the proposed solutions according to the previous fundamentals. Finally, all the aspects highlighted above (fundamentals, frameworks, and application scenarios) are analyzed with a view into the past, present, and future to extract trends, lessons learned, and challenges in the field of DFL. This is the last step of the followed methodology (step 4 in \figurename~\ref{fig:storyline}), which also shares the same structure and storyline as the document at hand. 
\section{Motivation and Contributions}
\label{sec:motivation}

DFL is an encouraging field of research that has repeatedly been reported as challenging in several review articles in recent years. While there are numerous surveys on FL, most focus primarily on the CFL-based approach and provide limited coverage of DFL. Therefore, to the best of our knowledge, this is the first survey providing a comprehensive literature review of DFL. The increase in the number of papers using DFL in different scenarios: healthcare \cite{Lian:efficient_privacy_healthcare:2022, Wang:ring_topology_healthcare:2022}, Industry 4.0 \cite{Kang:bl_metaverse_industrial:2022}, mobile services \cite{Wang:dfl_edge:2022}, military \cite{Wang:uav_military:2021}, and vehicles \cite{Chen:bdfl_autonomous_vehicle:2021} motivate the need to identify the fundamental elements of DFL and review the work done to date. 

\begin{table*}[htb!]
\caption{Comparison of surveys analyzing CFL and DFL.} \label{tab:surveys}
\resizebox{\textwidth}{!}{
\centering
\begin{threeparttable}
\begin{tabular}{ccccccccp{8cm}} 
\hline
Ref. & \makecell[t]{Year} & \makecell[t]{FL\\Approach} & \makecell[t]{Device\\Types / Area} & \makecell[t]{Fundamentals} & \makecell[t]{Frameworks} & \makecell[t]{Application\\Scenarios} & \makecell[t]{Key Performance\\Indicators} & \makecell[t]{Focus and Solution Categorization} \\ 
\hline \hline

\cite{Lim:fl_edge_survey:2020} & 
2020 & 
CFL & 
\makecell[t]{Mobile\\networks} & 
\yestick & 
\yestick &
\notick & 
\yestick & 
$\sbullet[0.75]$ A survey on the integration of FL and edge computing.\newline
$\sbullet[0.75]$ The applications of FL in IoT networks and services have not been explored and discussed.\\

\hline

\cite{Nguyen:fl_iot_survey:2021} & 
2021 &
CFL &
\makecell[t]{IoT\\devices} &
\notick & 
\notick &
\yestick & 
\notick & 
$\sbullet[0.75]$ A survey on FL in an IoT scenario, reviewing data distribution, ML models, privacy mechanism, and communication architecture.\newline
$\sbullet[0.75]$ The study has not discussed other approaches or frameworks capable of managing the scenario.\\ 

\hline

\cite{Khan:fl_iot_survey:2021} & 
2021 &
CFL &
\makecell[t]{IoT\\devices} &
\yestick & 
\notick &
\yestick & 
\yestick & 
$\sbullet[0.75]$ A systematic review of FL in IoT scenarios, detailing the application areas and limitations.\newline
$\sbullet[0.75]$ The study only considers CFL architectures without discussing the disadvantages compared to other approaches.\\ 

\hline

\cite{Mothukuri:survey_topology_architecture_privacy_decentralized:2021} & 
2021 &
CFL &
\makecell[t]{Security\\Privacy} & 
\notick & 
\yestick & 
\yestick &
\notick & 
$\sbullet[0.75]$  A survey of FL security and privacy, defining secure protocols.\newline
$\sbullet[0.75]$ The analysis of secure measures using different FL approaches is missing.\\

\hline

\cite{Boobalan:survey:2022} & 
2022 & 
CFL &
\makecell[t]{FL\\baselines} & 
\yestick & 
\yestick &
\partialtick & 
\partialtick & 
$\sbullet[0.75]$ A survey on the FL baselines with a basic introduction to definitions and architectures.\newline
$\sbullet[0.75]$ The application scenarios of FL have not been discussed.\\ 

\hline

\cite{Joshi:healthcare_survey:2022} & 
2022 & 
CFL &
\makecell[t]{Healthcare} & 
\partialtick & 
\notick & 
\yestick &
\notick & 
$\sbullet[0.75]$ An overview of the use of FL in Healthcare scenarios.\newline
$\sbullet[0.75]$ Limited overview of techniques for deploying realistic scenarios.\\ 

\hline

\cite{Witt:survey_frameworks:2022} & 
2022 & 
CFL &
\makecell[t]{Framework\\review} & 
\yestick & 
\yestick & 
\notick &
\notick & 
$\sbullet[0.75]$ An overview of frameworks for deploying DFL-based architectures.\newline
$\sbullet[0.75]$ Lack of identification of application scenarios.\\ 

\hline

\cite{Qu:bl_survey:2022} & 
2022 & 
DFL &
\makecell[t]{DLT} &
\notick & 
\notick & 
\notick &
\partialtick & 
$\sbullet[0.75]$ A description of the challenges and applications of Blockchain.\newline
$\sbullet[0.75]$ The paper only focuses on the Blockchain technology to ensure node decentralization.\\ 
\hline

\cite{Billah:bl_frameworks:2022} & 
2022 & 
DFL &
\makecell[t]{IoV} & 
\notick & 
\yestick & 
\yestick &
\partialtick & 
$\sbullet[0.75]$ A brief description of FL in fog radio access networks.\newline
$\sbullet[0.75]$ The applications of FL in IoT networks have not been presented.\\

\hline

\cite{Gupta:distributed_survey:2022} & 
2022 & 
DFL &
\makecell[t]{Wireless\\communications} & 
\notick &  
\notick &
\yestick &
\yestick & 
$\sbullet[0.75]$ A review of the techniques for adapting FL to distributed environments.\newline
$\sbullet[0.75]$ The paper only reviews techniques without giving specific application scenarios or frameworks.\\

\hline

\cite{Saraswat:uavs_survey:2022} & 
2022 & 
DFL &
\makecell[t]{UAV\\devices} & 
\notick &  
\notick &
\yestick &
\yestick & 
$\sbullet[0.75]$ A brief survey on the application of FL in UAV networks.\newline
$\sbullet[0.75]$ Other domains, such as smart city or IoT devices, are not considered.\\

\hline

\cite{Jiajun:survey_dfl_topologies:2023} & 
2023 & 
DFL &
\makecell[t]{DFL\\baselines} & 
\partialtick &  
\notick &
\partialtick &
\yestick & 
$\sbullet[0.75]$ A study of optimized DFL models and algorithms focusing on network topologies.\newline
$\sbullet[0.75]$ Limited review regarding federation architectures, device heterogeneity, and frameworks.\\

\hline

\cite{Huiming:survey_advances_dfl:2023} & 
2023 & 
DFL &
\makecell[t]{DFL\\baselines} & 
\partialtick & 
\notick & 
\yestick &
\yestick & 
$\sbullet[0.75]$ A comparison of CFL and DFL federation architectures regarding topologies, privacy, and security.\newline
$\sbullet[0.75]$ The study does not have enough information about DFL, including algorithms, optimizations, or metrics for evaluating its performance.\\
\hline

\cite{Witt:frameworks_dfl_bl:2023} & 
2023 & 
DFL &
\makecell[t]{Framework\\review} & 
\partialtick &  
\yestick &
\notick &
\notick & 
$\sbullet[0.75]$ A systematic review of DFL frameworks, highlighting differences, limitations, and future research directions.\newline
$\sbullet[0.75]$ The study focuses only on Blockchain-related frameworks and participant reward techniques.\\
\hline

This work & 
2023 & 
DFL &
\makecell[t]{Healthcare\\Industry 4.0\\Mobile services\\Military\\ Vehicles} & 
\yestick & 
\yestick & 
\yestick &
\yestick & 
$\sbullet[0.75]$ A survey on the baselines and fundamentals of DFL, generating a novel taxonomy in the literature.\newline
$\sbullet[0.75]$ The paper explores the application of DFL solutions and frameworks in various areas of interest.\\
\hline \hline
\end{tabular}
\begin{tablenotes}
\item \yestick\space fully addressed,\space\partialtick\space partially addressed,\space\notick\space not addressed by the work
\end{tablenotes}
\end{threeparttable}}
\end{table*}

\tablename~\ref{tab:surveys} compares the most relevant surveys analyzing both CFL and DFL approaches. Regarding CFL, several works have attracted immense interest from academia and industry \cite{Li:fl_challenges_methods_future:2020}. In this sense, in 2020, Lim \textit{et al.} \cite{Lim:fl_edge_survey:2020} reviewed the use of CFL in mobile networks using different end-user applications, while in 2021, researchers such as Nguyen \textit{et al.} \cite{Nguyen:fl_iot_survey:2021} and Khan \textit{et al.} \cite{Khan:fl_iot_survey:2021} examined the application of FL in more complex scenarios involving different heterogeneous IoT devices. Regarding security and privacy, Mothukuri \textit{et al.} \cite{Mothukuri:survey_topology_architecture_privacy_decentralized:2021} defined secure protocols between server and participants, while Boobalan \textit{et al.} \cite{Boobalan:survey:2022} provided solid CFL fundamentals. Other articles, such as \cite{Joshi:healthcare_survey:2022}, reviewed the applicability in specific medical scenarios, discussing sensitive data privacy mechanisms. Finally, Witt \textit{et al.} \cite{Witt:survey_frameworks:2022} detailed the frameworks most commonly used.

In contrast, most of the works covering DFL approaches focus on analyzing the use of Distributed Ledger Technology (DLT) technologies (e.g., Blockchain) \cite{Qu:bl_survey:2022}. Similarly, other authors explored the applicability of DFL in particular scenarios such as IoV \cite{Billah:bl_frameworks:2022}, wireless communications \cite{Gupta:distributed_survey:2022}, or UAV devices \cite{Saraswat:uavs_survey:2022}. Moreover, the article by Wu \textit{et al.} \cite{Jiajun:survey_dfl_topologies:2023} provided a comprehensive survey of optimized DFL models and algorithms focusing on network topologies. The authors introduced various topologies with different participants and potential optimizations for them. Additionally, Chen \textit{et al.} \cite{Huiming:survey_advances_dfl:2023} analyzed the advances in FL, including aspects such as communications, privacy, and security requirements in DFL. Also, they provided solutions for collaborative training using more generalized algorithms. Further research has also been conducted on the frameworks that enable DFL. Witt \textit{et al.} \cite{Witt:frameworks_dfl_bl:2023} proposed a set of solutions that address privacy and security issues in DFL. The authors also provided reward systems based on smart contracts to incentivize honest customer participation in the federated process. These latter works address the performance of DFL in selected application scenarios and detail critical elements of superiority over CFL approaches.

Despite the contributions of previous works, as illustrated in \tablename~\ref{tab:surveys}, none addresses a literature review of DFL. In addition, no previous research studies the fundamentals of the emerging DFL approach. Similarly, recent work does not review DFL solutions from a fundamentals perspective as elements of analysis and comparison for different application scenarios. These solutions are often deployed on frameworks capable of managing the proposed federated approach \cite{Dai:dispfl_framework:2022, Billah:bl_frameworks:2022}. However, no study highlights which frameworks are feasible for this new decentralized approach nor collects the latest developments. Within this context, and adhering to the storyline outlined in Section~\ref{sec:introduction}, it becomes essential to address the following research questions:

\begin{itemize}
    \item \textit{Q1. What are the fundamental aspects of DFL?} Depending on the federation architecture of the nodes, network topology, communication mechanisms between nodes, or security techniques, variations in the DFL approach may occur. Similarly, approach-based performance metrics and optimization techniques are needed. However, solutions in the literature ignore details about these elements and how they are combined.
    \item \textit{Q2. What DFL frameworks exist, and what fundamentals do they provide?} There is no study reviewing existing frameworks able to build DFL solutions, analyzing their characteristics, and defining which application scenarios can be utilized.
    \item \textit{Q3. Which are the main characteristics of the most relevant scenarios of DFL?} It is necessary to analyze and compare the previous DFL fundamentals and frameworks to solve the problems motivated by each application scenario. In addition, it is also necessary to detect the characteristics and limitations posed by existing solutions. The literature has not studied these approaches from a broad perspective to have a complete view.
    \item \textit{Q4. What trends, lessons learned, and challenges have emerged in DFL?} To establish the guidelines for future research, it is critical to describe how DFL has evolved over the last years and what trends and open challenges are in the field.
\end{itemize}

\begin{figure*}[!htb]
\centering
\includegraphics[width=0.85\textwidth]{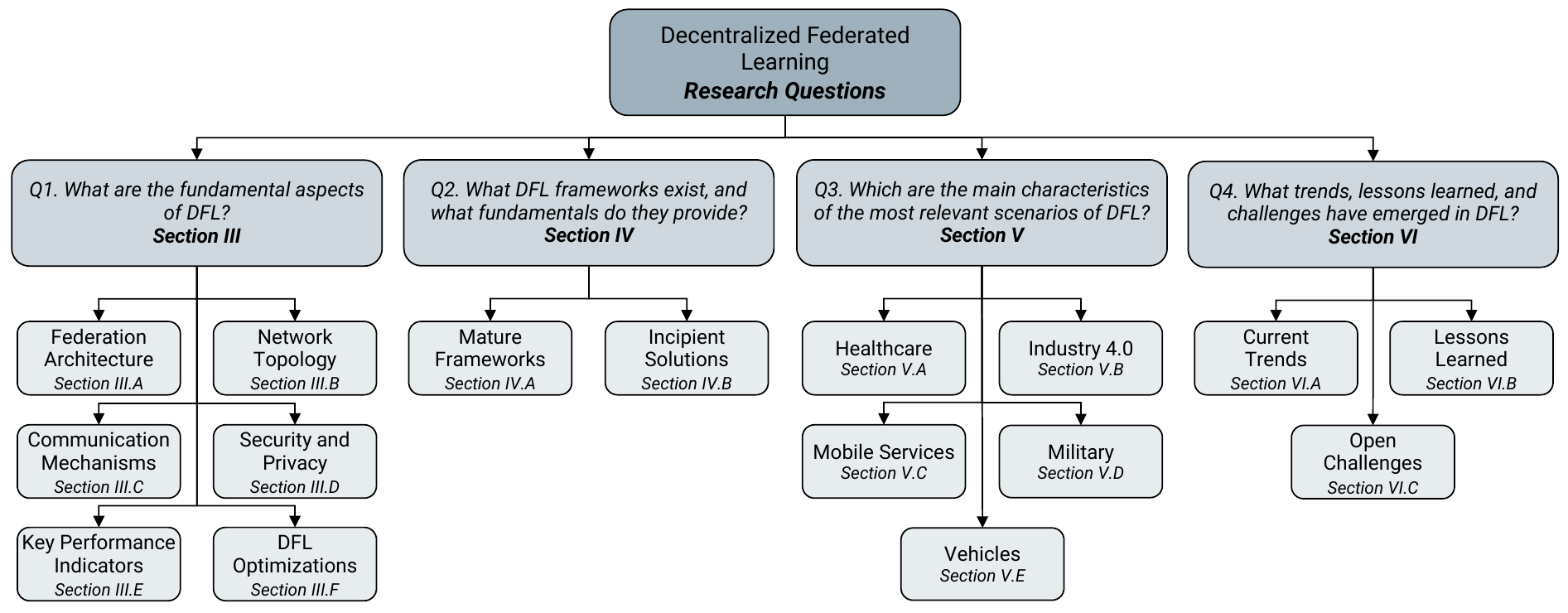}
\caption{Research questions and sections answering them.}
\label{fig:questions}
\end{figure*}

To answer the previous research questions and provide readers with an up-to-date vision of DFL, the main contributions of this manuscript are:

\begin{itemize}
    \item An analysis of the fundamentals that compose DFL: (i) federation architecture, (ii) network topology, (iii) communication mechanisms, (iv) security and privacy, (v) KPIs, and (vi) techniques to optimize the above KPIs, paying attention to the studies and applications that address each fundamental (answering \textit{Q1} in Section~\ref{sec:fundamentals}).
    \item A description of the main open-source frameworks to create and manage DFL solutions. This description is divided into mature literature solutions that have been redesigned to offer decentralized architectures and incipient solutions that address specific scenarios (answering \textit{Q2} in Section~{\ref{sec:frameworks}}).
    \item A comprehensive review and comparison of the characteristics, advantages, and limitations of the most relevant solutions in the literature on the most used application scenarios: (i) healthcare, (ii) Industry 4.0, (iii) mobile services, (iv) military, and (v) vehicles (answering \textit{Q3} in Section~{\ref{sec:scenarios}}).
    \item A set of current trends, lessons learned, and future challenges drawn from DFL works and frameworks reviewed (answering \textit{Q4} in Section~\ref{sec:challenges}).
\end{itemize}

\figurename~\ref{fig:questions} shows where and how the above questions are addressed in the paper at hand, acting as a table of contents. In particular, Section~\ref{sec:fundamentals} analyzes the fundamentals of DFL in terms of architectures, topologies, communications, security, performance indicators, and techniques for optimizing scenarios. Section~{\ref{sec:frameworks}} examines the main open-source frameworks that support the deployment of DFL with the previously defined fundamentals. After that, Section~{\ref{sec:scenarios}} describes and compares the leading solutions found in the state of the art, analyzing deployment considerations, the robustness of the proposed solution, and the results obtained. Section~\ref{sec:challenges} draws a set of lessons learned, current trends, and future challenges in the research area. Finally, Section~\ref{sec:conclusion} provides an insight into the conclusions extracted from the work.

\section{Fundamentals and Taxonomy}
\label{sec:fundamentals}

\begin{figure*}[!b]
\centering
\includegraphics[width=0.85\textwidth]{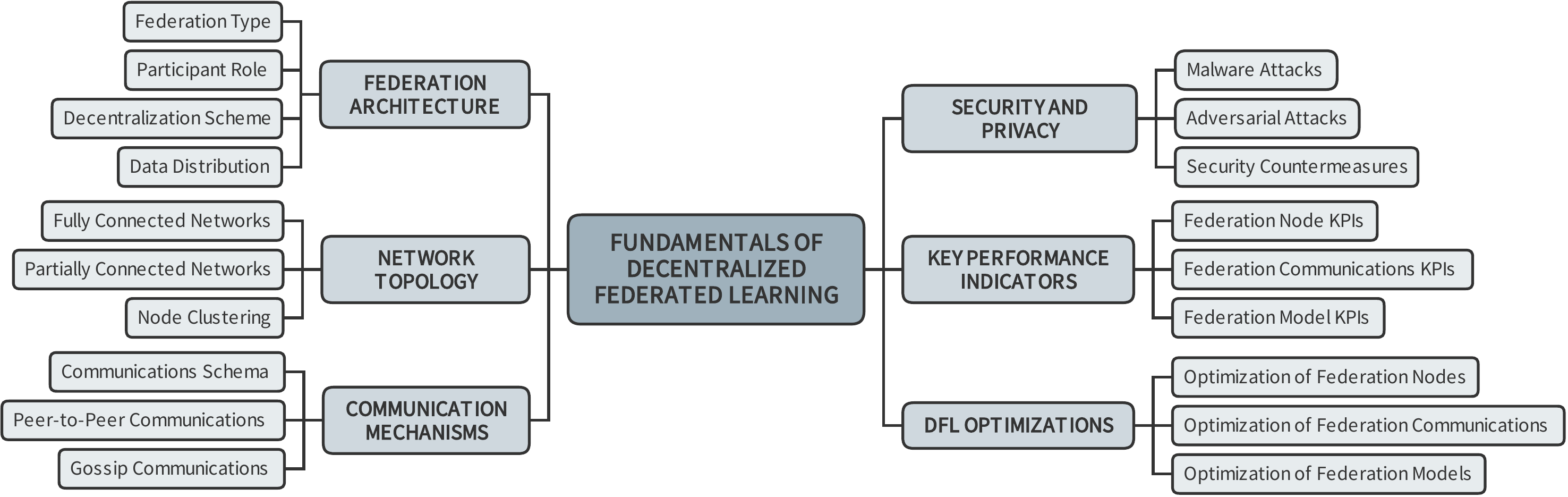}
\caption{Taxonomy with the fundamental aspects of DFL.}
\label{fig:taxonomy}
\end{figure*}

The fundamentals of DFL establish the groundwork for analyzing existing frameworks and solutions in the literature. Following the storyline presented in \figurename~\ref{fig:storyline}, this section responds to ``\textit{Q1. What are the fundamental aspects of DFL?},'' identifying and analyzing the following fundamentals: (i) federation architectures, (ii) network topology, (iii) communication mechanisms, (iv) security and privacy, (v) KPIs, and (vi) techniques to optimize the previous metrics. These fundamentals are detailed in \figurename~\ref{fig:taxonomy}, showing the subcategories that compose them and the resulting taxonomy. Additionally, \tablename~\ref{tab:fundamentals} details the above taxonomy with a detailed definition of the various fundamentals and the literature solutions that address them.

\subsection{Federation Architecture}

DFL solutions require an architecture that allows effective collaboration and communication between network participants. In this sense, this work analyzes the fundamentals based on the federation type in the communication, differentiating organizations or data centers from other devices. Subsequently, DFL architectures can also be categorized based on the roles of the participants. In this classification, each participant is assigned a specific task within the federated learning process. Furthermore, the decentralization schema differentiates between a decentralized, semi-decentralized, and centralized architecture. The semi-decentralized approach can be seen as a hybrid between centralized and decentralized architectures, offering a balance of both. Finally, the definition of DFL architectures is based on the data distribution during federation. In the following sections, each category will be discussed in greater detail.

\begin{table*}[ht!]
\caption{Comparison of solutions addressing different DFL fundamentals.} \label{tab:fundamentals}
\resizebox{\textwidth}{!}{
\centering
\begin{tabular}{p{2cm}p{5.1cm}p{11cm}p{2.1cm}} 
\hline
Taxonomy & Fundamentals & Definition and Solutions Analyzed in the Literature & References \\ 
\hline \hline

\multirow{1}{*}{\makecell[l]{Federation\\Architecture}} & 
\multirow{1}{*}{\makecell[l]{Federation Type}} & 
$\sbullet[0.75]$ Federation from data silos or data centers & \cite{Marfoq:dfl_optimization_comms:2020,Han:dfl_cross_silo:2022, Marfoq:dfl_optimization_comms:2020} \\
& & $\sbullet[0.75]$ Federation from different types of devices, such as mobile phones or IoT devices & \cite{Lian:efficient_privacy_healthcare:2022, Karimireddy:dfl_crossdevice:2021, Feng:dfl_horizontal_bl:2022}
\\ \cline{2-4}

& \multirow{1}{*}{\makecell[l]{Participant Role}} & 
$\sbullet[0.75]$ Identify the functionality of participants during federation & \cite{Wilt:scatterbrained_framework:2021, Kalra:proxyfl_proxy_model_sharing:2021} \\ \cline{2-4}

& \multirow{1}{*}{\makecell[l]{Decentralization Schema}} & 
$\sbullet[0.75]$ Techniques for enabling DFL while maintaining model convergence and data privacy & \cite{Hard:google_centralized_to_dfl:2021, Kanagavelu:dfl_mpc:2022, Xu:udfl_iot:2022} \\
& & $\sbullet[0.75]$ Balance the power and leadership in SDFL while maintaining the benefits of DFL & \cite{Yemini:dfl_semi:2022, Wang:dfl_select:2022} \\
& & $\sbullet[0.75]$ Combine centralized and decentralized elements to optimize FL performance and scalability & \cite{Lin:semi_d2d_aggregation:2021} \\ \cline{2-4}

& \multirow{1}{*}{\makecell[l]{Data Distribution}} & 
$\sbullet[0.75]$ Solutions for HFL, VFL, and TFL partitioning the dataset & \cite{Feng:dfl_horizontal_bl:2022, Sanchez:demeter:2022} \\
& & $\sbullet[0.75]$ Explore ways to sample data to ensure better distribution across nodes & \cite{Li:dfl_knowledge_transfer:2022}
\\ \hline

\multirow{1}{*}{\makecell[l]{Network\\Topology}} & 
\multirow{1}{*}{\makecell[l]{Fully Connected Networks}} & 
$\sbullet[0.75]$ Define the federated networks with participants connected to each other & \cite{Bellet:topology_data_heterogeneity:2021, Xiao:military_fullydl:2021} \\
& & $\sbullet[0.75]$ Measure of robustness, flexibility, fault tolerance, cost, or security & \cite{Georgatos:efficient_adaptive_local_links:2022, Ma:iot_selection_quantized:2021}
\\ \cline{2-4}

& \multirow{1}{*}{\makecell[l]{Partially Connected Networks}} & 
$\sbullet[0.75]$ Define the classification in star, ring, or random structured networks & \cite{Chow:expander_dfl:2016, Hua:topology_random:2021} \\
& & $\sbullet[0.75]$ Measure of robustness, flexibility, fault tolerance, cost or security & \cite{Vogels:toplogy_decision_convergence_models:2022, Cheng:secureboost:2019}
\\ \cline{2-4}

& \multirow{1}{*}{\makecell[l]{Node Clustering}} & 
$\sbullet[0.75]$ Define and classify the federated networks based on participants clusters & \cite{Kalra:proxyfl_proxy_model_sharing:2021, Briggs:hierarchy_cluster_fl:2020} \\
& & $\sbullet[0.75]$ Measure of robustness, flexibility, fault tolerance, cost or security & \cite{Al-Abiad:computation_opt:2023}
\\ \hline

\multirow{1}{*}{\makecell[l]{Communication\\Mechanisms}} & 
\multirow{1}{*}{\makecell[l]{Communications Scheme}} & 
$\sbullet[0.75]$ Data exchange where all participants transmit their models at the same time & \cite{Bonawitz:google_sync:2019, Pinyoanuntapong:trust_sync_async_edgeiot:2022} \\
& & $\sbullet[0.75]$ Update participants models independently at different times & \cite{Zehtabi:sync_async_thresholds:2022, Cao:dfl_async:2021}
\\ \cline{2-4}

& \multirow{1}{*}{\makecell[l]{Peer-to-Peer Communications}} & 
$\sbullet[0.75]$ Methods for designing P2P topologies in DFL & \cite{Chen:p2p:2022, Zhou:p2p_efficient:2022, Wang:p2p_graphs:2022} \\
& & $\sbullet[0.75]$ Develop techniques for disseminating data and models in P2P networks & \cite{Shi:over_the_air:2021, Li:p2p_adaptative_clients_neighbor:2022, Wink:p2p:2021, Qi:dfl_bl:2022}
\\ \cline{2-4}

& \multirow{1}{*}{\makecell[l]{Gossip Communications}} & 
$\sbullet[0.75]$ Methods to design gossip protocols in DFL & \cite{Hegedus:gossip_vs_centralized_fl:2021, Tang:gossipfl_framework:2023} \\
& & $\sbullet[0.75]$ Strategies to select nodes and optimize the efficiency in gossip-based communication & \cite{Jiang:bacombo_bandwidth_gossip:2020, Belal:recommender_gossip:2022, Khelghatdoust:socialnetworks_gossip:2022, koloskova:choco_gossip_communication_compression:2019}
\\ \hline

\multirow{1}{*}{\makecell[l]{Security and\\Privacy}} & 
\multirow{1}{*}{\makecell[l]{Malware Attacks}} & 
$\sbullet[0.75]$ Determine the impacts of malware threats in decentralized scenarios & \cite{Yin:botnet:2017, Rawat:botnet:2021, Chen:worm:2020} \\
& & $\sbullet[0.75]$ Identify the risks associated with fully connected, star-structured, and random networks & \cite{Sanchez:demeter:2022, Cheng:secureboost:2019, Qu:privacy_framework_iot:2022}
\\ \cline{2-4}

& \multirow{1}{*}{\makecell[l]{Adversarial Attacks}} & 
$\sbullet[0.75]$ Identify adversarial attacks affecting decentralized architectures during collaborative federation & \cite{Sanchez:demeter:2022, Alangot:eclipse_attack_defense:2021, Niu:eclipse_attack:2022} \\
& & $\sbullet[0.75]$ Compare the adversarial attacks between different FL approaches and their adaptive capacity & \cite{Verbraeken:bristle_noniid_attacks:2021, Lu:dfl_gradient_inversion_attack_defense:2023}
\\ \cline{2-4}

& \multirow{1}{*}{\makecell[l]{Security Countermeasures}} & 
$\sbullet[0.75]$ Secure techniques to protect the transmission of model parameters between participants & \cite{McMahan:communication_efficient:2016, Chen:wireless_privacy:2022, MartinezBeltran:dfl_mtd:2023, PeralesGomez:temporalfed:2023} \\
& & $\sbullet[0.75]$ Encryption mechanisms while preserving reliability between participants & \cite{Kanagavelu:dfl_mpc:2022, Geyer:client_privacy:2017, MartinezBeltran:dfl_mtd:2023}
\\ \hline

\multirow{1}{*}{\makecell[l]{Key Performance\\Indicators}} & 
\multirow{1}{*}{\makecell[l]{Federation Nodes KPIs}} & 
$\sbullet[0.75]$ Measure the node performance considering the heterogeneity and dynamism of the network & \cite{Verhaelen:kpi:2021, Zhao:network_system:2022} \\
& & $\sbullet[0.75]$ Quantify the resource capabilities and node mobility of participating nodes in DFL scenarios & \cite{Qu:bl_fog_computing:2020, Wang:edge_communication_optimization:2021}
\\ \cline{2-4}

& \multirow{1}{*}{\makecell[l]{Federation Communications KPIs}} & 
$\sbullet[0.75]$ Measure the reliability of model parameter exchanges between participants in DFL & \cite{Georgatos:efficient_adaptive_local_links:2022, Nadiradze:quantized_models:2021} \\
& & $\sbullet[0.75]$ Evaluate communications flexibility and overhead in DFL & \cite{Ye:dfl_unrealiable_comms:2022, Wang:edge_communication_optimization:2021}
\\ \cline{2-4}

& \multirow{1}{*}{\makecell[l]{Federation Models KPIs}} & 
$\sbullet[0.75]$ Measure the performance by solving different tasks or using multiple datasets for benchmarking & \cite{Zhao:network_system:2022, Kuo:healthcare_model_misconducts:2022, Vanhaesebrouck:multitask_collaboration_models:2016} \\
& & $\sbullet[0.75]$ Mechanisms to ensure the trustworthiness of federated models & \cite{Gholami:trusted:2022, Mothukuri:trusted_blockchain:2022, Kone:fl_improving_efficiency:2016, Sanchez:federatedtrust:2023, Wang:trust_DFL:2023}
\\ \hline

\multirow{1}{*}{\makecell[l]{DFL\\Optimizations}} & 
\multirow{1}{*}{\makecell[l]{Optimization of Federation Nodes}} & 
$\sbullet[0.75]$ Optimal methods for selecting nodes to transmit model parameters & \cite{Jiang:dfl_partial_gradient_aggregation:2020, Tedeschini:healthcare_tumor:2022} \\
& & $\sbullet[0.75]$ Optimize algorithms to suit the heterogeneous node resources and capabilities & \cite{Vepakomma:split_dl:2018, Mothukuri:survey_topology_architecture_privacy_decentralized:2021, Al-Abiad:computation_opt:2023}
\\ \cline{2-4}

& \multirow{1}{*}{\makecell[l]{Optimization of Federation Communications}} & 
$\sbullet[0.75]$ Efficient distribution schemes that minimize the number of network exchanges & \cite{Barbieri:selection_optimizer:2023, Jiang:bacombo_bandwidth_gossip:2020, Liu:comms_computing_cost:2022, Liu:comms_opt:2023} \\
& & $\sbullet[0.75]$ Compression techniques to reduce the amount of data that needs to be exchanged & \cite{Tang:communication_compression_efficiency:2018, Yuan:dgd:2016, Shi:ADMM:2014, Shi:EXTRA:2015, Deng:ADMM_plus:2017}
\\ \cline{2-4}

& \multirow{1}{*}{\makecell[l]{Optimization of Federation Models}} & 
$\sbullet[0.75]$ Explore efficient ML models to increase model performance and reliability during federation & \cite{Wu:network_gradient:2022, Alistarh:qsgd_optimization:2017, Basu:qsparse_local_sgd:2019,Shi:improving_model_dfl:2023, Nadiradze:quantized_models:2021} \\
& & $\sbullet[0.75]$ Use of meta-learning, multitasking, and federated distillation in federation tasks & \cite{Vanhaesebrouck:multitask_collaboration_models:2016, Li:dfl_metalearning_multitask:2022, Anil:google_knowdlege_destillation:2018, Seo:federated_knowledge:2020, Wu:knowledge_destillation:2022} \\ \hline \hline

\end{tabular}}
\end{table*}

\subsubsection{Federation Type}

The literature categorizes DFL architectures according to the type of federation, which is based on the participating nodes: cross-silo and cross-device. The differences lie in the number of participants and the amount of data stored in each one \cite{Marfoq:dfl_optimization_comms:2020}.

In cross-silo DFL, the nodes are organizations or data centers \cite{Han:dfl_cross_silo:2022}. There is usually a relatively small number of nodes ($<$100), each with a large amount of data (about millions of samples), perhaps distributed and aggregated from consumers in different businesses. At the same time, nodes use consistent, robust, and scalable computing over time. Likewise, these nodes have a high performance in the network, avoiding points of failure during communications between nodes. For example, it is applied in different hospitals by training a federated model for tumor classification while keeping their Positron Emission Tomography (PET) images locally \cite{Shiri:healthcare_framework:2022}.

In cross-device DFL, the number of nodes is relatively large ($>$100), where each node has a relatively small amount of data (about thousands of samples) and limited computational power \cite{Karimireddy:dfl_crossdevice:2021}. Due to concerns regarding power consumption, individual devices cannot be asked to perform complex training tasks. Furthermore, nodes could periodically disconnect from the network so that the network dropout rate would increase considerably, negatively impacting DFL performance. The nodes are usually on-edge devices or robots like UAVs \cite{Feng:dfl_horizontal_bl:2022}, where communications between devices are often weak if not kept in a close coverage radius.

\subsubsection{Participant Role}

Network nodes may have one or more roles that define their behavior and operation in DFL architectures. In particular, the existing roles are trainer, aggregator, proxy, and idle. A trainer node aims to train a local model with its local dataset and transmit the parameters to its neighboring nodes \cite{Wilt:scatterbrained_framework:2021}. The trainer node expects to receive the parameters of the updated federated model to incorporate it back into the local model. In contrast, the node with the aggregator role is responsible for obtaining the parameters from the neighboring nodes, aggregating them in the global model, and transmitting them to the neighboring nodes. In specific network topologies (see Section~\ref{sec:network_topology}), aggregators are not directly reachable by other nodes, and proxy nodes are needed. This role is intended to relay the received model parameters to neighboring nodes, allowing different nodes or network topologies to be interconnected \cite{Kalra:proxyfl_proxy_model_sharing:2021}. However, a participant may not have any of the above roles, being idle in the network and not participating in the federation.

\subsubsection{Decentralization Schema}

\begin{figure*}[!t]
\centering
\includegraphics[width=0.85\textwidth]{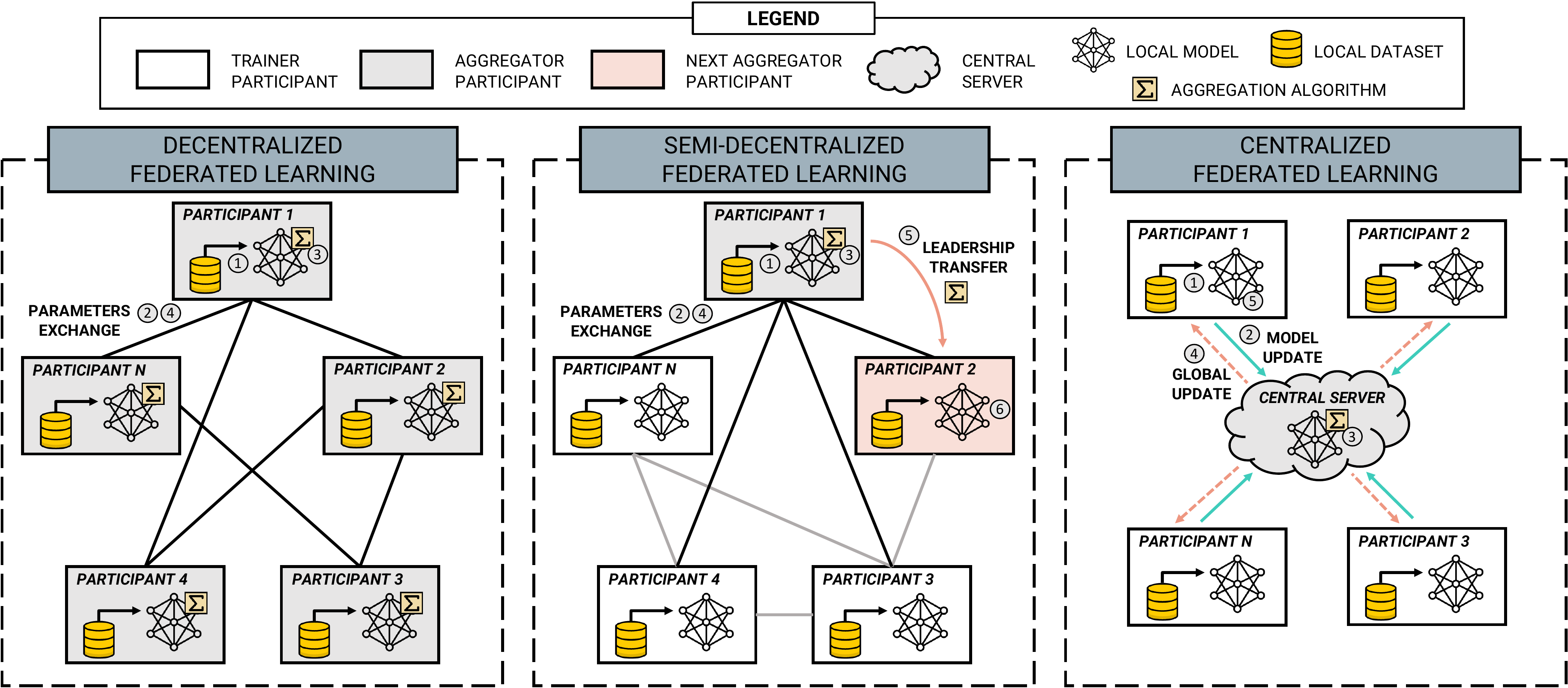}
\caption{Common approaches based on DFL architecture decentralization. The numbers inside the circles represent the training lifecycle.}
\label{fig:decentralization_approaches}
\end{figure*}

The federation architecture can be affected by the decentralization level maintained between participants, with three different approaches emerging (see \figurename~\ref{fig:decentralization_approaches}): DFL, Semi-Decentralized Federated Learning (SDFL), and CFL. In DFL, participants perform four steps independently: local model training, parameter exchange, local model aggregation, and parameter exchange again. In SDFL, participants perform the first two steps, while an aggregator participant handles the third step and transfers leadership for the aggregation functionality (step 5). In CFL, a central server handles parameter aggregation (step 3), with the rest of the network receiving and updating their local models accordingly (steps 4 and 5).

In DFL architectures, the network is fully autonomous, managing communications with other nodes and aggregating the model parameters independently of the other nodes. Each network node can select one or more nodes to transmit data, defining bidirectional communications between them. Furthermore, decentralization can be assumed to be fixed or dynamic, as the interconnections between nodes can change over time. Unlike the node-server architecture, the absence of a centralized server in DFL allows for the relaxation of the synchronous model update. Different solutions in the literature make several assumptions about network connectivity, particularly considering that each node is connected to all other nodes in the network or only to a set of neighboring nodes. Therefore, issues such as fixed or dynamic topology arise and the use of directed or undirected graphs \cite{Xu:udfl_iot:2022}. A fixed topology provides stability and predictability but limits adaptability, while a dynamic topology allows flexibility but requires additional overhead to maintain connectivity. In contrast, directed graphs model asymmetric relationships, and undirected graphs are more flexible for modeling symmetric relationships between participants.

In contrast to the previous approach, SDFL architectures maintain an aggregator role that rotates among the participants belonging to the network. The aggregator transmits the leadership (aggregation functionality) periodically during the federation, selecting the neighboring node randomly or based on its performance, such as network, computational, or power capacity \cite{Yemini:dfl_semi:2022, Lin:semi_d2d_aggregation:2021}. The leadership in SDFL architectures plays an essential role in determining which node is responsible for collecting the model updates and computing the next-generation model by aggregation. Also, the choice of leadership selection mechanism can significantly impact the performance, efficiency, and robustness of the system.

Finally, in CFL-based architectures, data decentralization is orchestrated by a single aggregator. This central node aggregates the model parameters shared periodically by each trainer and creates a robust central model distributed among the nodes \cite{Hard:google_centralized_to_dfl:2021}. In this approach, the aggregating participant is fixed during federation, with the central node being responsible for aggregating model parameters for each trainer in the network. Additionally, the CFL architecture makes it possible to ignore problems related to topology and device performance by not having decentralized aggregators. To improve this situation, techniques such as Multiparty Computation (MPC) can guarantee the immutability of model updates and coordination between participants directly, as could be the case for healthcare institutions \cite{Kanagavelu:dfl_mpc:2022}.

\subsubsection{Data Distribution}

The DFL approach, like CFL, has different configurations depending on the properties of the distributed data. Data can be Independent and Identically Distributed (IID) when the federation participants behave similarly. Therefore, the data distribution does not fluctuate ($x^{i} \sim D$), and the events associated with the data points are independent, i.e., they are not connected to each other in any way ($\nexists j \enspace p(x^{i}, x^{j}) = p(x^{i}) \; p(x^{j})$), where $p$ represents the joint probability distribution of the data points. However, in some application scenarios, nodes are usually heterogeneous. Hence, the data often have a non-IID format, where data vary in quality, diversity, and quantity in the network, thus increasing the complexity of modeling, analysis, and evaluation. The problem of heterogeneous data distribution poses a challenge for DFL. It can cause local model parameters to converge at different stationary points for different participants, making global convergence difficult. In addition, there might be multiple aggregated models during the federation where nodes are connected irregularly, leading to different global models receiving data with varying distributions. Addressing this challenge involves determining how to handle these various models. One option presented in the literature is to continue aggregating them at a higher level \cite{Hsieh:non_iid_optimization:2020}, while another option is to accept the presence of different federated models based on the network topology and characteristics of the nodes \cite{Gao:dfl_magazine:2023}. By considering these alternatives, DFL can better address the issues arising from non-IID data distributions.

Finally, depending on the nature of the problem to be solved and the organization of data (features and samples) among nodes, DFL is usually divided into three types: Horizontal Federated Learning (HFL) \cite{Feng:dfl_horizontal_bl:2022}, Vertical Federated Learning (VFL) \cite{Sanchez:demeter:2022}, and Transfer Federated Learning (TFL) \cite{Li:dfl_knowledge_transfer:2022}. HFL is the most commonly employed method in DFL, as it involves sample federation and is applicable when there are many overlapping features and few overlapping nodes, typically associated with cross-device scenarios. In contrast, VFL and TFL require more in-depth analysis and comparison, as they are more complex to adapt and deploy in application scenarios that take advantage of their unique data organization. VFL focuses on feature binding when there are many overlapping nodes and few overlapping features, while TFL is considered when there is a limited feature and sample intersection between nodes. The goal of TFL is to build efficient models for specific applications in cases where data are sparse, which makes it another relevant aspect of DFL strategies.

The successful transition from CFL to DFL involves the adaptation of commonly used CFL datasets. Techniques like decentralized sparse partitioning prove instrumental in aligning these datasets with the unique characteristics of DFL \cite{Dai:dispfl_framework:2022}. These techniques consider the aggregation algorithms, participants' roles, and the task to be solved. Similarly, the distinct data distribution across nodes in DFL could be addressed using federated sampling techniques, which help maintain data diversity while ensuring statistical accuracy \cite{Shi:personalization_dfl:2023}. These adaptations make it feasible to use CFL datasets in DFL scenarios efficiently, leading to optimized learning processes and improved performance of DFL implementations.

\subsection{Network Topology}
\label{sec:network_topology}

\begin{figure*}[!ht]
\centering
\includegraphics[width=0.85\textwidth]{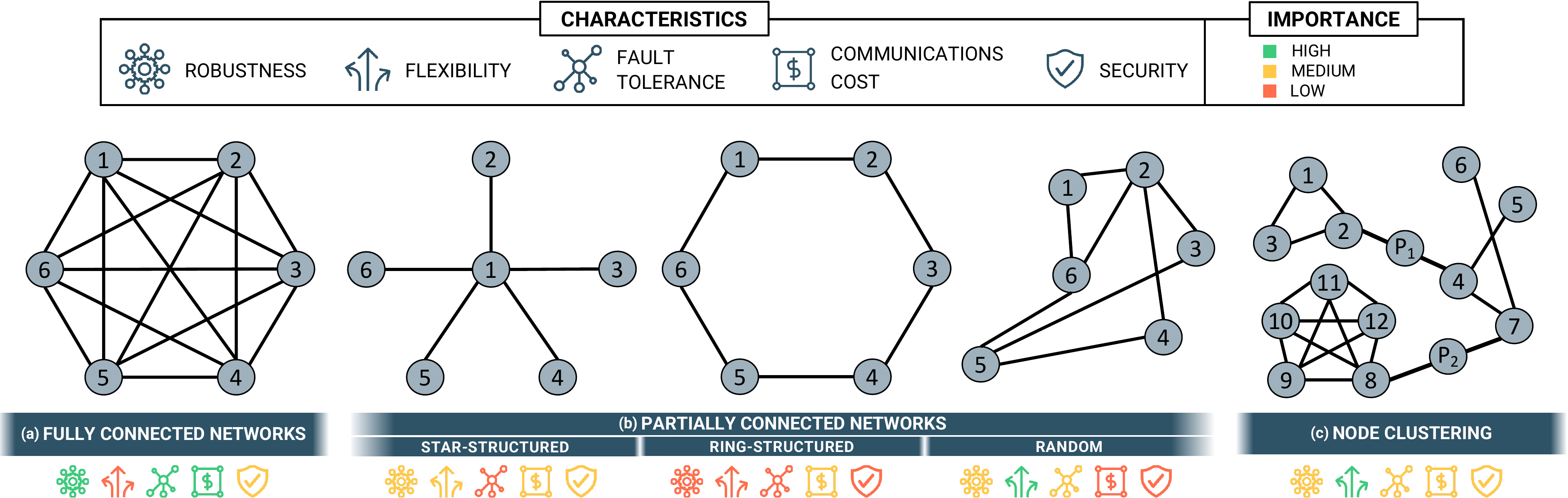}
\caption{DFL network topologies. (a) Fully connected networks, (b) Partially connected networks (star-structured network, ring-structured network, random network), (c) Node clustering. \textit{$P_n$} indicates the node acting as a proxy.}
\label{fig:network_topology}
\end{figure*}

In DFL, the network topology describes the organization of participants and determines their communications. Therefore, the importance lies in the impact on convergence, generalization, overhead, and robustness of the DFL approach. There are three network topologies used in DFL: fully connected networks, partially connected, and node clustering. \figurename~\ref{fig:network_topology} shows these network topologies and their characteristics. In particular, the topologies are evaluated depending on the importance (high, medium, low) of robustness, flexibility, fault tolerance, communications cost, and security.

\subsubsection{Fully Connected Networks}

This network topology maintains direct links between all pairs of nodes. When a new node is added to the system, a link must be added to each node (see \figurename~\ref{fig:network_topology}a) \cite{Georgatos:efficient_adaptive_local_links:2022}. Thus, the communication cost is high and grows with the number of nodes in the network. Moreover, adding new nodes increases the complexity of managing connections for each node, resulting in low flexibility. Despite the high communication cost and low flexibility, this topology is highly reliable and robust, as the network can still function when a few nodes or links fail \cite{Xiao:military_fullydl:2021}.

\subsubsection{Partially Connected Networks}

In some network topologies, participating nodes may only have direct links to some nodes in the network, regardless of the shape of the topology. Therefore, the basic transmission cost for each node is lower than in fully connected networks since the model parameters are not transmitted to all nodes. As a result, communication cost may be higher than in fully connected topologies since the model parameters transmitted may go through several intermediate trainer participants until they reach the aggregating participant, resulting in a longer delay \cite{Vogels:toplogy_decision_convergence_models:2022}. This work divides partially connected networks into star-structured, ring-structured, and random.

In a star-structured network, one of the nodes acts as a proxy participant and enables the federation communication with all other nodes (see the first subfigure of \figurename~\ref{fig:network_topology}b). When a new participant is added, only one link is needed to connect them to the central node. Consequently, the communication cost grows linearly as the number of nodes increases, and the resources of the node limit network scalability. However, since all communications between non-central nodes must pass through the central node, it becomes a potential bottleneck, reducing flexibility. Additionally, the network has low fault tolerance, as the failure of the central node disrupts the communication between all nodes. Some DFL solutions use temporary connections in star topologies, but establishing and terminating these connections adds significant overhead.

In a ring-structured network, nodes are connected circularly (see the second subfigure of \figurename~\ref{fig:network_topology}b). The communication cost grows linearly as the number of nodes increases, as each node only maintains two fixed links, regardless of network size. This contributes to medium flexibility. However, as the number of nodes grows, transmission delays for model parameters also increase. Ring networks can be unidirectional, where nodes transmit model parameters to only one neighbor, or bidirectional, where nodes send parameters to both neighbors, providing higher reliability and fault tolerance than unidirectional rings.

Sometimes a defined topology is not supported by the federation, and nodes generate connections following heuristics based on the proximity or computational capacity of nearby neighbors. Due to intermittent connections in DFL, the network maintains a dynamic node structure, creating random network topologies over time \cite{Chow:expander_dfl:2016}. In this sense, some topologies use expander graphs and deterministic local optimization algorithms to reduce communication overhead in these networks. The third subfigure of \figurename~\ref{fig:network_topology}b represents a network topology with six random nodes. Specifically, it describes a network based on the Erdös–Rényi model, where links between nodes are established randomly \cite{Hua:topology_random:2021}. This type of network offers high flexibility and moderate fault tolerance but can result in higher communication costs and variable security levels.

\subsubsection{Node Clustering}

Recent publications have identified another DFL network topology for creating hierarchical clusters with models adapted to the distribution of nodes. In this regard, the literature addresses two approaches: similarity-based clusters \cite{Duan:proxy_similarity:2021} and proxy-based clusters \cite{Kalra:proxyfl_proxy_model_sharing:2021}. In the former, clusters are determined by the similarity of the local model parameters of the nodes. As a result, each cluster is more individualized for the nodes that compose it, obtaining a homogeneous performance between nearby nodes. In contrast, proxy-based clusters aim to use nodes that interconnect different topologies, forwarding model parameters. In this regard, proxy nodes have to be evaluated in terms of performance since they transmit data from one cluster to another, generating an important bottleneck in the overall architecture \cite{Al-Abiad:computation_opt:2023}. The disadvantage of this clustering method implies that all nodes are initially linked to a coordinating node, which is not feasible in all application scenarios. Furthermore, since nodes within a cluster share similar data distributions, individually trained cluster models may be less generic and robust than models exposed to global data distributions \cite{Briggs:hierarchy_cluster_fl:2020}. \figurename~\ref{fig:network_topology}c shows the usage of proxy nodes to interconnect topologies in node clustering. 

\subsection{Communication Mechanisms}

This fundamental defines the procedures that enable DFL communications between network nodes. First, the communications scheme in DFL, based on synchronous, asynchronous, and semi-synchronous communications, is detailed. Then, this subsection presents two different decentralized communications mechanisms: peer-to-peer (P2P) \cite{Chen:p2p:2022} and gossip \cite{Hu:gossip:2019}.

\subsubsection{Communications Scheme}

The communications schemes determine the behavior of the nodes when transmitting or receiving model parameters from neighbors and then aggregating them. Depending on the type of scenario and the objective pursued, DFL can be supported by synchronous, asynchronous, or semi-synchronous communications. The first alternative requires each node to perform local training, consisting of multiple steps, usually expressed in epochs. An epoch is defined as a complete iteration through the entire local dataset. After completing local training, all nodes acting as trainers send the parameters of their local models to all their neighboring nodes and receive the new model they have to integrate. This training technique, known as synchronization points, is repeated for several federation rounds. In this sense, the nodes will have to wait for the end of the round to transmit their model parameters again. Therefore, a drawback of this scheme is its slow convergence due to waiting for slow or idle participants. To address this problem, over-selection techniques could increase participant selection but also lead to resource waste and selection bias \cite{Bonawitz:google_sync:2019}.

In asynchronous communications, no synchronization point exists, and nodes can transmit and receive model parameters independently of the other federation participants. Because there is no idle time for participating nodes, asynchronous protocols have a faster convergence speed \cite{Cao:dfl_async:2021}. However, they cause higher communication costs and lower generalization due to staleness \cite{Zehtabi:sync_async_thresholds:2022}. It is also necessary to detect stragglers (i.e., slow devices) in the federation, delaying data propagation through the network topology. In the same way, aggregators need aggregation policies adapted to the number of model parameters received, as they vary during federation rounds. Nevertheless, asynchronous communications are well suited for cross-device DFL scenarios, in which nodes have varying computational capability and intermittent availability \cite{Pinyoanuntapong:trust_sync_async_edgeiot:2022}.

Finally, semi-synchronous communications provide a balance between resource usage and communication costs. In this schema, each node trains locally until a predefined synchronization point is reached. Consequently, nodes with varying processing capacities and data volumes perform different numbers of epochs. The batch serves as the basic computation unit, allowing for finer control over when a node contributes to the federation. Some studies, such as \cite{Zehtabi:sync_async_thresholds:2022}, introduce thresholds for each device that compare the changes in local model parameters with available local resources to determine the benefits of aggregation at each round.

\subsubsection{Peer-to-Peer Communications}

Traditionally, the deployment of overlay networks has been used mainly for P2P message exchange, online social networks, and routing infrastructures. DFL leverages P2P message exchange networks for local model parameter exchanges at each node. However, unlike traditional P2P networks, DFL deals with a dynamic and heterogeneous topology, where participants often change their location or role in the federation \cite{Chen:p2p:2022}. In this regard, selecting $d$ random neighbors to exchange information is not always feasible. The main drawback of DFL is that nodes cannot randomly choose $d$ neighbors among existing nodes since there is no central coordinator to determine these associations \cite{Wink:p2p:2021}. However, each node can determine neighbors if global information is given when the topology is created, and neighbor discovery mechanisms are implemented during federation \cite{Wang:p2p_graphs:2022}. Other alternatives are to employ distributed Delaunay triangulation networks for wireless sensor networks or regular random topologies for data center and memory interconnection networks \cite{Schwab:delaunay_distributed:2021}.

\subsubsection{Gossip Communications}

Gossip communications enable asynchronous transmissions of model parameters between network nodes. This type of communication relies on the P2P sampling service with the objective that nodes in the federation communicate and exchange data \cite{Jiang:bacombo_bandwidth_gossip:2020}. The comparison of the DFL approach using gossip communications versus the CFL approach has been extensively studied, with superior performance obtained in the former when a better understanding of model parameters is employed \cite{Hegedus:gossip_vs_centralized_fl:2021}. However, this mechanism maintains a restrictive functionality where communication occurs between close neighboring nodes. To address this limitation, additional processes are presented to determine the location of a node using random walks \cite{Ye:randomwalk_optimization:2020} and node selection techniques, relying on the heuristics of the distance between nodes or underlying device features \cite{Belal:recommender_gossip:2022}. For instance, local models can be updated with partially received parameters and be optimized by mixing weights according to the matrix of communication link reliability \cite{Ye:dfl_unrealiable_comms:2022}. Finally, some application scenarios of gossip communications in DFL are providing accurate recommendations (e.g., location logs, movie ratings) \cite{Belal:recommender_gossip:2022} or in Distributed Online Social Networks (DOSN) where users can directly communicate only with their immediate neighbors in a social graph \cite{Khelghatdoust:socialnetworks_gossip:2022}.

\subsection{Security and Privacy}

Decentralized scenarios may be more susceptible to attacks than CFL scenarios as the number of participants grows within the federation. Nevertheless, the level of vulnerability largely depends on the network topology. For example, in a fully connected network, where each participant has numerous connections, the impact of an attacker might be less severe than in a centralized CFL scenario since the attacker can only compromise a limited portion of the network. Nonetheless, decentralized scenarios still face increased risk due to the numerous intermittent and often weak connections between the participants. 

In this sense, DFL can be subject to various attacks, including malware deployment and adversarial attacks, which can affect federation models and decentralized data, similar to the CFL approach \cite{Rawat:botnet:2021}. While the former exploits intrinsic DFL fundamentals in the architecture or communications to affect the federation adversely, the latter aims to infer federated models and local node data, manipulate preconditions or destroy the model. Notable among these attacks are model inversion or membership inference, which can potentially infer the raw data by accessing the model \cite{Sanchez:demeter:2022}. Despite these potential attacks, countermeasures have been adopted to address them, preserve data and model privacy, and develop recovery plans to minimize harm and disruption. In particular, this paper presents two main approaches adopted in current DFL scenarios for data protection: cryptographic methods and differential privacy.

\subsubsection{Malware Attacks}

Malware, also known as ``malicious software'', is a critical issue in the DFL approach due to the fundamentals based on decentralized architectures without a server coordinating the scenario. Malicious entities with high propagation through the network can affect behavior and interfere with its proper functioning. This issue differs in impact compared to more traditional approaches such as CFL, mainly because DFL does not have a centralized entity to coordinate the cooperation of participants in the federation. Thus, malware can be created and deployed directly among new network participants. Due to the number of connections to neighboring nodes, fully connected networks, star-structured networks, and random networks are the most affected in terms of security. While the first ones maintain communication links between each pair of nodes, the others establish links with either a central node or random nodes that could be malicious or dishonest.

Regarding communications between nodes, malware also has a significant impact due to the high number of exchanges in DFL. The exchanged messages are obtained by neighboring participants, extracted, and processed for aggregation or forwarding in the case of proxy participants. However, the information obtained may have been modified during the exchange, adding malicious data using rootkits or including the participant in a botnet \cite{Yin:botnet:2017, Rawat:botnet:2021}. In this way, the node would be infected and could spread easily and quickly among the federation participants, especially when there are many links between neighbors.

\subsubsection{Adversarial Attacks}

DFL must address adversarial attacks caused by malicious nodes participating in the federation. These attacks, inherited from traditional CFL operations, include poisoning attacks on HFL and VFL approaches \cite{Sanchez:demeter:2022}. Furthermore, DFL is prone to poisoning attacks for the following reasons: (i) DFL scenarios involve numerous participants, increasing the likelihood of faulty behavior from one or more nodes; (ii) local training data and training processes of participants are hidden from other nodes in the network, making it impossible to verify the authenticity of updates sent by participants; and (iii) multiple participants may generate vastly different local updates. Consequently, secure aggregation protocols might be negatively affected by aggregation performed by malicious nodes, rendering it infeasible to audit local updates \cite{Cheng:secureboost:2019}. In this sense, the paper at hand analyzes such attacks on DFL solutions to assess their impact on the network.

Adversarial attacks can be classified based on the timing of the attack (training or evaluation time) and frequency (one-shot or multiple) \cite{Rodriguez:survey_fl_threats:2023}. Among training-time attacks, Byzantine attacks are prevalent. In these attacks, malicious nodes disrupt model training by transmitting poisoned, corrupted, or fake model updates to other nodes in the network. Since malicious nodes in DFL keep their data and aggregations private, they can poison their model and distribute their parameters without consequences. These attacks can be executed in a one-shot manner each round or multiple times in asynchronous communication scenarios. Similarly, Sybil attacks create multiple fake pseudonym nodes that flood DFL networks. It is difficult to mitigate these threats since no central authority can verify node identities and distinguish between pseudonym and non-pseudonym nodes. Using pseudonyms, an attacker can launch DDoS attacks and disrupt the DFL network. The attacker can command multiple pseudonyms to flood a targeted node with pairing requests, rendering it unable to respond to and communicate with legitimate nodes \cite{Verbraeken:bristle_noniid_attacks:2021}. Additionally, an adversary can corrupt the reputation system of decentralized applications. Nodes could obtain a false high reputation, being forced to participate in the network and being queried by other nodes, thus degrading their performance.

Other attacks attempt to infer DFL communications with a high attack frequency. One of these threats is Eclipse attacks, which enable attackers to gain partial or full control of a participant by manipulating its connections with neighboring nodes \cite{Alangot:eclipse_attack_defense:2021}. In such attacks, the attacker nodes control many connections, which allows them to ``eclipse'' the victim, rendering messages that should reach it useless. The attackers could gain complete control of all traffic to the victim if all node connections belong to them \cite{Niu:eclipse_attack:2022}. Free Rider attacks are also a type of attack that takes advantage of DFL without contributing useful data to it during the training process \cite{Wang:freerider_attack:2022}. Finally, in evasion attacks, malicious nodes mimic participation in DFL by generating misleading updates, crafting inputs that manipulate the model into producing faulty collaborative models.

\subsubsection{Security Countermeasures}

The use of countermeasures is essential in the DFL approach to prevent malicious participation and preserve data privacy. DFL establishes a federation by sharing model parameters, such as gradients, weights, or meta-level information among participants. However, exchanging gradients or meta-level information during distributed training can reveal information about the training data and model updates, making exchanging model weights preferable for privacy preservation \cite{Lu:dfl_gradient_inversion_attack_defense:2023}.

Additional countermeasures comprise several techniques inherited from CFL architectures, as reported in \cite{Qu:privacy_framework_iot:2022}. Most of them are employed during decentralized federation to obfuscate updates at the cost of reducing the accuracy of each local model, relying on traditional Differential Privacy (DP) relaxations to inject less noise. However, this technique may degrade the performance of constrained participants, leading to suboptimal results in the federation. In DFL, each participant needs to perform a significant computation to add or suppress noise per set of model parameters transmitted on each communication link, making it more challenging to adapt than CFL \cite{Chen:wireless_privacy:2022}. Other techniques, such as data augmentation and obfuscation, can prevent sample reconstruction in local training. However, model parameters shared among neighbors would be equally exposed to malicious nodes. A promising solution is to combine DP mechanisms with secure aggregation and additive Homomorphic Encryption (HE), depending on the application scenario and required security level, and deploy them on a larger number of federation participants. In other cases, it is preferable to rely on anomaly detection mechanisms \cite{Wang:p2p_iot_cyber_anomaly_detection:2021} or model misbehavior detection \cite{Kuo:healthcare_model_misconducts:2022} to detect malicious actors that may deteriorate DFL performance. While the former focuses on the data supplied to the model, the latter analyzes the internal performance of the model depending on the task. In these cases, detection could involve the deployment of mitigation to reduce the severity of a cyberattack. Recent research has further augmented defensive capabilities by integrating Moving Target Defense (MTD) techniques. Specifically, random neighbor selection and dynamic IP/port switching offer enhanced resilience against communication-based threats \cite{MartinezBeltran:dfl_mtd:2023}. Moreover, anomaly detection has significantly improved by incorporating time series analysis techniques \cite{PeralesGomez:temporalfed:2023}.

Finally, cryptographic methods, such as Secure Multiparty Computation (SMC), are widely used in DFL to preserve the privacy of exchanges between neighboring nodes \cite{Kanagavelu:dfl_mpc:2022}. Federation participants must encrypt their messages before sending them, operate on them, and decrypt the encrypted output to obtain the final result. Despite the security provided by SMC, nodes can experience a significant computational overhead, similar to previous DP techniques. To address this issue, new approaches have emerged to provide adequate security performance while maintaining efficient resource usage. One such approach is based on deploying a Trusted Execution Environment (TEE) \cite{Chen:worm:2020}, such as Intel SGX processors, which can protect the code and data loaded inside it. Each node in the network can use this environment to increase its trustworthiness while allowing model parameters to be securely added during federation. In addition to cryptographic mechanisms and secure environments, DFL also requires mechanisms that guarantee the decentralization of the architecture while ensuring inter-participant reliability. DLT technologies, such as Blockchain, provide an ideal solution for this requirement by ensuring the immutability of exchanged models \cite{Qi:dfl_bl:2022}. It is accomplished through a P2P consensus scheme governing the network, where the model parameters of each participant can only be updated through consensus. In this sense, a novel solution uses a lightweight Blockchain with a unique consensus protocol and storage strategy, significantly reducing overhead while guaranteeing privacy. This approach optimizes performance by mitigating slower device impacts and maximizing storage utilization \cite{Ma:torr_bl:2023}. At the same time, a transparent and immutable reward system ensures the trust of the participants by detecting malicious activities.

\subsection{Key Performance Indicators}

\begin{figure*}[!ht]
\centering
\includegraphics[width=0.8\textwidth]{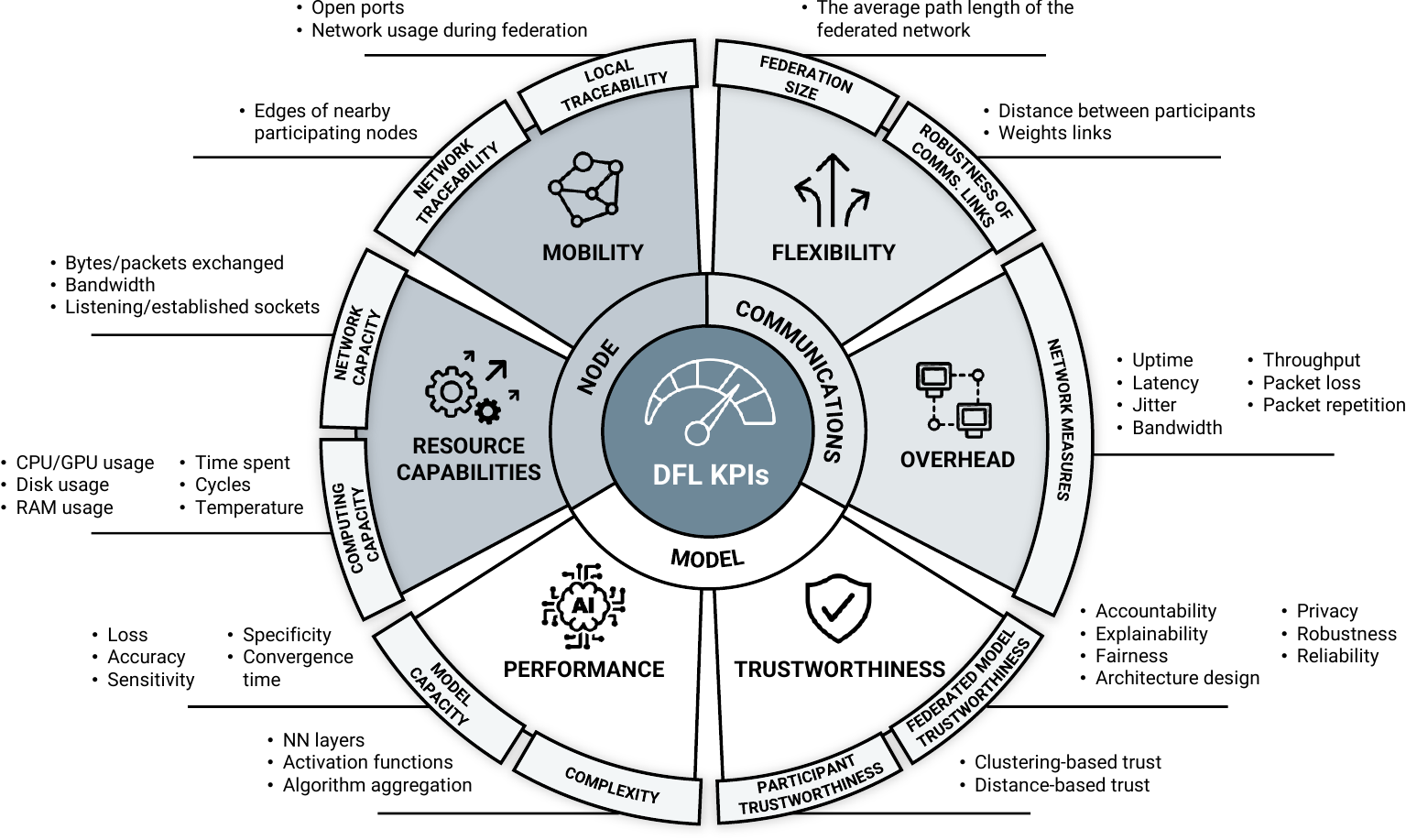}
\caption{Key Performance Indicators for DFL.}
\label{fig:kpis}
\end{figure*}

KPIs are necessary to measure the efficiency and suitability of the activities and components of DFL architectures. It is essential to know these metrics to detect and resolve federation, communications, or security issues more quickly, reducing the frequency and duration of incidents while increasing efficiency \cite{Verhaelen:kpi:2021}. In this sense, KPIs are utilized for the proper evaluation of the main components of the DFL approach: (i) the federation nodes, detailing the evaluation in terms of mobility and resource capabilities; (ii) federation communications, defining the flexibility and overhead of the deployed architecture; and (iii) the federation models, reviewing their performance and trustworthiness in solving distributed ML tasks (see \figurename~\ref{fig:kpis}).

\subsubsection{Federation Nodes KPIs}

DFL must have KPIs to evaluate the participating nodes of the federation. Thus, the heterogeneity and dynamism of the nodes in DFL scenarios are relevant factors adding complexity to the network. In this sense, \textit{resource capabilities} and \textit{node mobility} are identified as promising KPIs to determine node performance. The \textit{resource capabilities} KPI comprises computational and network capacity.

\begin{itemize}
    \item \textit{Computing capacity}. It provides internal participant resource indicators for local training computation and model parameter aggregation during federation. Some metrics related to this category are CPU/GPU, disk, or RAM usage. One of the aspects of computational capacity is the autonomy of the node to address the tasks. In this sense, time, cycles, and temperature are considered \cite{Qu:bl_fog_computing:2020}.
    \item \textit{Network capacity}. It determines the node ability to manage communications with one or more neighbors. In this regard, the following metrics are important to analyze: the number of bytes transmitted and received, the number of packets transmitted/received, the bandwidth, or the number of listening/established sockets \cite{Wang:edge_communication_optimization:2021}. 
\end{itemize}

In addition to \textit{resource capabilities}, DFL presents the \textit{node mobility} KPI to evaluate the dynamism of the network. Therefore, nodes can be easily affected in communications with other nodes, generating intermittent inputs and outputs in the network. The \textit{node mobility} KPI is determined by node traceability in both the network and local.

\begin{itemize}
    \item \textit{Network traceability}. It monitors the state of each node and traces its mobility in the network. For this purpose, this metric computes the edges of nearby participating nodes to determine their link with other participants in the federation, which denote the number of near neighbors. In this way, the network can extract information about the position, direction, and absolute velocity of each node \cite{Zhao:network_system:2022}.
    \item \textit{Local traceability}. It allows determining the DFL network to contribute by monitoring local interfaces and their characteristics. Thus, it is possible to determine the mobility intent. Metrics such as the number of open ports or network usage during federation are identified \cite{Huang:fl_healthcare_mobility:2019}. Network traceability metrics usually complement them in the evaluation of node performance.
\end{itemize}

\subsubsection{Federation Communications KPIs}

Model parameter exchanges between nodes in a DFL approach might experience instability due to asynchronous communications, leading to nodes processing and exchanging data at varying rates. Consequently, this could result in delays and inconsistencies in the learning process, as some nodes may handle outdated information. Furthermore, the aggregation process is complicated by heterogeneous aggregator devices with diverse computational capabilities and communication protocols. Such heterogeneity can cause inconsistencies across different network segments, ultimately having a negative impact on the accuracy of the models for a group of participants. Moreover, asymmetric topologies may cause uneven data distributions or processing capabilities among nodes, posing challenges in attaining a uniformly distributed learning process. In this sense, \textit{communications flexibility} and \textit{network overhead} are essential for optimal balance. Regarding the \textit{communications flexibility}, the DFL approach is evaluated according to the following elements.

\begin{itemize}
    \item \textit{Federation size}. It provides an evaluation of the predisposition of the architecture to exchange model parameters between nodes. This metric accurately calculates the average path length of the network, representing a stable number strongly affected by the movement of the nodes and, thus, the flexibility of the network. Reduced network size improves parameter transmission performance by reducing interference \cite{Georgatos:efficient_adaptive_local_links:2022}.
    \item \textit{Robustness of communication links}. It determines the strength of the links with the other neighbors participating in the federation. Some metrics that allow its evaluation are the distance between nodes, reactance, and reliability. Link weights are often used to support evaluations, with high weights being maintained for links with a high percentage of correctly received transmissions \cite{Ye:dfl_unrealiable_comms:2022}.
\end{itemize}

In the context of DFL, several scenarios require decentralized technologies, which can face significant performance issues. One of the main challenges arises from the communication-intensive nature of DFL applications, which require frequent model parameter transmissions involving large amounts of data. As the federation progresses or the model becomes more complex, this communication becomes even more intensive, leading to high \textit{network overhead}, where a high overhead can lead to applications failing to meet response time, quality, or resource utilization requirements. \textit{Network overhead} is evaluated from the following metrics \cite{Wang:edge_communication_optimization:2021}.

\begin{itemize}
    \item \textit{Uptime}, often known as availability, reflects the time of use of the federated network since it became available to participants.
    \item \textit{Latency} measures the time a data packet travels between network nodes.
    \item \textit{Jitter} indicates the variability of a latency time and measures the consistency of a network transfer rate. Very low jitter is preferred when a real-time reaction is required (e.g., warning of enemy UAVs).
    \item \textit{Bandwidth and throughput}. While the first indicates the quantity of data a network path is anticipated to support or successfully convey from one point to another at a specific time, the second describes the amount of data transported between nodes inside a network path.
    \item \textit{Packet loss} can denote congestion, low bandwidth, and interference. It refers to packets transferred from one device to another but fail to arrive at their destination. Packet loss is the ratio of packets received at the destination to packets sent from the source.
    \item \textit{Packet repetition}. When a data packet is successfully sent but does not reach its destination, it must be retransmitted. Retransmissions occur in all networks; however, they are more common in wireless networks due to low signal strength, concealed nodes, and interference.
\end{itemize}

Both \textit{communications flexibility} and \textit{network overhead} are mainly limited to the model parameters exchanges during federation for each $K$ step. Assuming $M$ is the parameter size of the model, there would be $N-1$ times communications in a round, being the average overhead $M\cdot(N-1)$. Thus, $K$ could reduce the communication frequency, decreasing the performance of DFL. A summary of the communication complexity comparison with CFL architectures is shown in \tablename~\ref{tab:communication_complexity}, where $\left [\;\right]$ denotes the standard rounding function. The table shows that the total volume of data transferred per DFL round is comparable across all three architectures. However, the communication overhead per node is different among decentralized architectures. Specifically, SDFL has the potential to achieve better performance in node communication overhead, which could benefit system bandwidth utilization and increase robustness. Overall, it is important to consider the trade-off between communication overhead and other features when designing decentralized architectures with FL.

\renewcommand{\arraystretch}{1.8}
\begin{table}[ht!]
\caption{Comparison of the flexibility of federated communications between DFL, SDFL, and CFL.} \label{tab:communication_complexity}
\resizebox{\columnwidth}{!}{
\centering
\begin{threeparttable}
\begin{tabular}{cccc} 
\hline
\makecell{Federated\\Architecture} & \makecell{Communication\\Times/Round} & \makecell{Node\\Overhead} & \makecell{Total Transferred Data\\Volume per Round (MB)} 
\\ \hline \hline

DFL &
1 &
$M\cdot(N-1)$ &
$M\cdot(N-1)^2$

\\[3pt] \hline 

SDFL &
$\left [\frac{N-1}{2}\right]$ &
$2\cdot M$ &
$2\cdot M\cdot N \left [\frac{N-1}{2}\right]$ *

\\[3pt] \hline

CFL &
1 &
$M\cdot N$ &
$M\cdot N$

\\[3pt] \hline \hline

\end{tabular}
\begin{tablenotes}
\item \textit * It considers the additional communication required for leadership transmission, which is not present in the DFL and CFL architectures.
\end{tablenotes}
\end{threeparttable}}
\end{table}
\renewcommand{\arraystretch}{1}

\subsubsection{Federation Models KPIs}

The \textit{performance} of the federated model in solving the task for which it has been created is critical. In other words, the performance indicates the success of DFL in generating accurate predictions or solving the intended problem. The following elements are used for measurement \cite{Kuo:healthcare_model_misconducts:2022}.

\begin{itemize}
    \item \textit{Model capacity}. It is based on comparing the model predictions with known values of the dependent variable in a dataset. Some relevant metrics are model loss, accuracy, sensitivity, specificity, and the time associated with the convergence of the model.
    \item \textit{Complexity}. It is influenced by both the functions of the model trying to learn and the nature of the training data. Some metrics are the number of layers in the NN, activation functions, and algorithm aggregation implemented by the federated model.
\end{itemize}

In addition to the above elements, federation models must be \textit{trustworthy} to ensure the technology is used to contribute productively to the task. If they are not trustworthy, nodes will not collaborate with these systems. The problem increases when model parameters are deliberately distributed among nodes in a topology. Therefore, it is necessary to determine the confidence of the participants and the federated models to ensure trustworthiness, where robust privacy and security measures significantly contribute to this \cite{Gholami:trusted:2022, Mothukuri:trusted_blockchain:2022}.

\begin{itemize}
    \item \textit{Participant trustworthiness}. It defines the confidence level of neighboring participants based on the model parameters exchanged. There are two main metrics, clustering-based trust, which determines the federation nodes trustworthiness relying on clusters of participating nodes; and distance-based trust, which is determined based on the distance between participants and the transmission time of the model parameters.
    \item \textit{Federated model trustworthiness}. It provides confidence in the local models of each participant in the federation. Normally it is calculated in the aggregator nodes before supplying the federated model to the rest of the participants. The metrics are divided into several pillars: accountability, analyzing the quality of the model life cycle (client registry, anomaly monitoring); explainability, providing clarification of the decision rationale (algorithm class, model size, DFL configuration); fairness, ensuring impartial and just decisions without discrimination of protected groups (participation rate, discrimination index); architecture design, creating a baseline behavior for the management of participants in the federation (data augmentation, regularized local loss); privacy, protecting federated model data (perturbation, anonymization, entropy); robustness, checking the resilience against adversarial inputs (confidence score, loss sensitivity); and reliability, predicting when an asset will fail or deteriorate (client dropout rate, data provenance, data quality).
    \item \textit{Security and privacy}. The preservation of security and privacy is paramount in DFL, even while allowing effective model training \cite{Sanchez:federatedtrust:2023}. DP is a widely used method that includes a privacy budget parameter, denoted as epsilon $(\epsilon)$, serving as a metric to quantify the maximum amount of privacy loss allowed. This budget diminishes with each data-sharing instance, with smaller budgets enhancing privacy but potentially impeding data utility. Another critical parameter in DP is delta $(\delta)$, which represents the probability of information inadvertently being leaked. Other privacy-preserving methods, such as HE and anonymization, also have vital performance indicators. In HE, the degree of noise introduced can be a significant parameter. Anonymization techniques, such as $k$-anonymity and $l$-diversity, use $k$ and $l$ values as performance metrics, ensuring privacy in DFL by making it difficult to re-identify models within a federation \cite{Wang:trust_DFL:2023}.

\end{itemize}

\subsection{DFL Optimizations}

\begin{figure}[!ht]
\centering
\includegraphics[width=0.8\columnwidth]{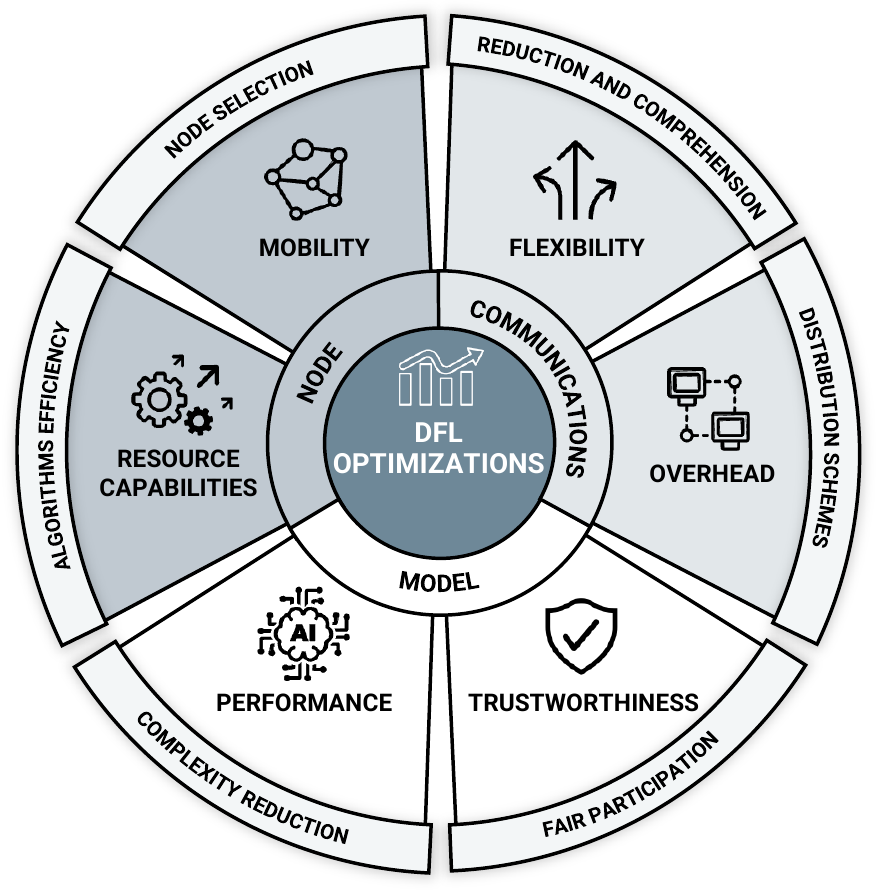}
\caption{DFL optimizations.}
\label{fig:optimization}
\end{figure}

Despite the advantages of DFL, a series of optimization mechanisms are necessary to guarantee the full efficiency of the approach. In this sense, DFL includes three main elements to optimize: federation nodes, federation communications, and federation models (see \figurename~\ref{fig:optimization}).

\subsubsection{Optimization of Federation Nodes}

An important optimization for improving node \textit{resources capabilities} and \textit{mobility} in the federation is the selection of nodes to transmit the model parameters. Node selection is divided into three different strategies.

\begin{itemize}
    \item \textit{Sequential selection of nodes}. Participants are selected at the beginning of the federation process, which iterates over time \cite{Mothukuri:trusted_blockchain:2022}.
    \item \textit{Random selection of nodes}. It excludes resource capabilities or network constraints from the participants. In this sense, a subset of nodes is obtained in a fixed time of the federated process to send the model parameters.
    \item \textit{Scheme schedule selection of nodes}. This selection strategy requires a node to act as an orchestrator of the communication links and the characteristics of each neighboring node. Furthermore, the controller can select participants based on a preassigned probability calculated according to participation in the federation \cite{McMahan:communication_efficient:2016}.
\end{itemize}

Additionally, it is relevant to optimize traditional aggregation algorithms such as Federated Averaging (FedAvg) to make it more suitable for the node resource capabilities \cite{McMahan:communication_efficient:2016}. FedAvg is a widely accepted heuristic algorithm used as a baseline for star-structured topologies due to its simplicity and empirical effectiveness. In this case, the aggregator node obtains the model parameters of the neighboring nodes, aggregating the average of the updates and computing the resulting global model. It is also worth noting that while in CFL, the FedAvg algorithm has been widely accepted as a baseline, in DFL, no algorithm has risen among the others. Customizations of FedAvg and new techniques focused on decentralized scenarios have emerged in the literature intending to adapt to DFL solutions.

\begin{itemize}
    \item \textit{Decentralized Stochastic Gradient Descent (DSGD)}. It provides optimization for a set of objective functions distributed over a network. Each node maintains estimates of the optimal optimization step on information concerning its cost function and exchanges these estimates directly with the other nodes in the network \cite{Nedic:alorithm_dfl:2009}.
    \item \textit{FedPGA}. It defines a partial gradient exchange mechanism that leverages node-to-node bandwidth to speed up communication time. At the same time, it reduces the convergence rate by adapting the step size of the stable gradient descent direction \cite{Jiang:dfl_partial_gradient_aggregation:2020}.
    \item \textit{Dynamic Average Consensus-based FL (DACFL)}. It transforms the aggregation of the DFL model into a local training procedure as a discrete time series, whose convergence guarantees convex objectives \cite{Chen:dacfl_algorithm:2022}. In the algorithm, nodes compute and update a local solution to a subproblem and share the solution with neighboring nodes. Moreover, the authors demonstrate that the algorithm adapts to dynamic networks and heterogeneous scenarios by tuning a node-specific local parameter based on the node resources \cite{He:fully_dfl_algorithm:2018}.
    \item \textit{Split Learning (SL)}. This method, called SplitNN \cite{Vepakomma:split_dl:2018}, enables collaborative training of NNs without sharing raw private data, but emphasizes the suitability for DFL architectures. It employs split models instead of full model replication in a DFL architecture. The training participants hold replications of the shallower layers up to a certain layer (i.e., the cut layer), while a specific node holds the deeper layers.
    \item \textit{Decentralized FedAvg with Momentum (DFedAvgM)}. It uses SGD with momentum to train the local models of the federation participants, communicating only with their neighbors in an undirected graph. It provides a better convergence rate and communication efficiency than DSGD \cite{Sun:DFedAvgM:2022}.
    \item \textit{DeceFL}. It focuses on neighbor communications to improve efficiency for both convex and non-convex loss functions. It applies to a wide range of real-world medical and industrial applications \cite{Yuan:decefl_framework:2021}.
\end{itemize}

\subsubsection{Optimization of Federation Communications}

Optimizing network flexibility and overhead is crucial for addressing performance issues caused by continuous data exchanges between participants. Concerning \textit{network flexibility}, it is necessary to have techniques that reduce the number of exchanges in the network without losing functionality during federation \cite{Minsker:dfl_convergence:2017}. Several studies have proposed reducing the number of bits transferred for each worker update through data compression. However, these studies do not consider the impact of data loss from compression results, which could impair the learning accuracy and affect convergence \cite{Tang:communication_compression_efficiency:2018}.

In the literature, different distributed optimization schemes aim to maintain acceptable convergence rates in terms of recurring iterations and device computation time. Some examples are Decentralized Gradient Descent (DGD) \cite{Yuan:dgd:2016}, decentralized Alternating Direction Method of Multipliers (ADMM) \cite{Shi:ADMM:2014}, EXTRA \cite{Shi:EXTRA:2015}, and ADMM based on Jacobi-Proximal \cite{Deng:ADMM_plus:2017}. A recent contribution includes a method that applies the Lloyd-Max algorithm to DFL to minimize quantization distortion, with the resulting LM-DFL algorithm able to adjust quantization levels adaptively \cite{Liu:comms_opt:2023}. Other communication optimization mechanisms use an incremental learning method to reduce costs by activating and linking agents while keeping other nodes and links inactive. These include Random Walk ADMM (WADMM) \cite{Mao:walkman_randomwalk_optimization:2020}, Parallel Random Walk ADMM (PW-ADMM) \cite{Ye:randomwalk_optimization:2020} and Walk Proximal Gradient (WPG) \cite{Mao:walk_proximal_gradient:2019} which are commonly used in increment-based approaches. Recent research has introduced mechanisms that aim to optimize the balance between improving the quality of the model and saving communication resources, which can result in a more sustainable federation policy \cite{Barbieri:selection_optimizer:2023}. These mechanisms typically offer an independent selection of participants and fragments of the NN to be transmitted, providing a promising alternative to traditional optimizations. One such technique is FL-EOCD, leveraging D2D communications and overlapped clustering to enable decentralized aggregation, thereby reducing overall energy consumption and latency while maintaining convergence rates \cite{Al-Abiad:computation_opt:2023}. Further approaches, such as \cite{Nguyen:dfl_bl_latency_opt:2022}, implement latency optimization strategies by incorporating Blockchain. The latency optimization is achieved by strategically managing data offloading decisions, power transmission, bandwidth allocation, and computational resources, resulting in a highly responsive system.

Regarding optimizing \textit{network overhead}, the DFL approach presents limitations in communication bandwidth between nodes \cite{Jiang:bacombo_bandwidth_gossip:2020}. To address this issue, researchers have explored different distribution schemes, such as quantization and sparsification. For example, Quantized Sparse SGD (QSGD) \cite{Alistarh:qsgd_optimization:2017} and quantized ADMM \cite{Zhu:quantized_consensus_admm:2016} were proposed to reduce network overhead when exchanging model parameters. Furthermore, the Qsparse-local-SGD algorithm \cite{Basu:qsparse_local_sgd:2019} includes a combination of aggressive sparsification with quantization and local computation along with error compensation. However, these methods may decrease accuracy to achieve lower communication costs \cite{koloskova:choco_gossip_communication_compression:2019}. With adequate communication infrastructure, decentralized learning can achieve accuracies like its centralized counterpart. In this sense, the use of advanced technologies such as 5G \cite{Singh:bl_5g_dfl:2023} or 6G \cite{Ridhawi:6g_dfl_metaverse:2023} can help reduce communication limitations, improving network bandwidth and enable faster and more efficient data exchange. Other studies, such as \cite{Lian:algorithms_centralized_vs_decentralized:2017}, have shown that decentralized SGD can outperform centralized approaches in low-bandwidth networks. However, these studies do not consider typical federated settings, such as non-IID data among participants, limiting the applicability of their results.

\subsubsection{Optimization of Federation Models}

It is necessary to optimize federated models to avoid high processing demand on the most constrained participants. In this sense, optimizations are responsible for increasing the efficiency of the models without sacrificing performance and trustworthiness.

In DFL, nodes often aim to train a state-of-the-art ML model on a specified task. To increase \textit{model performance} and \textit{trustworthiness}, researchers have developed new models and improved existing solutions. The most popular ML model is a NN, which achieves state-of-the-art results in many tasks, such as image classification. Many DFL studies are based on Stochastic Gradient Descent (SGD), which can be used to train NNs \cite{Basu:qsparse_local_sgd:2019}. Another widely used algorithm is a decision tree, which is more efficient to train and easier to interpret than NNs. In particular, a tree-based DFL is designed for federated training of single or multiple decision trees, such as Gradient Boosting Decision Trees (GBDTs) and Random Forests (RFs). GBDTs have been especially popular recently due to their superior performance in many classification and regression tasks \cite{Mothukuri:survey_topology_architecture_privacy_decentralized:2021}. Besides NNs and trees, linear models such as Linear Regression (LR), Logistic Regression, or Support Vector Machine (SVM) are classic and easy-to-use models. To enhance the consistency of models generated by each federation participant, algorithms such as DFedSAM utilize gradient perturbation to produce local flat models with uniformly low loss values \cite{Shi:improving_model_dfl:2023}. In non-IID scenarios where adaptive training models can be challenging on resource-constrained devices, these models can be particularly useful \cite{Wu:network_gradient:2022}. Furthermore, techniques such as SwarmSGD jointly leverage non-blocking communication, quantization, and local steps, allowing for better compensation by training on more samples but with a lower quantization level of model parameters. This technique is effective in heterogeneous node data distributions and random topologies \cite{Nadiradze:quantized_models:2021}.

In addition to the efficient models mentioned above, new techniques have emerged in recent years to improve the efficiency and performance of DFL. These techniques include meta-learning, multitasking, and federated distillation. Meta-learning aims to improve the efficiency and performance of the model generated at each node by creating less complex models capable of performing tasks in parallel. It uses information from the activities performed to better adapt to new tasks \cite{Li:dfl_metalearning_multitask:2022}. As a result, meta-learners generalize better when trained with a more extensive set of practical tasks and more data for each task. In this sense, the learner can acquire knowledge and processing capacity distributed among agents, adapting its model to the best-performing ones in the network. Meta-learning often involves multitasking when ML models are trained to perform multiple tasks simultaneously. These models optimize metrics for each task and generate a corresponding output \cite{Vanhaesebrouck:multitask_collaboration_models:2016}. Finally, federated distillation, first introduced in 2015 as knowledge distillation, aims to optimize federated models by performing model compression. Specifically, it is a procedure to train a model using a second NN as a reference instead of learning directly from the ground truth of the data \cite{Hinton:first_knowledge_destillation:2015}. In this way, it allows several advantages over more complex models: (i) more information can be extracted from a single sample, (ii) training can be performed with fewer examples, and (iii) labeled data are not needed. All this offers less computation time and energy consumption in the federated node \cite{Anil:google_knowdlege_destillation:2018}. A decentralized version of this approach has been proposed in \cite{Seo:federated_knowledge:2020}, where instead of having a pre-trained model to imitate it, each model considers the aggregated knowledge of the remaining participants' models.

Finally, the most relevant optimization of federated models for increasing \textit{trustworthiness} is using DLT techniques, providing reliability during model parameter updates in the federation. They also facilitate the fair participation of nodes during the federation. In such trust-based relationships, there are two entities: the trustor and trustee node. For each observation from the trustee, the trustor determines whether it is an outlier, while the trustee node validates the contributions provided by the trustor. To maintain the reputation of the participating devices and ensure that all participants make active and honest contributions to the model, the trustor node $i$ can generate a smart contract with the parameters of the local model, which is validated by a set of trustee nodes $j$ \cite{Rehman:trustfd_trusted:2021}. DLT techniques also provide an added explanation and robustness to federated models. The immutability of the initialization and update of federated models is guaranteed by DLT, maintaining a chronological log in the Blockchain. Moreover, the fair participation of nodes in the federation is further ensured by DLT.
\section{Open-Source Frameworks for DFL}
\label{sec:frameworks}

Building upon the storyline presented in \figurename~\ref{fig:storyline}, this section responds to ``\textit{Q2. What DFL frameworks exist, and what fundamentals do they provide?}'' In this context, frameworks serve as the bridge between the ``Theoretical Overview'' with its fundamentals and ``Practical Application'' with its application scenarios. This section introduces the most relevant existing open-source frameworks for DFL. First, it discusses the most mature and highly customized frameworks supported by large companies. Then, it focuses on those more incipient with limited functionality but notable research advances. \tablename~\ref{tab:frameworks} compares the main aspects of the frameworks analyzed. The double horizontal line separates the mature frameworks from the incipient ones.


\begin{table*}[ht!]
\caption{Comparison of mature and incipient DFL frameworks.} \label{tab:frameworks}
\resizebox{\textwidth}{!}{
\centering
\begin{threeparttable}
\begin{tabular}{cccccccccc} 
\hline
\multirow{2}{*}{Reference} & \multirow{2}{*}{\makecell[t]{OS}} & \multicolumn{2}{c}{\makecell{Federation Architecture}} &
\multirow{2}{*}{\makecell[t]{Aggregation\\Algorithms}} & \multirow{2}{*}{\makecell[t]{Communication\\Protocol}} &
\multirow{2}{*}{\makecell[t]{Security and\\Privacy}} & \multirow{2}{*}{\makecell[t]{Data Type}} & \multirow{2}{*}{\makecell[t]{Scenario}} & \multirow{2}{*}{\makecell{Benchmarking}}
\\
& & \makecell[t]{Participant Type} & \makecell[t]{Aggregator Node} & & & & & & \\
\hline \hline

TFF \cite{google:tff:2022} & \makecell[t]{Linux\\MacOS} & \makecell[t]{Cross-silo} & \makecell[t]{Centralized\\Decentralized} & \makecell[t]{Median\\FedAvg\\FedProx} & \makecell[t]{gRPC} & \makecell[t]{\yestick} & \makecell[t]{Time series\\Images} & \makecell[t]{Simulation} & \makecell[t]{\yestick} 
\\ \hline

PySyft \cite{openminded:pysyft:2021} & \makecell[t]{Windows\\Linux\\MacOS\\Mobile} & \makecell[t]{Cross-silo\\Cross-device} & \makecell[t]{Centralized\\Decentralized} & \makecell[t]{FedAvg} & \makecell[t]{Websockets} & \makecell[t]{\yestick} & \makecell[t]{Images} & \makecell[t]{Simulation\\Real} & \makecell[t]{\yestick} 
\\ \hline

SecureBoost \cite{Cheng:secureboost:2019} & \makecell[t]{Linux\\MacOS} & \makecell[t]{Cross-silo} & \makecell[t]{Centralized\\Decentralized} & \makecell[t]{FedAvg\\GBDT} & \makecell[t]{gRPC} & \makecell[t]{\yestick} & \makecell[t]{Time series} & \makecell[t]{Simulation} & \makecell[t]{\notick} 
\\ \hline

FederatedScope \cite{Xie:federatedscope:2022} & \makecell[t]{Windows\\Linux\\MacOS\\Mobile} & \makecell[t]{Cross-silo\\Cross-device} & \makecell[t]{Centralized} & \makecell[t]{FedAvg\\FedOpt} & \makecell[t]{gRPC} & \makecell[t]{\yestick} & \makecell[t]{Time series\\Images} & \makecell[t]{Simulation\\Real} & \makecell[t]{\yestick} 
\\ \hline

FedML \cite{He:fedml:2020} & \makecell[t]{Linux\\MacOS} & \makecell[t]{Cross-silo} & \makecell[t]{Centralized\\Decentralized} & \makecell[t]{FedAvg\\FedOpt\\FedNova} & \makecell[t]{gRPC\\MPI\\MQTT} & \makecell[t]{\yestick} & \makecell[t]{Time series\\Images} & \makecell[t]{Simulation\\Real} & \makecell[t]{\yestick} 
\\ \hline

LEAF \cite{Caldas:leaf:2018} & \makecell[t]{Linux\\MacOS} & \makecell[t]{Cross-silo} & \makecell[t]{Centralized\\Decentralized} & \makecell[t]{-} & \makecell[t]{-} & \makecell[t]{-} & \makecell[t]{Time series\\Images} & \makecell[t]{Simulation} & \makecell[t]{\yestick} 
\\ \hline \hline

BrainTorrent \cite{Roy:dfl_braintorrent:2019} & \makecell[t]{Windows\\Linux\\MacOS} & \makecell[t]{Cross-device} & \makecell[t]{Decentralized} & \makecell[t]{FedAvg} & \makecell[t]{\textit{N/S}} & \makecell[t]{\notick} & \makecell[t]{Time series} & \makecell[t]{Simulation\\Real} & \makecell[t]{\notick} 
\\ \hline

Scatterbrained \cite{Wilt:scatterbrained_framework:2021} & \makecell[t]{Windows\\Linux\\MacOS} & \makecell[t]{Cross-device} & \makecell[t]{Centralized\\Decentralized} & \makecell[t]{FedAvg} & \makecell[t]{ZeroMQ} & \makecell[t]{\notick} & \makecell[t]{Time series\\Images} & \makecell[t]{Simulation} & \makecell[t]{\notick} 
\\ \hline

IPLS \cite{Pappas:IPLS_framework:2021} & \makecell[t]{Windows\\Linux\\MacOS} & \makecell[t]{Cross-device} & \makecell[t]{Decentralized} & \makecell[t]{FedAvg} & \makecell[t]{P2P} & \makecell[t]{\yestick} & \makecell[t]{Time series\\Images} & \makecell[t]{Simulation\\Real} & \makecell[t]{\yestick} 
\\ \hline

TrustFed \cite{Rehman:trustfd_trusted:2021} & \makecell[t]{Windows\\Linux\\MacOS} & \makecell[t]{Cross-device} & \makecell[t]{Centralized\\Decentralized} & \makecell[t]{FedAvg} & \makecell[t]{P2P} & \makecell[t]{\yestick} & \makecell[t]{Time series\\Images} & \makecell[t]{Simulation} & \makecell[t]{\notick} 
\\ \hline

FLoBC \cite{Ghanem:bl_dfl:2021} & \makecell[t]{Windows\\Linux\\MacOS} & \makecell[t]{Cross-device} & \makecell[t]{Decentralized} & \makecell[t]{FedAvg} & \makecell[t]{HTTP (REST API)} & \makecell[t]{\notick} & \makecell[t]{Time series\\Images} & \makecell[t]{Simulation\\Real} & \makecell[t]{\yestick} 
\\ \hline

BLADE-FL \cite{Li:blade_bl:2022} & \makecell[t]{Windows\\Linux} & \makecell[t]{Cross-device} & \makecell[t]{Decentralized} & \makecell[t]{Custom\\algorithm} & \makecell[t]{P2P} & \makecell[t]{\yestick} & \makecell[t]{Images} & \makecell[t]{Simulation} & \makecell[t]{\yestick} 
\\ \hline

DISCO \cite{epfl:disco_framework:2022} & \makecell[t]{Windows\\Linux\\MacOS\\Mobile} & \makecell[t]{Cross-device} & \makecell[t]{Centralized\\Decentralized} & \makecell[t]{Custom\\FedAvg} & \makecell[t]{peer.js} & \makecell[t]{\yestick} & \makecell[t]{Time series\\Images} & \makecell[t]{Simulation} & \makecell[t]{\notick} 
\\ \hline

CMFL \cite{Che:CMFL:2022} & \makecell[t]{Windows\\Linux\\MacOS} & \makecell[t]{Cross-device} & \makecell[t]{Decentralized} & \makecell[t]{Median\\Trimmed Mean\\Krum\\Multi-Krum} & \makecell[t]{P2P} & \makecell[t]{\yestick} & \makecell[t]{Time series\\Images} & \makecell[t]{Simulation} & \makecell[t]{\yestick}
\\ \hline

DeFL \cite{Han:dfl_cross_silo:2022} & \makecell[t]{Windows\\Linux\\MacOS} & \makecell[t]{Cross-device} & \makecell[t]{Decentralized} & \makecell[t]{Custom\\algorithm} & \makecell[t]{\textit{N/S}} & \makecell[t]{\notick} & \makecell[t]{Time series\\Images} & \makecell[t]{Simulation\\Real} & \makecell[t]{\yestick} 
\\ \hline

FL-SEC \cite{Qu:privacy_framework_iot:2022} & \makecell[t]{Windows\\Linux\\MacOS} & \makecell[t]{Cross-device} & \makecell[t]{Decentralized} & \makecell[t]{FedAvg} & \makecell[t]{\textit{N/S}} & \makecell[t]{\yestick} & \makecell[t]{Time series} & \makecell[t]{Simulation} & \makecell[t]{\yestick} 
\\ \hline

DisPFL\cite{Dai:dispfl_framework:2022} & \makecell[t]{Windows\\Linux\\MacOS} & \makecell[t]{Cross-device} & \makecell[t]{Decentralized} & \makecell[t]{FedAvg\\Ditto\\FOMO\\Sub-FedAvg} & \makecell[t]{P2P} & \makecell[t]{\notick} & \makecell[t]{Time series\\Images} & \makecell[t]{Simulation\\Real} & \makecell[t]{\notick} 
\\ \hline

GossipFL\cite{Tang:gossipfl_framework:2023} & \makecell[t]{Windows\\Linux\\MacOS} & \makecell[t]{Cross-device} & \makecell[t]{Decentralized} & \makecell[t]{FedAvg\\S-FedAvg\\D-PSGD\\CHOCO-SGD} & \makecell[t]{P2P} & \makecell[t]{\notick} & \makecell[t]{Images} & \makecell[t]{Simulation} & \makecell[t]{\yestick} 
\\ \hline

Fedstellar\cite{MartinezBeltran:fedstellar:2023} & \makecell[t]{Windows\\Linux\\MacOS} & \makecell[t]{Cross-silo\\Cross-device} & \makecell[t]{Centralized\\Decentralized\\Semi-Decentralized} & \makecell[t]{FedAvg\\Krum\\TrimmedMean\\Median} & \makecell[t]{P2P\\HTTP (REST API)} & \makecell[t]{\yestick} & \makecell[t]{Time series\\Images} & \makecell[t]{Simulation\\Real} & \makecell[t]{\yestick}

\\ \hline \hline

\end{tabular}
\begin{tablenotes}
\item \textit{N/S} (Not Specified) by the authors. The double horizontal line indicates the separation between mature and incipient frameworks
\end{tablenotes}
\end{threeparttable}}
\end{table*}

\subsection{Mature Frameworks}

TensorFlow Federated (TFF) \cite{google:tff:2022}, developed by Google, provides the building blocks for FL based on TensorFlow. It includes three key components: models, federated computation builders, and datasets. TFF can be deployed on multiple machines, creating a rotating aggregator. Currently, TFF does not consider any adversaries during FL training nor provides privacy mechanisms. Other frameworks, such as PySyft \cite{openminded:pysyft:2021}, provide interfaces for developers to implement their training algorithm. While TFF is based on TensorFlow, PySyft can work well with PyTorch and TensorFlow. This framework provides multiple optional privacy mechanisms, including secure multi-party computation and differential privacy. Moreover, it can be deployed on a single machine or multiple machines, where the communication between nodes is performed using the WebSocket API. For privacy preservation, SecureBoost \cite{Cheng:secureboost:2019} builds multi-part reinforcement trees with an encryption strategy between different nodes. This solution allows for different feature sets corresponding to a vertically partitioned dataset. One of the advantages of SecureBoost is that it provides remarkable accuracy across different topology architectures without revealing information about each private data provider.

Other frameworks, such as FederatedScope \cite{Xie:federatedscope:2022} and FedML \cite{He:fedml:2020}, allow lower-level management of the deployed architecture. The first one employs an event-driven architecture to provide great flexibility in describing node behaviors. It allows learning objectives and backends to be added and coordinated in an FL life cycle with synchronous or asynchronous training strategies. In addition, it includes privacy protection, attack simulation, and automatic tuning. In addition, FedML offers a platform for FL benchmarking based on PyTorch and utilizes the Message Passing Interface (MPI) protocol, gRPC, or MQTT for communication. This framework comprises two main modules: FedML-core, which implements the training engine and the distributed communication infrastructures, and FedML-API, which is built on top of FedML-core and provides training models, datasets, and FL algorithms. Finally, FedML supports three computing paradigms: standalone simulation, distributed computing, and on-device training. Along the same lines, LEAF \cite{Caldas:leaf:2018} provides only benchmark functionality of the deployed architectures. In this sense, it uses distributed datasets and partitioning mechanisms that other frameworks can leverage to complement its functionality.

\subsection{Incipient Solutions}

BrainTorrent \cite{Roy:dfl_braintorrent:2019} is an FL framework without a central server explicitly aimed at medical applications. With no need for a central authority to oversee the training process, BrainTorrent provides highly dynamic P2P communications where each medical center communicates directly, also offering robust training of participants through asynchronous updates achieving performance similar to on-device model training. Although the framework focuses on image segmentation, it can be extended to other data and models. Similarly, Scatterbrained \cite{Wilt:scatterbrained_framework:2021} facilitates the construction of both CFL and DFL systems. In the latter, it can create a customized communication system based on ZeroMQ, corresponding to a communications library based on UDP. Nodes can have a specific role: (i) leech, in which nodes listen for broadcasts but do not share data; (ii) offline, where nodes do not listen or broadcast data; (iii) peer, where participants listen and broadcast data; and (iv) seeding, where nodes broadcast but do not listen to incoming data. In addition, it allows the use of different ML frameworks such as TensorFlow, Scikit-learn, or PyTorch.

Other studies present DFL models based on the InterPlanetary File System (IPFS) \cite{Pappas:IPLS_framework:2021}. In this sense, the authors deployed a framework called IPLS, whose nodes are interconnected and perform federated model training based on the operation of one or more network participants. Any participant can initiate the federated training process or join an existing one. This solution shows robustness in the face of dynamic nodes and intermittent connectivity, ensuring limited resource usage and convergence of the federated model. Similarly, TrustFed \cite{Rehman:trustfd_trusted:2021} uses IPFS and Blockchain smart contracts to maintain the reputation of participants in a centralized or decentralized network. In this way, compelling participants to make active and honest contributions to the federated model. In contrast, other frameworks such as FLoBC \cite{Ghanem:bl_dfl:2021} deploy a Blockchain mechanism to ensure the decentralization of nodes. In this sense, the framework compares and contrasts the effects of the trainer-validator relationship, reward-penalty policy, and model synchronization schemes on overall system performance. The authors demonstrate that using this framework enables the deployment of DFL systems with performance similar to more centralized architectures. Similarly, BLADE-FL \cite{Li:blade_bl:2022} uses Blockchain to mitigate vulnerabilities during federation. It offers robustness against malfunctions, distrustful connections, and external threats by having each client compete to generate a block before local training in the next federation round. The system also addresses training deficiency caused by lazy clients and optimizes resource allocation.

In addition to the above solutions, some frameworks explore new applicability technologies to generate DFL scenarios. DISCO \cite{epfl:disco_framework:2022} is a framework presented by the EPFL Machine Learning and Optimization Laboratory that creates an easy-to-use platform that allows non-specialists to participate in collaborative learning. This platform is based on Javascript, supporting arbitrary DL tasks and model architectures via TF.js and relying on P2P communications. Additionally, DISCO provides mechanisms to ensure communication efficiency, privacy preservation, fault tolerance, and deployment customization. Moreover, frameworks seek to improve the robustness of their privacy techniques. In this case, \cite{Che:CMFL:2022} presented a serverless framework named Committee Mechanism-based Federated Learning (CMFL), orchestrated by a committee system deploying selection strategies of model parameters during federation. In contrast to the other frameworks, this one implements several aggregation algorithms such as median, trimmed mean, Krum, and Multi-Krum, evaluating them in terms of robustness and efficiency. Regarding the security and privacy of the scenarios, \cite{Qu:privacy_framework_iot:2022} and \cite{epfl:disco_framework:2022} provide frameworks capable of ensuring the integrity and confidentiality of the communications and data exchanged. Both alternatives present custom aggregation algorithms to process image and time series data.

Finally, other frameworks in the literature allow the deployment of cross-silo-based solutions, creating robust topologies of nodes that exchange information using P2P networks. In this case, \cite{Han:dfl_cross_silo:2022} removed the central server by aggregating the weights at each participating node, and the weights of only the current training round are maintained and synchronized among all nodes. This framework has been evaluated on two widely adopted public datasets, CIFAR-10 \cite{krizhevsky:cifar10-100:2009}, and Sentiment140 \cite{Go:sentiment140dataset:2009}, demonstrating convergence against more common threats with minimal accuracy loss, achieving up to 100x reduction in storage overhead and up to 12x reduction in network overhead compared to state-of-the-art DFL approaches. Similar to the previous solution, the authors of \cite{Dai:dispfl_framework:2022} allow further savings in communication and computational costs. In this sense, the authors proposed a decentralized sparse training technique, where each local model only maintains a fixed number of active parameters throughout the local training and P2P communication process. They employ aggregation algorithms such as Ditto, FOMO, and Sub-FedAvg to enhance efficiency. Furthermore, the authors demonstrated that the framework allows easy adaptation to heterogeneous local clients with different computational complexities. Similarly, GossipFL \cite{Tang:gossipfl_framework:2023} presents a novel approach to address communication challenges in IoT scenarios characterized by intermittent connections between participants. The framework utilizes sparsification techniques and a unique gossip matrix generation method, employing algorithms such as S-FedAvg, D-PSGD, and CHOCO-SGD to optimize communication traffic while preserving convergence. The efficacy is demonstrated across various datasets, such as MNIST and CIFAR10, achieving model accuracies of up to 98\%. Building on these advancements, Fedstellar \cite{MartinezBeltran:fedstellar:2023} offers a versatile platform for DFL training that addresses common limitations in existing frameworks. It efficiently manages heterogeneous topologies and adapts the federation to physical or virtual deployments. In a real-world cyberattack detection employing single-board devices and a fully connected network, Fedstellar obtained a 91\% F1 score. Furthermore, in a virtualized scenario using the MNIST dataset, the platform delivered a compelling F1 score of 98\% with DFL and 97.3\% with SDFL while reducing model convergence training time by 32\% relative to CFL.

\section{Application scenarios}
\label{sec:scenarios}

Following the storyline presented in \figurename~\ref{fig:storyline}, this section responds to ``\textit{Q3. Which are the main characteristics of the most relevant scenarios of DFL?}'' In particular, this section provides a review of existing works classified according to the application scenario: (i) healthcare, (ii) Industry 4.0, (iii) mobile services, (iv) military, and (v) vehicles. These scenarios combine most of the literature work, thus giving a great overview of current DFL applicability. For each scenario, the following aspects are highlighted: the goal, the participants during federation, the ML algorithm used, and the aggregation method employed.

Beyond the previously discussed scenarios, DFL is also relevant in emerging areas such as smart cities and metaverse, where it enhances urban planning and service provision, all while facilitating privacy-preserving data sharing between real and virtual entities \cite{Gurung:metaverse_dfl:2023}. In the realm of swarm robotics, DFL enables asynchronous knowledge sharing and task coordination \cite{Zhou:swarm_robots:2023}. In the energy and utilities sector, DFL supports predictive maintenance and demand forecasting, contributing to operational efficiency across companies \cite{Liu:pv_stations:2022}. Finally, the telecommunications sector benefits from DFL in optimizing network performance and anomaly detection, allowing network providers to share insights without compromising proprietary information \cite{Nakip:intrusion_dfl:2023}. However, the challenges of these domains need to be further explored to determine the scope of DFL.

\subsection{Healthcare}

Cooperation between research institutes, hospitals, and federal agencies is essential in modern healthcare systems to improve healthcare standards. However, transferring patient information between these entities can be challenging due to regulations like the General Data Protection Regulation (GDPR) or the European Health Data Space (EHDS). Through DFL, each hospital or institute only needs to share the parameters of the local model to obtain an accurate diagnosis model, thereby preserving patients' privacy. Notably, the successful application of DFL-based approaches in the healthcare sector is evident from recent studies, as demonstrated in the relevant articles presented in \tablename~\ref{tab:dfl_healthcare}.

\begin{table*}[ht!]
\caption{Comparison of DFL solutions focused on healthcare.} \label{tab:dfl_healthcare}
\resizebox{\textwidth}{!}{
\centering
\begin{threeparttable}
\begin{tabular}{ccp{5cm}cccccccc}
\hline
Ref. & 
 Year & 
 \makecell[t]{Goal} & 
 \makecell[t]{Federation\\Architecture} & 
 \makecell[t]{Topology} & 
 \makecell[t]{Scenario} & 
 \makecell[t]{ML\\Model} & 
 \makecell[t]{Aggregation\\Algorithm} & 
 \makecell[t]{Local Model\\Aggregation} &
 \makecell[t]{Client\\Selection} &
 \makecell[t]{Dataset} \\ 
\hline \hline

\cite{Liu:fadl_distributed:2018} & 2018 & Improve the accuracy of the CFL approximation by training parts of the model in a decentralized way and others using specific data sources & \makecell[t]{Cross-silo} & \makecell[t]{Centralized\\Decentralized} & \makecell[t]{Simulation} & \makecell[t]{Three-layered NN} & \makecell[t]{Custom\\algorithm} & \makecell[t]{Synchronous} & \makecell[t]{Sequential} & \makecell[t]{eICU-CRD \cite{Pollard:eICU:2018}} \\ \hline

\cite{Huang:fl_healthcare_mobility:2019} & 2019 & Decrease the cost of communications between hospitals and the aggregator by using a clinical community data clustering algorithm for disease detection & \makecell[t]{Cross-silo} & \makecell[t]{Centralized\\Decentralized} & \makecell[t]{Simulation} & \makecell[t]{K-Means} & \makecell[t]{FedAvg} & \makecell[t]{Synchronous} & \makecell[t]{Sequential} & \makecell[t]{eICU-CRD \cite{Pollard:eICU:2018}} \\ \hline

\cite{Lu:healthcare_records:2020} & 2020 & Improve the communication efficiency for fully DFL over a graph without loss of optimality of the solutions & \makecell[t]{Cross-device} & \makecell[t]{Decentralized} & \makecell[t]{Real} & \makecell[t]{Custom\\model} & \makecell[t]{Custom\\algorithm} & \makecell[t]{Synchronous\\Asynchronous} & \makecell[t]{Sequential} & \makecell[t]{Private} \\ \hline

\cite{Elayan:iot_healthcare:2021} & 2021 & Deep FL framework for decentralized healthcare systems that maintain user privacy & \makecell[t]{Cross-device} & \makecell[t]{Decentralized} & \makecell[t]{Simulation} & \makecell[t]{Custom\\model} & \makecell[t]{Custom\\algorithm} & \makecell[t]{Synchronous} & \makecell[t]{Scheduling} & \makecell[t]{Private} \\ \hline

\cite{Warnat-Herresthal:swarm_healthcare:2021} & 2021 & Facilitate the integration of any medical data from any data owner worldwide without violating privacy laws & \makecell[t]{Cross-silo} & \makecell[t]{Decentralized} & \makecell[t]{Real} & \makecell[t]{Custom\\model} & \makecell[t]{Custom\\aggregator} & \makecell[t]{Asynchronous} & \makecell[t]{Scheduling} & \makecell[t]{PBMC transcriptome \cite{ncbi:geo:2016}\\Chest X-ray \cite{Wang:chestxray:2019}} \\ \hline

\cite{Salim:healthcare:2022} &
2022 &
Secure valuable hospital biomedical data useful for clinical research organizations, use of Blockchain and smart contracts & \makecell[t]{Cross-device} & \makecell[t]{Decentralized} & \makecell[t]{Simulation} & \makecell[t]{CNN} & \makecell[t]{FedAVG} & \makecell[t]{Asynchronous} & \makecell[t]{Sequential} & \makecell[t]{Private} \\ \hline

\cite{Nguyen:healthcare_nature:2022} & 2022 & Privacy and protection of medical data exchanged between healthcare entities & \makecell[t]{Cross-silo} & \makecell[t]{Decentralized} & \makecell[t]{Real} & \makecell[t]{ResNet18\\ResNet50\\DenseNet121} & \makecell[t]{Custom\\algorithm} & \makecell[t]{Synchronous} & \makecell[t]{Sequential} & \makecell[t]{
ImageNet \cite{Deng:imagenet:2009}\\Private} \\ \hline

\cite{Tedeschini:healthcare_tumor:2022} & 2022 & Detect brain tumor segmentation, using representative clinical datasets & \makecell[t]{Cross-device} & \makecell[t]{Centralized\\Decentralized} & \makecell[t]{Real} & \makecell[t]{U-Net} & \makecell[t]{CFA\\FedPer} & \makecell[t]{Synchronous\\Asynchronous} & \makecell[t]{Random} & \makecell[t]{BRATS \cite{Menze:brats:2015}\\Private} \\ \hline

\cite{Kuo:healthcare_model_misconducts:2022} & 2022 & Detect misconduct model on specific sites from any learning iteration & \makecell[t]{Cross-device} & \makecell[t]{Centralized\\Decentralized} & \makecell[t]{Simulation} & \makecell[t]{GloreChain} & \makecell[t]{FedAvg} & \makecell[t]{Asynchronous} & \makecell[t]{Sequential} & \makecell[t]{Cancer Biomarker \cite{Zou:Cancerbiomarkerdataset:2011}} \\ \hline

\cite{Lian:efficient_privacy_healthcare:2022} & 2022 & Collaboratively training between healthcare devices for creating a robust model without raw data exchange & \makecell[t]{Cross-device} & \makecell[t]{Decentralized} & \makecell[t]{Simulation} & \makecell[t]{3D CNN} & \makecell[t]{Custom FedAvg\\(RingAVG)} & \makecell[t]{Synchronous\\Asynchronous} & \makecell[t]{Random} & \makecell[t]{COVID-19 CT \cite{Yang:covidctdataset:2020}\\Eye Disease \cite{Dos:eyediseasedataset:2021}\\Skin Cancer \cite{Jennings:skincancerdataset:2021}} \\ \hline

\cite{Wang:ring_topology_healthcare:2022} & 2022 & Ring topology for DGM to ensure security and privacy in medical scenarios & \makecell[t]{Cross-device} & \makecell[t]{Decentralized} & \makecell[t]{Simulation} & \makecell[t]{GAN} & \makecell[t]{FedAvg} & \makecell[t]{Asynchronous} & \makecell[t]{Sequential} & \makecell[t]{MNIST \cite{Deng:MNIST:2012}} \\ \hline

\cite{Pennisi:healthcate_privacy:2022} & 2022 & Exploit concepts from experience repetition and adversarial generation research & \makecell[t]{Cross-device} & \makecell[t]{Centralized\\Decentralized} & \makecell[t]{Simulation} & \makecell[t]{GAN} & \makecell[t]{FedAvg\\FedProx\\FedBN} & \makecell[t]{Synchronous\\Asynchronous} & \makecell[t]{Random} & \makecell[t]{X-ray data \cite{Candemir:xraydataset1:2014, Jaeger:xraydataset2:2014}\\HAM10000 \cite{Tschandl:ham10000:2018}} \\ \hline

\cite{Tian:healthcare_dfl:2023} &
2023 & Privacy-preserving decentralized approach while supporting efficient communications & \makecell[t]{Cross-silo} & \makecell[t]{Decentralized} & \makecell[t]{Simulation} & \makecell[t]{\textit{N/S}} & \makecell[t]{Custom\\algorithm} & \makecell[t]{Synchronous} & \makecell[t]{Random} & \makecell[t]{MNIST \cite{Deng:MNIST:2012}} \\ \hline \hline

\end{tabular}
\end{threeparttable}}
\end{table*}

The first study to present the privacy issues of server use in healthcare is \cite{Liu:fadl_distributed:2018}. The authors noted the challenges in achieving a trade-off between global learning of the model and local information from each data source due to the large number of data sources with varying amounts and properties of data. To address this, they proposed a new strategy called Federated-Autonomous Deep Learning (FADL), where the first layer of the NN model is trained in a federated way using data from all sources. In contrast, the other model layers are trained locally in each data source. Their approach demonstrated an accuracy similar to the centralized analysis and outperformed regular FL for distributed electronic health records.

In the same way as the previous work, Huang \textit{et al.} \cite{Huang:fl_healthcare_mobility:2019} proposed a community-based FL algorithm to predict mortality and hospital stay time. Electronic medical records were clustered into communities inside each hospital based on standard medical aspects. Each cluster learned and shared a particular ML model customized for each community rather than a general global model shared among all hospitals. In this work, a participant with an aggregator role was in charge of converting patients' drug features into privacy-preserving representations. The authors of \cite{Lu:healthcare_records:2020} introduced an algorithm to perform local updates during several iterations, enabling communication between nodes and improving communication efficiency for DFL. This approach was tested on simulations with real-world electronic medical records, demonstrating its effectiveness in Alzheimer's disease detection compared to classical methods. Similarly, Elayan \textit{et al.} \cite{Elayan:iot_healthcare:2021} proposed a framework to train on skin lesion images using smartphones. They further utilized TFL to overcome the need for large, labeled data. They also proposed an automated training data-acquiring process and evaluated the algorithm on skin diseases, leveraging TL techniques to address the problem of a lack of healthcare data in generating DL models. Moreover, Warnat-Herresthal \textit{et al.} \cite{Warnat-Herresthal:swarm_healthcare:2021} followed a similar approach to ensure the privacy of medical data exchanged between cross-silos.

The advancement of clinical research and new scientific discoveries relies on enabling data mobility, including medical records. In this context, various schemes have been proposed to exchange medical records using Blockchain and smart contracts to ensure appropriate data privacy and protection. Salim and Park \cite{Salim:healthcare:2022} compared the results of their decentralized Convolutional Neural Network (CNN) model for Electronic Health Records (EHR) with a private Interplanetary File System (IPFS) and found promising results in accuracy, sensitivity, and specificity, similar to the traditional centralized model. In contrast, Nguyen \textit{et al.} \cite{Nguyen:healthcare_nature:2022} discussed the use of federated distillation to decrease model complexity, using a private dataset called The Noisy Blind consisting of 1198 original images collected from four clinics. Their solution provided superior performance to traditional centralized training when nodes comprise low-quality data, which is common in healthcare. Similarly, the authors of \cite{Tedeschini:healthcare_tumor:2022} focused on medical image prediction for brain tumor segmentation, enabling the collaboration of multiple institutions by sharing their local computational models and training a U-Net model. This approach reached a target Dice Similarity Coefficient (DSC) above 85\% using private data from patients' devices (e.g., heart rate, blood oxygen saturation).

\subsection{Industry 4.0}

The use of federated and decentralized techniques has increased considerably in Industry 4.0 to overcome the lack of communication between devices and sensors without needing a central node. As shown in \tablename~\ref{tab:dfl_industry}, several recent studies have shown promising results in applying DFL-based approaches in industrial settings.

\begin{table*}[ht!]
\caption{Comparison of DFL solutions focused on Industry 4.0.} \label{tab:dfl_industry}
\resizebox{\textwidth}{!}{
\centering
\begin{threeparttable}
\begin{tabular}{ccp{5cm}cccccccc}
\hline
Ref. & 
 Year & 
 \makecell[t]{Goal} & 
 \makecell[t]{Federation\\Architecture} & 
 \makecell[t]{Topology} & 
 \makecell[t]{Scenario} & 
 \makecell[t]{ML\\Model} & 
 \makecell[t]{Aggregation\\Algorithm} & 
 \makecell[t]{Local Model\\Aggregation} &
 \makecell[t]{Client\\Selection} &
 \makecell[t]{Dataset} \\  
\hline \hline

\cite{Savazzi:cooperative_fl_robots_drones:2021} &
2021 & Explore emerging FL opportunities using next-generation networked and autonomous industrial systems (e.g., robots, vehicles, drones) & \makecell[t]{Cross-device} & \makecell[t]{Centralized\\Decentralized} & \makecell[t]{Real} & \makecell[t]{DNN} & \makecell[t]{FedAvg} & \makecell[t]{Synchronous\\Asynchronous} & \makecell[t]{Random} & \makecell[t]{MIMO radar \cite{Savazzi:MIMOdataset:2020}} \\ \hline

\cite{Qu:bl_cognitive_computing_industry:2021} &
2021 & Decentralized paradigm for big data-based cognitive computing & \makecell[t]{Cross-device} & \makecell[t]{Decentralized} & \makecell[t]{Simulation} & \makecell[t]{CNN} & \makecell[t]{DANE} & \makecell[t]{Synchronous} & \makecell[t]{Random} & \makecell[t]{CIFAR-10 \cite{krizhevsky:cifar10-100:2009}} \\ \hline

\cite{Ma:iot_selection_quantized:2021} &
2021 & A method for D2D network in industrial IoT devices for improving the convergence of communications & \makecell[t]{Cross-device} & \makecell[t]{Decentralized} & \makecell[t]{Real} & \makecell[t]{MobileNet} & \makecell[t]{FedAvg\\(consensus)} & \makecell[t]{Asynchronous} & \makecell[t]{Random} & \makecell[t]{MNIST \cite{Deng:MNIST:2012}} \\ \hline

\cite{Wang:p2p_iot_cyber_anomaly_detection:2021} &
2021 & Anomaly detection ML models in non-IID scenarios, including mechanisms to locally rebalance training datasets & \makecell[t]{Cross-device} & \makecell[t]{Decentralized} & \makecell[t]{Simulation} & \makecell[t]{Four-layered FCNN} & \makecell[t]{P2PK-SMOTE} & \makecell[t]{Synchronous\\Asynchronous} & \makecell[t]{Random} & \makecell[t]{N-BaIoT \cite{Meidan:nbaiot:2018}} \\ \hline

\cite{Lian:iot_anomaly_detection:2022} &
2022 & A method for decentralized anomaly detection using neural networks, offering a comparison with traditional federated architectures & \makecell[t]{Cross-device} & \makecell[t]{Decentralized} & \makecell[t]{Simulation} & \makecell[t]{Custom\\model} & \makecell[t]{FedAvg} & \makecell[t]{Asynchronous} & \makecell[t]{Sequential} & \makecell[t]{IoT23 \cite{Garcia:IoT23:2020}} \\ \hline

\cite{Abdel-Basset:bl_edgeiot_cyber:2022} &
2022 & A BoEI framework that integrates an innovative DFL for cyberattack detection in IIoT & \makecell[t]{Cross-device} & \makecell[t]{Centralized\\Decentralized} & \makecell[t]{Simulation} & \makecell[t]{Custom\\model} & \makecell[t]{Fed-Trust} & \makecell[t]{Synchronous\\Asynchronous} & \makecell[t]{Random} & \makecell[t]{TON\_IoT \cite{Alsaedi:TONIOT:2020}\\LITNET-2020 \cite{Damasevicius:LITNET:2020}} \\ \hline

\cite{Mothukuri:trusted_blockchain:2022} &
2022 & Secure application approach to seal and sign asynchronous and synchronous FL collaborative tasks & \makecell[t]{Cross-silo\\Cross-device} & \makecell[t]{Decentralized} & \makecell[t]{Real} & \makecell[t]{Custom\\model} & \makecell[t]{Custom\\algorithm} & \makecell[t]{Asynchronous} & \makecell[t]{Sequential} & \makecell[t]{Private} \\ \hline

\cite{Pinyoanuntapong:trust_sync_async_edgeiot:2022} &
2022 & Generic decentralized FL framework for training distributed models in inherently heterogeneous IoT environments & \makecell[t]{Cross-device} & \makecell[t]{Decentralized} & \makecell[t]{Real} & \makecell[t]{Custom\\model} & \makecell[t]{Custom\\algorithm} & \makecell[t]{Synchronous\\Asynchronous} & \makecell[t]{Random} & \makecell[t]{Private} \\ \hline

\cite{Ochiai:adhoc_iot:2022} &
2022 & Cooperative machine learning organized by physically close nodes & \makecell[t]{Cross-device} & \makecell[t]{Decentralized} & \makecell[t]{Simulation} & \makecell[t]{Two-layered FCNN} & \makecell[t]{WAFL} & \makecell[t]{Synchronous\\Asynchronous} & \makecell[t]{Random} & \makecell[t]{MNIST \cite{Deng:MNIST:2012}} \\ \hline

\cite{Kang:bl_metaverse_industrial:2022} &
2022 & User-defined privacy-preserving framework for industrial metaverses & \makecell[t]{Cross-silo\\Cross-device} & \makecell[t]{Decentralized} & \makecell[t]{Simulation} & \makecell[t]{\textit{N/S}} & \makecell[t]{Custom\\FedAvg} & \makecell[t]{Asynchronous} & \makecell[t]{Scheduling} & \makecell[t]{Private} \\ \hline

\cite{Ranathunga:dfl_bl_industry:2023} &
2023 & A method that leverages Blockchain to improve model quality and reduce latency & \makecell[t]{Cross-silo} & \makecell[t]{Decentralized} & \makecell[t]{Simulation} & \makecell[t]{Two-layered MLP} & \makecell[t]{FedAvg} & \makecell[t]{Asynchronous} & \makecell[t]{Random} & \makecell[t]{MNIST \cite{Deng:MNIST:2012}} \\ \hline

\cite{Singh:bl_5g_dfl:2023} &
2023 & Cooperative use of Blockchain and FL while preserving privacy using 5G & \makecell[t]{Cross-silo} & \makecell[t]{Decentralized} & \makecell[t]{Simulation} & \makecell[t]{\textit{N/S}} & \makecell[t]{Custom\\algorithm} & \makecell[t]{Synchronous} & \makecell[t]{Random} & \makecell[t]{CIFAR-10 \cite{krizhevsky:cifar10-100:2009}\\FEMNIST \cite{Caldas:leaf:2018}} \\ \hline \hline

\end{tabular}
\begin{tablenotes}
\item \textit{N/S} (Not Specified) by the authors
\end{tablenotes}
\end{threeparttable}}
\end{table*}

In \cite{Savazzi:cooperative_fl_robots_drones:2021}, the authors discussed the opportunities and applications of DFL in automated and networked industrial systems supported by Device-to-Device (D2D) communications. The authors demonstrated the integration of the DFL approach into the sensing-decision-action loop, improving knowledge discovery operations. The system evaluation showed a convergence of the collaboration model in cross-device environments with minimal synchronous and asynchronous communications. Following the same direction, the authors of \cite{Qu:bl_cognitive_computing_industry:2021} applied cognitive computing in a simulated industrial scenario. This technique models the reasoning process of the human brain, having a progressive welcome in Industry 4.0 automation. This work implemented a DFL scenario supported by Blockchain and a Distributed Approximate Newton (DANE) aggregation method with advanced checks and random node selection. The results of an extensive evaluation and assessment showed accuracy values between 0.74 and 0.82. Ma, Wang, and Li \cite{Ma:iot_selection_quantized:2021} defined Q-DFL for a quantized selection of network nodes. The proposal contained two phases: (i) a local model was trained with the SGD algorithm on each IIoT device, and then the parameters of the quantized model were exchanged among neighboring nodes; and (ii) a consensus mechanism was designed to ensure that the local models converged to the same global model. Subsequent simulation for the MobileNet model revealed its performance trade-off between system information flow consumption, time delay, and energy cost.

Cybersecurity is also an important consideration in Industry 4.0, and several studies have investigated the use of DFL for anomaly and attack detection, such as the work published in \cite{Wang:p2p_iot_cyber_anomaly_detection:2021}. The authors developed anomaly detection ML models for non-IID scenarios and included mechanisms to locally rebalance the training datasets through the synthetic generation of data points from the minority class. Using different metrics to evaluate the model performance on the N-BaIoT dataset, they obtained a convergence close to 75\% of False Negative Rate (FNR) and recall. Another study, conducted by Lian and Su \cite{Lian:iot_anomaly_detection:2022}, discussed a similar approach with the IoT23 dataset \cite{Garcia:IoT23:2020}, obtaining a model convergence 30\% faster than in the previous study with an increase of only 3\% in the overhead of the participants.

Apart from ensuring the reliability of IIoT and the connectivity of industrial objects, Abdel-Basset, Moustafa, and Hawash \cite{Abdel-Basset:bl_edgeiot_cyber:2022} proposed a Blockchain-orchestrated Edge Intelligence (BoEI) framework for cyberattack detection in IIoT. The framework uses a temporal convolutional generative network to enable semi-supervised learning from semi-labeled data. The authors used TON\_IoT \cite{Alsaedi:TONIOT:2020} and LITNET-2020 \cite{Damasevicius:LITNET:2020} to validate the robustness and efficiency of the framework over different cyberattack detection approaches. Several studies have also addressed the synchronous or asynchronous operation of nodes within the network topology. In this regard, Pinyoanuntapong \textit{et al.} \cite{Pinyoanuntapong:trust_sync_async_edgeiot:2022} implemented a solution to mitigate high communication congestion in decentralized networks, leading to robust distributed model training in heterogeneous IoT environments where stragglers (i.e., slow devices) are commonplace due to varying computation and network speeds of IoT devices. Similarly, Ochiai \textit{et al.} \cite{Ochiai:adhoc_iot:2022} proposed a similar approach for ad-hoc wireless environments, achieving an average accuracy of 95\% for the test IID dataset compared to on-device training, which obtained 84.7\%. Mothukuri \textit{et al.} \cite{Mothukuri:trusted_blockchain:2022} proposed a Blockchain-based approach and Hyperledger Fabric with a gamification component for integrating a secure application to seal and sign asynchronous and synchronous DFL collaborative tasks. Their results demonstrated that security enhancements improve the FL process using a custom aggregation algorithm and asynchronous communications. Finally, Kang \textit{et al.} \cite{Kang:bl_metaverse_industrial:2022} enhanced privacy in the industrial metaverse by using a DFL framework with cross-chain power for decentralized, secure, and privacy-preserving data training in physical and virtual spaces. The framework uses a hierarchical Blockchain architecture with the main chain and multiple sub-chains. Ranathunga \textit{et al.} \cite{Ranathunga:dfl_bl_industry:2023} introduced a similar cross-silo architecture using a network of aggregators to optimize model quality, minimize convergence time, and improve system performance in various industrial applications. Similarly, Singh, Yang, and Park \cite{Singh:bl_5g_dfl:2023} proposed an approach for Industry 5.0. This scheme allowed local learning updates within the different departments of the industry using 5G. The validation outcomes showed an accuracy of 93.5\% in a 50\% active node.

\subsection{Mobile Services}

The proliferation of interconnected mobile devices and the development of advanced intelligent models have facilitated the emergence of decentralized approaches that offer increased efficiency and optimal training of models involved in various everyday tasks. To this end, DFL-based approaches have gained significant attention in mobile services, as demonstrated by the literature in \tablename~\ref{tab:dfl_mobile}.

\begin{table*}[ht!]
\caption{Comparison of DFL solutions focused on mobile services.} \label{tab:dfl_mobile}
\resizebox{\textwidth}{!}{
\centering
\begin{threeparttable}
\begin{tabular}{ccp{5cm}cccccccc} 
\hline
Ref. & 
 Year & 
 \makecell[t]{Goal} & 
 \makecell[t]{Federation\\Architecture} & 
 \makecell[t]{Topology} & 
 \makecell[t]{Scenario} & 
 \makecell[t]{ML\\Model} & 
 \makecell[t]{Aggregation\\Algorithm} & 
 \makecell[t]{Local Model\\Aggregation} &
 \makecell[t]{Client\\Selection} &
 \makecell[t]{Dataset} \\  
\hline \hline

\cite{Wang:edge_communication_optimization:2021} &
2021 & Reduce the number of directly connected end devices, avoiding the overhead of unnecessary local updates & \makecell[t]{Cross-silo\\Cross-device} & \makecell[t]{Centralized\\Decentralized} & \makecell[t]{Real} & \makecell[t]{CNN\\MLP} & \makecell[t]{FedAvg} & \makecell[t]{Synchronous\\Asynchronous} & \makecell[t]{Random} & \makecell[t]{MNIST \cite{Deng:MNIST:2012}} \\ \hline

\cite{Monschein:p2p_continous_authentication:2021} &
2021 & Novel approach based on continuous authentication of users trained by an organized peer-to-peer federation involving different organizations & \makecell[t]{Cross-device} & \makecell[t]{Centralized\\Decentralized} & \makecell[t]{Simulation} & \makecell[t]{Custom\\model} & \makecell[t]{FedAvg} & \makecell[t]{Asynchronous} & \makecell[t]{Scheduling} & \makecell[t]{Private} \\ \hline

\cite{Lu:blockchain_digitaltwin:2021} &
2021 & Blockchain-based FL scheme to strengthen communication security and data privacy protection in DITENs & \makecell[t]{Cross-silo\\Cross-device} & \makecell[t]{Centralized\\Decentralized} & \makecell[t]{Simulation} & \makecell[t]{DNN} & \makecell[t]{Custom\\algorithm} & \makecell[t]{Synchronous\\Asynchronous} & \makecell[t]{Random} & \makecell[t]{MNIST \cite{Deng:MNIST:2012}\\FMNIST \cite{Xiao:fmnist:2017}} \\ \hline

\cite{Wilhelmi:dfl_bl:2022} &
2022 & Simulation tool to capture the decentralized and asynchronous nature of FL operation in conjunction with DTL techniques & \makecell[t]{Cross-device} & \makecell[t]{Decentralized} & \makecell[t]{Simulation} & \makecell[t]{U-Net} & \makecell[t]{FedAvg} & \makecell[t]{Asynchronous} & \makecell[t]{Sequential} & \makecell[t]{BRATS \cite{Menze:brats:2015}\\Private} \\ \hline

\cite{Qi:dfl_bl:2022} &
2022 & Reputation scheme to balance the weights between trust values of parameters and bid prices & \makecell[t]{Cross-device} & \makecell[t]{Centralized\\Decentralized} & \makecell[t]{Simulation} & \makecell[t]{Four-layered FNN} & \makecell[t]{FedAvg} & \makecell[t]{Synchronous} & \makecell[t]{Scheduling} & \makecell[t]{MNIST \cite{Deng:MNIST:2012}} \\ \hline

\cite{Wang:dfl_edge:2022} &
2022 & Novel distributed validation weighting scheme to evaluate the performance of a mobile node in the federation versus a distributed validation set & \makecell[t]{Cross-device} & \makecell[t]{Centralized\\Decentralized} & \makecell[t]{Simulation} & \makecell[t]{Two-layered CNN} & \makecell[t]{Custom\\algorithm} & \makecell[t]{Synchronous\\Asynchronous} & \makecell[t]{Sequential} & \makecell[t]{CIFAR-10 \cite{krizhevsky:cifar10-100:2009}\\CIFAR-100 \cite{krizhevsky:cifar10-100:2009}\\EMNIST \cite{Cohen:emnist:2017}} \\ \hline

\cite{Belal:recommender_gossip:2022} &
2022 & Decentralized recommender system based on the principles of gossip learning & \makecell[t]{Cross-device} & \makecell[t]{Decentralized} & \makecell[t]{Simulation} & \makecell[t]{GMF\\PRME-G} & \makecell[t]{FedAvg\\FedFast\\Reptile} & \makecell[t]{Synchronous} & \makecell[t]{Scheduling} & \makecell[t]{Foursqueare-NYC \cite{Yang:foursquearedataset:2015}\\Gowalla-NYC \cite{Cho:Gowalladataset:2011}\\MovieLens \cite{Harper:movielens:2015}} \\ \hline

\cite{Khelghatdoust:socialnetworks_gossip:2022} &
2022 & Gossip-based system to maximize social awareness among nodes, improving reliability and latency & \makecell[t]{Cross-device} & \makecell[t]{Decentralized} & \makecell[t]{Simulation} & \makecell[t]{\textit{N/S}} & \makecell[t]{Custom\\algorithm} & \makecell[t]{Asynchronous} & \makecell[t]{Sequential} & \makecell[t]{Facebook \cite{Rozemberczki:facebook_dataset:2021}} \\ \hline

\cite{Pan:Lumos_FGL:2023} & 2023 & Federated GNN supporting supervised and unsupervised learning with security measures & \makecell[t]{Cross-device} & \makecell[t]{Centralized\\Decentralized} & \makecell[t]{Simulation} & \makecell[t]{GCN\\GAT} & \makecell[t]{Custom\\algorithm} & \makecell[t]{Synchronous} & \makecell[t]{Squential} & \makecell[t]{Facebook \cite{Rozemberczki:facebook_dataset:2021}\\LastFM \cite{Rozemberczki:LastFM:2020}} \\ \hline

\cite{Arapakis:p4l_p2p_telefonica:2023} & 2023 & Privacy-preserving P2P learning for asynchronous and collaborative tasks. & \makecell[t]{Cross-device} & \makecell[t]{Centralized\\Decentralized} & \makecell[t]{Simulation} & \makecell[t]{MobileNetv2\\BLSTM} & \makecell[t]{Custom\\algorithm} & \makecell[t]{Synchronous\\Asynchronous} & \makecell[t]{Random} & \makecell[t]{CIFAR-10 \cite{krizhevsky:cifar10-100:2009}\\Avito \cite{Avito:dataset:2021}\\IMDb \cite{Maas:IMDB_dataset:2011}} \\ \hline

\cite{Jeong:dfl_personalized_distillation:2023} & 2023 & Personalized and fully decentralized FL algorithm using knowledge distillation & \makecell[t]{Cross-device} & \makecell[t]{Decentralized} & \makecell[t]{Simulation} & \makecell[t]{Three-layered NN\\CNN} & \makecell[t]{Custom\\algorithm} & \makecell[t]{Synchronous\\Asynchronous} & \makecell[t]{Random} & \makecell[t]{IoT identification \cite{Kaggle:device_identification:2021}\\EMNIST \cite{Cohen:emnist:2017}} \\ \hline

\cite{Salama:dfl_mesh_topology:2023} & 2023 & Decentralized and privacy-preserving FL for improved performance and scalability & \makecell[t]{Cross-device} & \makecell[t]{Decentralized} & \makecell[t]{Simulation} & \makecell[t]{CNN} & \makecell[t]{FedAvg} & \makecell[t]{Synchronous} & \makecell[t]{Random} & \makecell[t]{MNIST \cite{Deng:MNIST:2012}} \\ \hline \hline

\end{tabular}
\begin{tablenotes}
\item \textit{N/S} (Not Specified) by the authors
\end{tablenotes}
\end{threeparttable}}
\end{table*}

Most research in this field has focused on leveraging edge computing to bring processing and data storage closer to the source of the request, thereby improving response times and conserving bandwidth on constrained devices. In this context, Wang \textit{et al.} \cite{Wang:edge_communication_optimization:2021} proposed a novel solution that deployed mobile edge nodes at various network locations to act as communication hubs between the cloud and end devices. This approach effectively sidesteps the latency associated with high server concurrency. Moreover, the proposed method filters unnecessary models and communications using cosine similarity. Experimental results showed that the proposed scheme reduces the number of local updates by 60\% compared to CFL and increases the convergence speed of the evaluated model by 10.3\%. Additionally, Monschein \textit{et al.} \cite{Monschein:p2p_continous_authentication:2021} addressed the challenge of acquiring large amounts of data to train powerful ML models, which is often an overwhelming task for a single organization. The authors proposed a methodology combining the establishment of continuous user-based authentication with federated and decentralized data governance.

As in the previous scenarios, DLT is recurrently used to ensure communications in decentralized scenarios. In this sense, Lu \textit{et al.} \cite{Lu:blockchain_digitaltwin:2021} conducted a theoretical analysis to enhance communication security and protect data privacy in Digital Twin Edge Networks (DITENs). The proposed scheme demonstrated significant improvements in communication efficiency and data security for IoT applications, as confirmed by numerical results. Wilhelmi, Guerra, and Dini \cite{Wilhelmi:dfl_bl:2022} studied the impact of ledger inconsistencies on DFL performance. The study employed a reliable simulation tool that captures the decentralized and asynchronous nature of Blockchain operation. Qi \textit{et al.} \cite{Qi:dfl_bl:2022} proposed a hybrid Blockchain-based incentive mechanism that addresses similar challenges. The authors leveraged smart contracts and dynamic reputation score calculation of each DFL participant. In contrast to previous studies, performance is evaluated against malicious or non-honest nodes. In edge computing settings where P2P communication is commonly used for exchanging model parameters between nodes, an efficient algorithm named CoCo was introduced in \cite{Wang:dfl_edge:2022}. This algorithm integrates topology construction optimization and model compression to accelerate DFL. Extensive simulation results show that CoCo achieves a ten-fold speedup and reduces the communication cost by 50\% on average compared to existing DFL baselines.

The rise of decentralized scenarios has enabled new applications such as recommender systems and social networks. In particular, Belal \textit{et al.} \cite{Belal:recommender_gossip:2022} introduced PEPPER, a decentralized recommender system based on gossip learning principles. The system extracts relevant content for users to aid them in their daily activities, such as finding relevant places to visit, content to consume, or items to buy. In PEPPER, users gossip about model updates and aggregate them asynchronously. By conducting experiments on three real datasets implementing two use cases, location registration recommendation and movie recommendation, the authors demonstrated that their solution converges up to 42\% faster than other decentralized solutions. Furthermore, Khelghatdoust and Mahdavi \cite{Khelghatdoust:socialnetworks_gossip:2022} proposed a Decentralized Online Social Network (DOSN) based on gossip to address the privacy and scalability issues of centralized social networks. The authors observed almost a 30\% reduction in search latency and a 10\% improvement in communication reliability through trusted contacts. These results demonstrate the potential of decentralized approaches in providing efficient solutions for modern applications such as recommender systems and social networks. To improve the previous scenarios, Pan \textit{et al.} \cite{Pan:Lumos_FGL:2023} utilized a tree constructor to improve representation capability given the limited structural information and a Monte Carlo Markov Chain-based algorithm to mitigate workload imbalance caused by degree heterogeneity. Finally, it is worth mentioning that other studies have explored the use of innovative techniques in DFL, such as P2P with knowledge distillation, for various applications such as social network scenarios or IoT device identification \cite{Arapakis:p4l_p2p_telefonica:2023, Jeong:dfl_personalized_distillation:2023, Salama:dfl_mesh_topology:2023}. 

\subsection{Military}

The military scenario has also yielded numerous improvements in DFL communications between different devices on the battlefield. A clear example is the European Future Combat Air System (FCAS) program \cite{Koch:military_FCAS:2021}, the largest European armament effort since World War II, where unmanned aircraft belong to a decentralized and collaborative system during combat missions. \tablename~\ref{tab:dfl_military} highlights the most relevant items in the military scenario where DFL-based approaches are applied.

\begin{table*}[ht!]
\caption{Comparison of DFL solutions related to the military scenario.} \label{tab:dfl_military}
\resizebox{\textwidth}{!}{
\centering
\begin{threeparttable}
\begin{tabular}{ccp{5cm}cccccccc}
\hline
Ref. & 
 Year & 
 \makecell[t]{Goal} & 
 \makecell[t]{Federation\\Architecture} & 
 \makecell[t]{Topology} & 
 \makecell[t]{Scenario} & 
 \makecell[t]{ML\\Model} & 
 \makecell[t]{Aggregation\\Algorithm} & 
 \makecell[t]{Local Model\\Aggregation} &
 \makecell[t]{Client\\Selection} &
 \makecell[t]{Dataset} \\  
\hline \hline

\cite{Mowla:adhoc_uav:2020} & 2020 & Federated learning-based security architecture for jamming attack detection for FANET & \makecell[t]{Cross-device} & \makecell[t]{Centralized\\Decentralized} & \makecell[t]{Simulation} & \makecell[t]{Three-layered NN} & \makecell[t]{FedAvg} & \makecell[t]{Asynchronous} & \makecell[t]{Random} & \makecell[t]{CRAWDAD \cite{Puñal:crawdadjamming:2014}\\NS3 FANET \cite{Riley:ns3dataset:2010}} \\ \hline

\cite{Sharma:bl_military:2020} & 2020 & Distributed defense solution for sustainable society using the features of Blockchain technology and federated learning & \makecell[t]{Cross-device} & \makecell[t]{Centralized\\Decentralized} & \makecell[t]{Simulation} & \makecell[t]{CNN} & \makecell[t]{FedAvg} & \makecell[t]{Synchronous} & \makecell[t]{Random} & \makecell[t]{Custom\\dataset} \\ \hline

\cite{Xiao:military_fullydl:2021} & 2021 & DFL solution with a fully connected network between battlefield devices employing a method of random walks and alternate directions during federation & \makecell[t]{Cross-device} & \makecell[t]{Decentralized} & \makecell[t]{Real} & \makecell[t]{ELM} & \makecell[t]{ISPW-ADMM} & \makecell[t]{Asynchronous} & \makecell[t]{Random} & \makecell[t]{Private} \\ \hline

\cite{Qu:military_uav_architecture:2021} & 2021 & Novel architecture that allows FL within UAV networks without a central entity & \makecell[t]{Cross-device} & \makecell[t]{Decentralized} & \makecell[t]{Simulation} & \makecell[t]{CNN} & \makecell[t]{FedAvg} & \makecell[t]{Asynchronous} & \makecell[t]{Random} & \makecell[t]{\textit{N/S}} \\ \hline

\cite{Wang:uav_military:2021} & 2021 & Secure framework for UAV-assisted mobile crowdsensing to promote high-quality model sharing & \makecell[t]{Cross-device} & \makecell[t]{Decentralized} & \makecell[t]{Simulation} & \makecell[t]{Custom\\model} & \makecell[t]{Custom\\FedAvg} & \makecell[t]{Synchronous} & \makecell[t]{Scheduling} & \makecell[t]{\textit{N/S}} \\ \hline

\cite{Shi:over_the_air:2021} & 2021 & Precoding and decoding strategies for D2D communication & \makecell[t]{Cross-device} & \makecell[t]{Decentralized} & \makecell[t]{Real} & \makecell[t]{Custom\\model} & \makecell[t]{FedAvg} & \makecell[t]{Synchronous} & \makecell[t]{Scheduling} & \makecell[t]{BRATS \cite{Menze:brats:2015}} \\ \hline

\cite{Zhu:military_blockchain:2022} & 2022 & Novel integration of Blockchain and FL in UAV edge computing networks and associated challenges and solutions & \makecell[t]{Cross-device} & \makecell[t]{Decentralized} & \makecell[t]{Simulation} & \makecell[t]{\textit{N/S}} & \makecell[t]{Custom\\FedAVG} & \makecell[t]{Asynchronous} & \makecell[t]{Sequential} & \makecell[t]{\textit{N/S}} \\ \hline

\cite{Feng:dfl_horizontal_bl:2022} & 2022 & Cross-domain authentication of UAVs using multi-signature smart contracts & \makecell[t]{Cross-device} & \makecell[t]{Decentralized} & \makecell[t]{Simulation} & \makecell[t]{CNN} & \makecell[t]{Custom\\algorithm} & \makecell[t]{Synchronous\\Asynchronous} & \makecell[t]{Random} & \makecell[t]{EMNIST \cite{Cohen:emnist:2017}} \\ \hline

\cite{al-abiad:uav_dfl:2023} & 2023 & Resource-efficient framework for mmWave aerial-terrestrial integrated networks using UAVs & \makecell[t]{Cross-device} & \makecell[t]{Decentralized} & \makecell[t]{Simulation} & \makecell[t]{\textit{N/S}} & \makecell[t]{Custom\\algorithm} & \makecell[t]{Asynchronous} & \makecell[t]{Scheduling} & \makecell[t]{MNIST \cite{Deng:MNIST:2012}\\CIFAR-10 \cite{krizhevsky:cifar10-100:2009}} \\ \hline

\cite{Giannopoulos:maritime:2023} & 2023 & Feasibility and adaptability of DFL in maritime transportation & \makecell[t]{Cross-device} & \makecell[t]{Centralized\\Decentralized} & \makecell[t]{Simulation} & \makecell[t]{Fully-connected ANN} & \makecell[t]{Custom\\algorithm} & \makecell[t]{Synchronous} & \makecell[t]{Random} & \makecell[t]{Private} \\ \hline \hline

\end{tabular}
\begin{tablenotes}
\item \textit{N/S} (Not Specified) by the authors
\end{tablenotes}
\end{threeparttable}}
\end{table*}

Flying Ad-hoc Network (FANET) is a type of network that applies DFL as a promising way to train collaborative and decentralized models. One challenge in this network is the evasion of malicious attacks, such as jamming. In jamming attacks, adversaries disrupt the victim network communication by generating interference at the receiver side, making it difficult for the intended signals to be correctly received and processed. For this problem, Mowla \textit{et al.} \cite{Mowla:adhoc_uav:2020} proposed security measures to prevent node jamming by creating a client group prioritization technique leveraging the Dempster-Shafer theory and using asynchronous communications to exchange model parameters. The results provided an accuracy of 82.01\% for the CRAWDAD dataset and 89.73\% for the NS3 FANET dataset. Additionally, Sharma, Park, and Cho \cite{Sharma:bl_military:2020} discussed using DFL in the Internet of Battlefield Things (IoBT) as a defense system employing AI to strengthen the armed forces. The authors rely on the characteristics of Blockchain technology to obtain high accuracy and a random device selection. A custom dataset consisting of several drone detection datasets \cite{Reiser:creiserdronedetection:2022} and images \cite{Fisher:cvonline:2019} is used to evaluate the effectiveness of the proposed model, obtaining an accuracy rate of 99\% at the fog layer. Xiao \textit{et al.} \cite{Xiao:military_fullydl:2021} presented an Inexact Parallel Random Walk Alternate Direction Multiplier Method (ISPW-ADMM) applied to fully connected networks. Then, a UAV communication scenario was deployed and compared with traditional methods such as W-ADMM, PW-ADMM, DGD, and distributed-ADMM (D-ADMM). Qu \textit{et al.} \cite{Qu:military_uav_architecture:2021} proposed a similar solution, where the authors provided three critical considerations: (i) the average loss achieved by DFL is similar to the one obtained in a centralized approach; (ii) for each UAV, the loss value after 60 rounds is similar to the achieved in the centralized approach; and (iii) the training latency is always more negligible in DFL since it does not need to broadcast the model parameters.

Alternative solutions have been proposed to address the issue of secure DFL for UAV-assisted Mobile Crowdsensing (MC). For instance, Wand \textit{et al.} \cite{Wang:uav_military:2021} introduced a DFL approach that utilizes Blockchain to securely exchange local model updates and verify contributions without a central node in a cross-device scenario. In the same way, Shi \textit{et al.} \cite{Shi:over_the_air:2021} proposed a consensus phase based on additive noise at each iteration of the algorithm, which enhanced the robustness of the solution to changes in wireless network topology. The solution converged linearly in a binary classification using MNIST. Furthermore, Zhu \textit{et al.} \cite{Zhu:military_blockchain:2022} used the same technology among multiple untrusted parties with anonymous, immutable, and distributed characteristics, achieving similar results in performance for UAV intelligent edge computing networks to the previous study. Another study by Feng \textit{et al.} \cite{Feng:dfl_horizontal_bl:2022} followed a similar approach, utilizing HFL and non-IID data in a fully connected network. The authors implemented cross-domain UAV authentication through multi-signature smart contracts, with global model updates computed using these smart contracts instead of a centralized server. Recently, decentralized model dissemination has emerged as a promising approach for DFL in mmWave aerial-terrestrial integrated networks. To this end, Al-Abiad \textit{et al.} \cite{al-abiad:uav_dfl:2023} presented an algorithm that makes use of UAVs as local model aggregators through UAV-to-UAV communications and reduces the energy consumption of DFL using Radio Resource Management (RRM) under the constraints of over-the-air learning latency. In maritime environments, optimizing transportation focusing on certain performance metrics may lead to non-convex problems due to the large number and heterogeneity of network nodes and vessels. To tackle this issue, Giannopoulos \textit{et al.} \cite{Giannopoulos:maritime:2023} presented and analyzed various use cases in these scenarios and demonstrated the superiority of DFL over traditional ML approaches using datasets from an enterprise specializing in the maritime industry.

\subsection{Vehicles}

In recent years, concerns regarding road safety have significantly increased, prompting significant efforts towards developing automated solutions to detect distractions while driving and to alert vehicles on the road intelligently \cite{Martinez:SAFECAR:2022}. In this sense, decentralized solutions and FL have emerged as promising approaches in this sector, as illustrated by the latest vehicle solutions highlighted in \tablename~\ref{tab:dfl_vehicles}.

\begin{table*}[ht!]
\caption{Comparison of DFL solutions dealing with vehicular scenarios.} \label{tab:dfl_vehicles}
\resizebox{\textwidth}{!}{
\centering
\begin{threeparttable}
\begin{tabular}{ccp{5cm}cccccccc} 
\hline
Ref. & 
 Year & 
 \makecell[t]{Goal} & 
 \makecell[t]{Federation\\Architecture} & 
 \makecell[t]{Topology} & 
 \makecell[t]{Scenario} & 
 \makecell[t]{ML\\Model} & 
 \makecell[t]{Aggregation\\Algorithm} & 
 \makecell[t]{Local Model\\Aggregation} &
 \makecell[t]{Client\\Selection} &
 \makecell[t]{Dataset} \\  
\hline \hline

\cite{Yu:p2p_vehicles:2020} & 2020 & Proactive caching scheme where vehicles are trained from sparse data, mitigating privacy risks & \makecell[t]{Cross-device} & \makecell[t]{Decentralized} & \makecell[t]{Real} & \makecell[t]{CF-VAE} & \makecell[t]{FedAvg} & \makecell[t]{Asynchronous} & \makecell[t]{Random} & \makecell[t]{MovieLens \cite{Harper:movielens:2015}} \\ \hline

\cite{Lu:vehicles_bl:2020} & 2020 & Asynchronous FL-based approach for efficient and secure data sharing & \makecell[t]{Cross-device} & \makecell[t]{Decentralized} & \makecell[t]{Simulation} & \makecell[t]{Custom NN} & \makecell[t]{FedAvg} & \makecell[t]{Asynchronous} & \makecell[t]{Scheduling} & \makecell[t]{\textit{N/S}} \\ \hline

\cite{Pokhrel:vehicles_bl:2020} & 2020 & Autonomous Blockchain-based powered learning design for a vehicular communication network & \makecell[t]{Cross-device} & \makecell[t]{Decentralized} & \makecell[t]{Simulation} & \makecell[t]{\textit{N/S}} & \makecell[t]{oVML} & \makecell[t]{Synchronous\\Asynchronous} & \makecell[t]{Random} & \makecell[t]{\textit{N/S}} \\ \hline

\cite{Chen:bdfl_autonomous_vehicle:2021} & 2021 & Novel method proposed by DFL with P2P communication between autonomous vehicles & \makecell[t]{Cross-device} & \makecell[t]{Centralized\\Decentralized} & \makecell[t]{Simulation} & \makecell[t]{CNN} & \makecell[t]{Custom\\algorithm} & \makecell[t]{Synchronous} & \makecell[t]{Schematic} & \makecell[t]{MNIST \cite{Deng:MNIST:2012}\\KITTI \cite{Geiger:kittidataset:2012}} \\ \hline

\cite{Barbieri:vehicles:2022} & 2022 & DFL system to increase road user and object classification capability based on Lidar data & \makecell[t]{Cross-device} & \makecell[t]{Centralized\\Decentralized} & \makecell[t]{Simulation} & \makecell[t]{PointNet ML} & \makecell[t]{Custom\\FedAvg} & \makecell[t]{Synchronous} & \makecell[t]{Random} & \makecell[t]{nuScenes \cite{Caesar:nuscenes:2019}} \\ \hline

\cite{Bragato:vehicles_qos:2023} & 2023 & Reinforcement Learning with FL agent for predictive QoS in teleoperated driving scenarios & \makecell[t]{Cross-device} & \makecell[t]{Decentralized} & \makecell[t]{Simulation} & \makecell[t]{DNN} & \makecell[t]{Custom\\algorithm} & \makecell[t]{Asynchronous} & \makecell[t]{Random} & \makecell[t]{KITTI \cite{Geiger:kittidataset:2012}} \\ \hline 

\cite{Hu:dfl_bl_vehicles:2023} &
2023 & Blockchain-enhanced DFL for efficient and privacy-protective data sharing & \makecell[t]{Cross-device} & \makecell[t]{Decentralized} & \makecell[t]{Simulation} & \makecell[t]{CNN} & \makecell[t]{FedAvg} & \makecell[t]{Asynchronous} & \makecell[t]{Scheduling} & \makecell[t]{FMNIST \cite{Xiao:fmnist:2017}} \\ \hline \hline

\end{tabular}
\begin{tablenotes}
\item \textit{N/S} (Not Specified) by the authors
\end{tablenotes}
\end{threeparttable}}
\end{table*}

The unique characteristics of Vehicle-to-Everything (V2X) communication, including the high mobility of vehicles and limited storage capacity of nodes, present particular communication and processing challenges that can be addressed through DFL. Building on recent advances in ML, Yu \textit{et al.} \cite{Yu:p2p_vehicles:2020} proposed a proactive caching scheme based on SDFL and P2P communications. In this solution, a vehicle acts as a parameter server to aggregate the updated global model from peers instead of an edge node. Experimental results show that the solution outperforms typical baselines, gaining efficiency and autonomy. Furthermore, Blockchain supports data integrity when transmitting model parameters, thus addressing privacy concerns when sharing private data between vehicles. This mechanism has been utilized in several studies, such as \cite{Lu:vehicles_bl:2020} and \cite{Pokhrel:vehicles_bl:2020}. While the former used deep Reinforcement Learning (RL) to select the participating nodes of DFL, thereby improving the efficiency of the data-sharing process, the latter minimized the system delay by exploiting the channel dynamics.

Finally, other studies consider new ways to optimize resources to present a solution comparable to server-based architectures. Chen \textit{et al.} \cite{Chen:bdfl_autonomous_vehicle:2021} proposed a novel approach to DFL based on a P2P network designed to be resilient to byzantine faults. The authors evaluated their method on the MNIST and KITTI datasets, assessing the performance of the proposed P2P approach in terms of accuracy, convergence speed, and fault tolerance and comparing it to traditional server-based methods. Similarly, Barbieri \textit{et al.} \cite{Barbieri:vehicles:2022} investigated the use of DFL methods to improve the classification of road users (such as pedestrians, cyclists, and vehicles) and objects based on Lidar data. The authors simulated a realistic V2X network using the collective perception service to share PointNet model parameters. The results showed a low latency training compared to existing distributed ML approaches. Another study that investigated the use of DFL in vehicular networks is presented in \cite{Bragato:vehicles_qos:2023}. In this work, the authors focused on ensuring the Quality of Service (QoS) for communication between vehicles and remote drivers in remote driving scenarios. Specifically, they proposed using Predictive Quality of Service (PQoS) to predict and react to unanticipated degradation of the QoS. To implement PQoS in vehicular networks, the authors designed an RL agent to identify the optimal compression level for sending automotive data under low latency and reliability constraints. Finally, Hu et al. \cite{Hu:dfl_bl_vehicles:2023} integrated Blockchain into a fully connected vehicle topology. The authors demonstrated robust privacy protection and system reliability without significant latency, ensuring efficient and accurate data exchange, even with data of varying quality.
\section{Trends, Lessons Learned, and Open Challenges}
\label{sec:challenges}

Based on the different aspects of DFL analyzed through research questions Q1-Q3, and following the storyline presented in \figurename~\ref{fig:storyline}, this section responds to ``\textit{Q4. What trends, lessons learned, and challenges have emerged in DFL?}'' To provide a clear structure, each trend, lesson, and challenge is related to a specific category, such as \textit{[Fund.]} for fundamentals, \textit{[Fram.]} for frameworks, and \textit{[Sce.]} for application scenarios.

\subsection{Current Trends}

\begin{figure*} 
    \centering
  \subfloat[Fundamentals per application scenario.\label{fig:fundamentals-graph}]{%
        \includegraphics[width=0.48\linewidth]{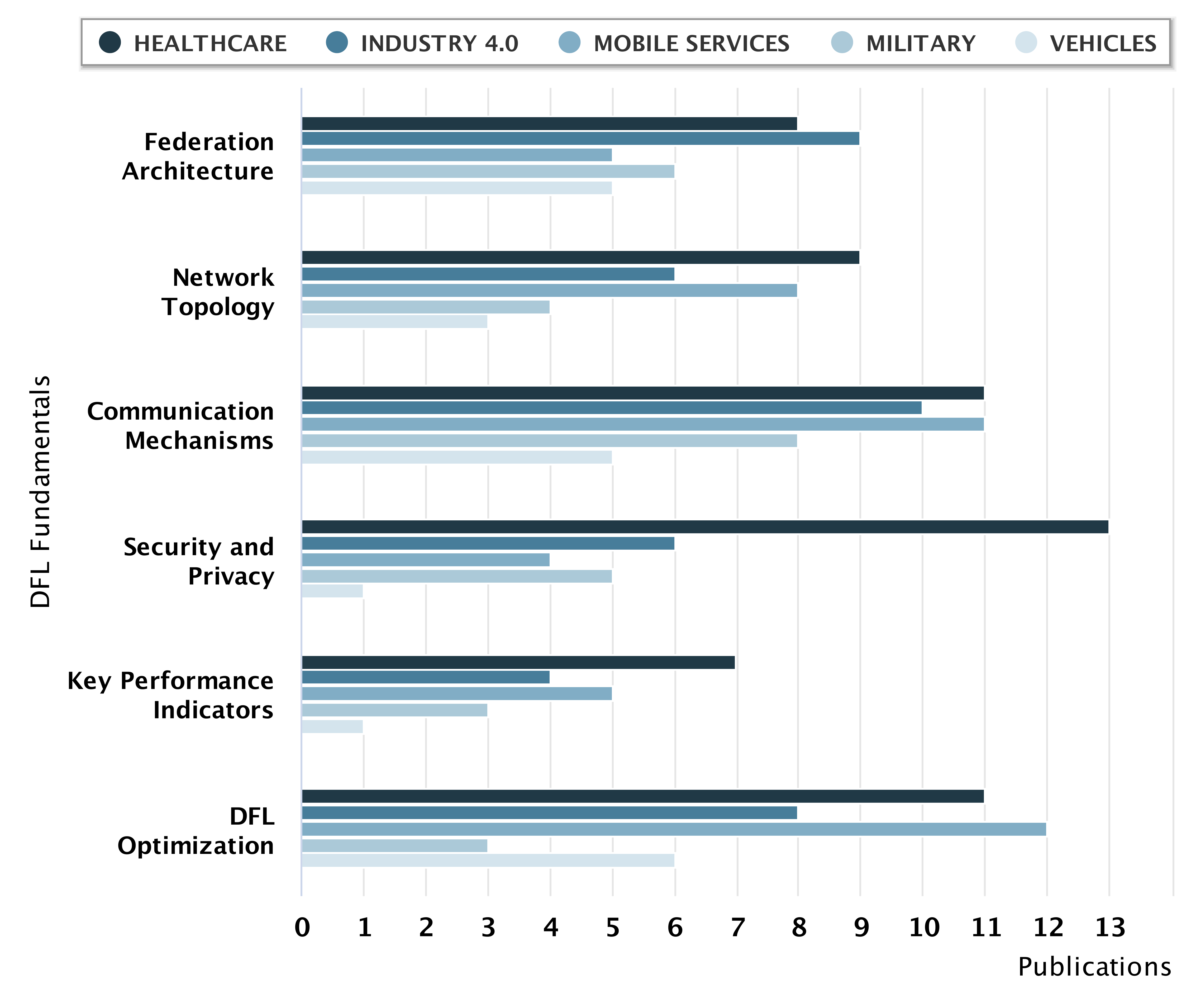}}
    \hfill
  \subfloat[Main objectives per application scenario. \label{fig:scenarios-graph}]{%
       \includegraphics[width=0.48\linewidth]{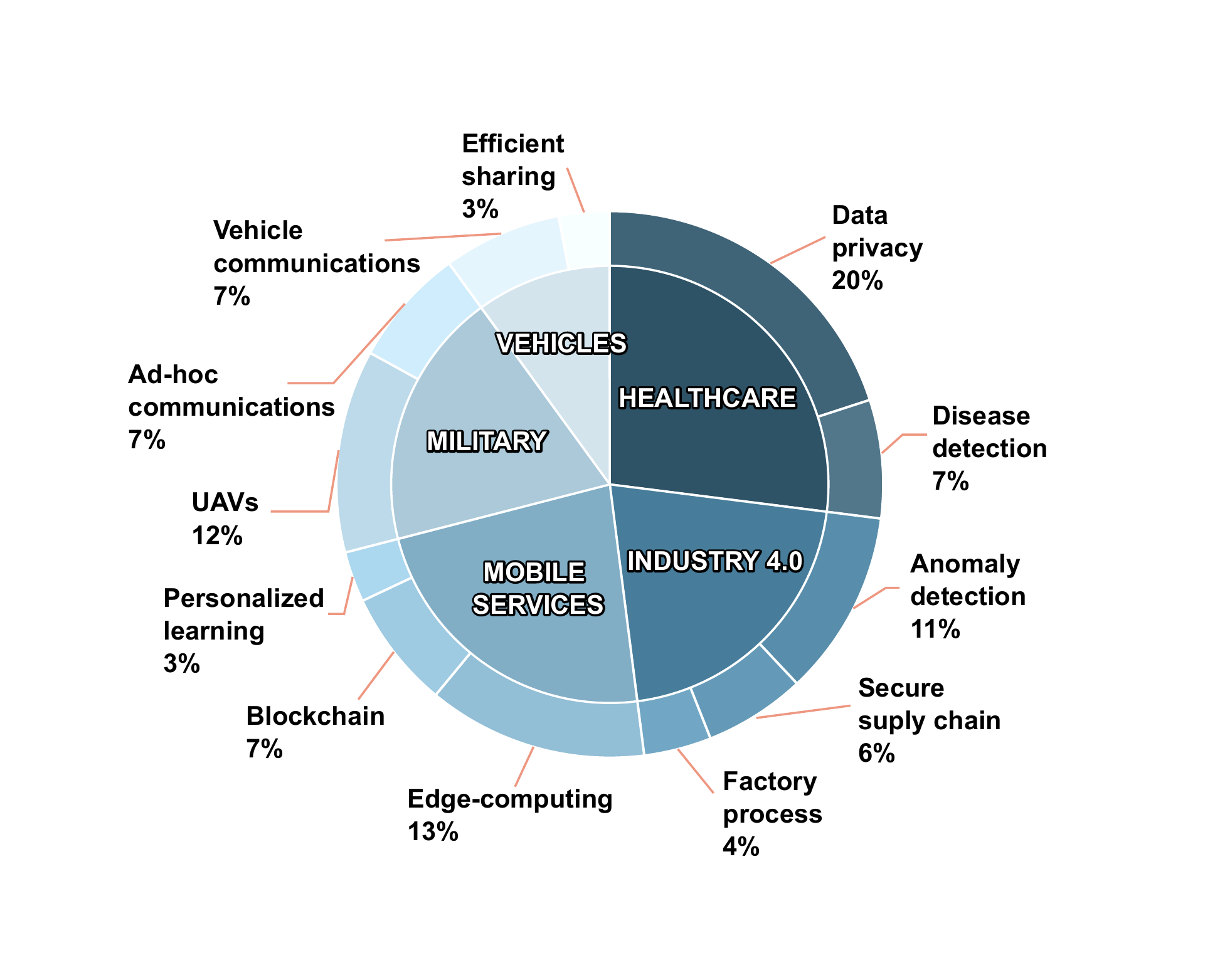}}
  \caption{Distribution of DFL publications.}
  \label{fig:papers_distribution} 
\end{figure*}

The main trends in current work, based on the evolution of recent solutions, are as follows (see \figurename~\ref{fig:papers_distribution}).

\textbf{\textit{[Fund.]} Federation architectures, network topologies, and communication mechanisms are extensively studied}. Most works analyze the communications between network nodes, the network topologies (mostly fully connected), and the architecture it supports, with DFL being the main approach addressed. \figurename~\ref{fig:fundamentals-graph} shows the trend of the above fundamentals in different application scenarios.

\textbf{\textit{[Fund.]} Fully connected network topologies are widely applied to DFL scenarios}. Network topologies where all nodes are connected to each other are the most commonly used in DFL. Its versatility and simplicity of construction have made it a usable approach in most application scenarios. \tablename~\ref{tab:fundamentals} shows how about 50\% of the papers analyzed in the network topology fundamental belong to this approach.

\textbf{\textit{[Fund.]} The optimization of communications is predominant in recent work on DFL}. The inherent functionality of DFL requires communication optimization techniques between federation participants. In this sense, \tablename~\ref{tab:fundamentals} shows how more than 65\% of the analyzed solutions address complexity reduction in model parameter exchanges. The healthcare and mobile services are the most optimized application scenarios in terms of communications (see \figurename~\ref{fig:fundamentals-graph}).

\textbf{\textit{[Fram.]} Most of the frameworks are adapted to cross-device environments}. The reviewed frameworks offer DFL for deploying simulated or real federated scenarios with limited devices (see \tablename~\ref{tab:frameworks}). They also typically provide models fed by images or timelines, aggregation algorithms, and various communication technologies between participants.

\textbf{\textit{[Sce.]} DFL is widely used in healthcare scenarios}. DFL meets its goals in healthcare data analytics, with a significant impact on the accuracy of medicine and, ultimately, improved treatment and diagnosis (see \figurename~\ref{fig:scenarios-graph}). Electronic medical records, medical imaging, disease detection, and collaborative drug discovery are the most significant use cases for DFL.

\textbf{\textit{[Sce.]} DFL gains momentum in mobile services and Industry 4.0}. The high performance offered by DFL has led to the increasing relevance of both application scenarios. In mobile services, DFL improves personalization and recommendation systems in edge devices while protecting user privacy. In Industry 4.0, DFL enables collaboration between multiple organizations without sharing sensitive data, reducing costs, and improving efficiency.

\textbf{\textit{[Sce.]} Anomaly detection and deployment on UAVs are prominent trends}. \figurename~\ref{fig:scenarios-graph} shows how both topics maintain remarkable applicability in different application scenarios. Specifically, anomaly detection represents the 11\% of works documented in industrial scenarios, while the use of UAVs represents the 12\% of works in military scenarios.

\subsection{Lessons Learned}

After reviewing and analyzing the state of the art, the following lessons have been learned:

\textbf{\textit{[Fund.]} The use of specific aggregation algorithms for DFL is still limited}. Although FedAvg is widely used in existing frameworks, applications in certain DFL scenarios need to obtain similar results. Studies of different application scenarios (see \tablename~\ref{tab:dfl_healthcare}, \ref{tab:dfl_industry}, \ref{tab:dfl_mobile}, \ref{tab:dfl_military}, and \ref{tab:dfl_vehicles}) have opted for customizations of the algorithm to adapt it to the peculiarities of the federation models.

\textbf{\textit{[Fund.]} Limited analysis of improving decentralized systems with DFL}. Decentralized systems face challenges such as consistency and coordination between nodes. In the literature, there are not enough studies comparing and evaluating DFL enhancement in these systems. There is a need to determine the resilience, robustness, and overall security provided by this approach that reduces dependency on a server.

\textbf{\textit{[Fram.]} There is a limited number of solutions providing realistic federation benchmarks}. As more DFL architectures are developed, it is important to have a benchmark with representative datasets and workloads to evaluate existing systems and direct future development. Although many DFL frameworks provide benchmarks (see \tablename~\ref{tab:frameworks}), no single benchmark has been widely used in all studies reviewed. In addition, existing benchmarks often ignore metrics such as system efficiency, reliability, or architecture robustness. The evaluation of model performance on non-IID datasets and system security needs further research.

\textbf{\textit{[Fram.]} There is no consensus on frameworks in the literature for deploying DFL architectures}. The frameworks analyzed in \tablename~\ref{tab:frameworks} show heterogeneous characteristics, mostly adapted to validation scenarios. Therefore, there are limited open-source DFL frameworks with sufficient maturity to be network, node, and data agnostic.

\textbf{\textit{[Sce.]} The military and vehicular scenarios are complex application scenarios for deploying DFL solutions}. The solutions presented in the military (see \tablename~\ref{tab:dfl_military}) and vehicles scenarios (see \tablename~\ref{tab:dfl_vehicles}) lack the robustness to be put into practice, so they usually recreate simulated deployments to adapt nodes and models. In addition, they present limited bandwidth, unstable network connections, and high-security requirements in simulations. Therefore, strategies such as edge computing, hybrid approaches, and Blockchain technology are used to reduce complexity.

\textbf{\textit{[Sce.]} There is a lack of literature using unsupervised learning in DFL architectures}. The literature has not addressed the applicability of unsupervised learning in DFL-based scenarios where federation participants do not operate locally labeled data. \tablename~\ref{tab:dfl_industry}, \ref{tab:dfl_mobile}, \ref{tab:dfl_military}, and \ref{tab:dfl_vehicles} indicate that the literature only defines solutions using supervised ML models such as MLP or CNN for classification tasks.

\subsection{Open Challenges}

Based on the current state of the art, the following points represent the main challenges that future DFL solutions might consider. In addition, \tablename~\ref{tab:challenges} summarizes the open challenges and their future developments, indicating the importance of their application in the future.

\newcommand{\noimportant}{{\color{ao}$\mathord{!}$}}
\newcommand{\important}{{\textbf{\color{amber}$\mathord{!!}$}}}
\newcommand{\critical}{{\color{red}$\mathord{!!!}$}}

\begin{table}[ht!]
\caption{Open challenges in DFL and future developments.} \label{tab:challenges}
\resizebox{\columnwidth}{!}{
\centering
\begin{threeparttable}
\begin{tabular}{p{4.5cm}p{5.5cm}} 
\hline
\makecell{Challenge} & \makecell{Future Developments}
\\ \hline \hline
\multicolumn{2}{c}{Fundamentals \textit{[Fund.]}}
\\ \hline

Scalability of DFL with increasing participants (\critical) &
$\sbullet[0.75]$ Dynamic participant selection\newline
$\sbullet[0.75]$ Personalized local model learning
\\ \hline 

Cybersecurity mechanisms for a secure DFL (\critical) &
$\sbullet[0.75]$ Detect attacks in DFL scenarios\newline
$\sbullet[0.75]$ Different treatment based on privacy
\\ \hline

Trustworthiness among federation participants (\critical) &
$\sbullet[0.75]$ Maintain trust policies\newline
$\sbullet[0.75]$ Prevent dishonest behavior
\\ \hline

Homogeneous node participation (\important) &
$\sbullet[0.75]$ Quantization and gradient compression\newline
$\sbullet[0.75]$ Use of SDFL
\\ \hline

Address participant mobility in DFL scenarios (\important) &
$\sbullet[0.75]$ Topology-aware node reconfiguration\newline
$\sbullet[0.75]$ Resilient synchronization methods
\\ \hline

Study of adversarial attacks (\important) &
$\sbullet[0.75]$ Identify the techniques and their impacts\newline
$\sbullet[0.75]$ Compare against traditional approaches
\\ \hline

Explore the use of Reinforcement Learning (\important) &
$\sbullet[0.75]$ Optimize the federated model performance\newline
$\sbullet[0.75]$ Improve the selection of participants
\\ \hline

DFL standardization efforts (\important) &
$\sbullet[0.75]$ Promote comprehensive DFL standards\newline
$\sbullet[0.75]$ Involve standard-setting bodies (ISO, IEEE)
\\ \hline

5G and 6G technologies for communications (\noimportant) &
$\sbullet[0.75]$ Network slicing utilization\newline
$\sbullet[0.75]$ 5G/6G-integrated edge computing
\\ \hline

\multicolumn{2}{c}{Frameworks \textit{[Fram.]}}
\\ \hline

Modular, scalable, and efficient frameworks (\critical) &
$\sbullet[0.75]$ Implement and manage DFL fundamentals\newline
$\sbullet[0.75]$ Application in practical scenarios
\\ \hline

Heterogeneous datasets in decentralized participants (\important) &
$\sbullet[0.75]$ Data preprocessing and normalization\newline
$\sbullet[0.75]$ Advanced data augmentation techniques
\\ \hline

Dynamic scheduling of federated network (\noimportant) &
$\sbullet[0.75]$ Adaptable federation architecture\newline
$\sbullet[0.75]$ Resilient algorithms
\\ \hline

\multicolumn{2}{c}{Application scenarios \textit{[Sce.]}}
\\ \hline

Exploration of new DFL application scenarios (\important) &
$\sbullet[0.75]$ Evaluate DFL in smart city technologies\newline
$\sbullet[0.75]$ Combine AI, IoT, and DFL for testing tools
\\ \hline \hline

\end{tabular}
\begin{tablenotes}
\item \noimportant\space low importance,\space\important\space high importance,\space\critical\space critical
\end{tablenotes}
\end{threeparttable}}
\end{table}

\textbf{\textit{[Fund.]} Improve the scalability of the solution when the number of participants in the federation increases}. The development of algorithms that can dynamically select participants based on their availability, network connection, and trustworthiness would help to ensure that the federation remains resilient to dropouts and the unavailability of participants. In addition, exploring new techniques for compressing and aggregating models across multiple participants can reduce communication and computation overhead, particularly for large-scale federations. Finally, including personalized local model learning can improve training while minimizing the impact of low network dependability and client availability.

\textbf{\textit{[Fund.]} Improve the cybersecurity mechanisms depending on the participant and the application scenario}. The security of the nodes participating in DFL and their ability to detect and prevent attacks or threats in heterogeneous scenarios are crucial to the success of this approach. A promising solution would be to design an architecture that treats participants differently according to their privacy restrictions, allowing for a more personalized and secure approach. This is particularly relevant for application scenarios such as social media applications, industrial IoT, and Blockchain, where user privacy protection is critical.

\textbf{\textit{[Fund.]} Enhance trustworthiness among federation participants in DFL approaches}. To ensure the accuracy and reliability of the federated model, participating nodes must establish trust policies with one another. These policies allow participants to aggregate model parameters based on reputation and performance history only from those they trust. By doing so, participants can avoid dishonest behavior and maintain the integrity of the FL process, thereby encouraging the participation of other nodes in the network.

\textbf{\textit{[Fund.]} Ensure the homogeneous participation of the constrained nodes in the federation}. It is challenging to guarantee the participation of nodes with limited autonomy or bandwidth. Some solutions could use quantization methods and gradient compression techniques to reduce communication overhead. Another option is to consider other federation architectures, such as SDFL, where nodes with limited resources can participate in the federation while still maintaining some level of centralization.

\textbf{\textit{[Fund.]} Address participant mobility}. The dynamic nature of DFL, where participants may join or leave the federation at any time, poses challenges in maintaining learning stability and reliability. Future DFL solutions need to manage mobility and develop strategies to handle the continuous flow of participants while ensuring efficient learning.

\textbf{\textit{[Fund.]} In-depth study of adversarial attacks applied in DFL approaches}. The literature on cyberattacks affecting DFL is limited, with relatively little attention given to adversarial attacks and their impact on DFL performance. Thus, evaluating their impact compared to other traditional approaches, such as CFL, can help understand the vulnerabilities of DFL better and inform the development of more robust security measures.

\textbf{\textit{[Fund.]} Explore the use of RL}. The literature regarding RL applied in decentralized collaborative scenarios is scarce. This learning approach can optimize the performance of the federated model by dynamically allocating resources and adjusting the learning process. Additionally, it can be used to optimize the selection of participants in the federation, as well as to dynamically adjust the communication and aggregation parameters based on the performance of the participants.

\textbf{\textit{[Fund.]} Research the use of 5G/6G networks to improve DFL communications}. The rise of 5G and subsequent 6G networks presents a significant potential to enhance communications performance by providing faster and more reliable communication between nodes. Thus, exploring techniques for efficient data aggregation and transmission is necessary using ultra-low latency and high bandwidth capabilities. Moreover, researchers could examine the possibility of utilizing network slicing to allocate dedicated resources for DFL tasks and integrate edge computing within these networks. Furthermore, with 5G/6G-powered DFL, a deep focus on the security and integrity of P2P communications between edge devices becomes critical. Given the potential vulnerability of these networks to novel security threats, dedicated research on defense mechanisms is essential to safeguard secure and trusted P2P exchanges. To this end, it is necessary to deploy appropriate defense mechanisms, such as encryption, SMPC, or Blockchain technologies.

\textbf{\textit{[Fram.]} Create modular, scalable, and efficient frameworks for diverse application scenarios}. The literature has not addressed an agnostic federation for different solutions in terms of DFL fundamentals. Therefore, it is necessary to design and implement a robust and reliable solution to generate realistic scenarios, considering elements such as model creation, data storage, framework usage, model parameter exchange, and model aggregation. The framework should focus on efficiency, considering communication overhead, computational resources, and storage requirements.

\textbf{\textit{[Fram.]} Handle heterogeneous datasets in decentralized participants}. Participants may possess varying sizes, distributions, and quality data, impacting the model performance. Addressing the challenge of dividing or adapting datasets to account for these differences is crucial to achieving optimal performance in DFL scenarios. It involves creating methods for data preprocessing, normalization, and augmentation that can accommodate the distinct characteristics and variations inherent in the datasets of each participating node.

\textbf{\textit{[Fram.]} Adapt the dynamic scheduling of the federated network to the application scenario}. Due to the instability of the application scenario, the number of participants may not be fixed during the learning process in DFL. Therefore, it is necessary to develop algorithms that can dynamically adjust the number of participants in the federation based on changes in inputs or outputs from the nodes. This would require building resilient algorithms that handle low network dependability and client availability. In addition, exploring the use of adaptive FL techniques that can adjust the level of centralization or decentralization in response to changes in the application scenario could be beneficial.

\textbf{\textit{[Sce.]} Explore new DFL application scenarios}. DFL can be effectively applied in a multitude of application scenarios. For instance, DFL can help to increase the intelligence of robotic and autonomous systems, making them more efficient and reliable. In smart city technologies, DFL can exchange local model parameters between devices, enhancing overall performance. Additionally, DFL can be evaluated using Digital Twin (DT) technology, combining AI, IoT, and DFL to create a simulated environment where tools can be tested in real time without exposing people or facilities to unknown risks.

\textbf{\textit{[Fund.]} Explore standardization activities for DFL}. It is necessary to have comprehensive standards to ensure uniformity, interoperability, and trustworthiness across various implementations. These standards should address key elements like privacy-preserving methods, security features, data quality standards, model training processes, and performance indicators. Standardizing DFL could encourage consistency in research outcomes and establish a universal framework for assessing DFL methodologies.
\section{Conclusion and Future Work}
\label{sec:conclusion}

This work studies the evolution of DFL in recent years, providing the basic and distinctive fundamentals versus traditional federated architectures, its current application scenarios, and the frameworks that manage the deployment of DFL architectures. In this context, the present work has answered the following research questions.

\textit{Q1. What are the fundamental aspects of DFL?} Section~\ref{sec:fundamentals} studies the main aspects of DFL in the most current and representative works in the literature, clarifying the differences with CFL. The analysis provides a set of fundamentals related to federation architectures, network topologies, communication mechanisms, and security techniques. It also proposes KPIs to evaluate the approach and mechanisms to optimize DFL. With all this, the work proposes the first taxonomy that defines and details the fundamentals of DFL.

\textit{Q2. What DFL frameworks exist, and what fundamentals do they provide?} Section~\ref{sec:frameworks} discusses the main frameworks currently used to deploy DFL scenarios. This section shows the most mature frameworks in FL versus the incipient solutions that favor the inclusion of DFL. Thus, this paper highlights the prominence of current solutions and the DFL fundamentals contemplated to generate robust solutions to the scenario.

\textit{Q3. Which are the main characteristics of the most relevant scenarios of DFL?} Section~\ref{sec:scenarios} describes, analyzes, and compares the most relevant and recent solutions according to their application scenario. The predominant application scenarios are healthcare, mobile services, and Industry 4.0. Also, this work determines that fields such as military or vehicles have grown significantly in recent years. Regarding fundamentals, decentralized cross-device architectures in fully connected network topologies are predominant. Optimization mechanisms are applied to the above aspects, mainly to communications and aggregation algorithms.

\textit{Q4. What trends, lessons learned, and challenges have emerged in DFL?} Lessons learned, current trends, and future challenges have been documented in Section~\ref{sec:challenges}. It details how to create more sophisticated federation architectures, topologies, optimizations, and their applicability in different scenarios. At the same time, certain limitations in the recent literature are also detailed. For instance, the scarce comparison between DFL scenarios or the lack of frameworks allowing their deployment. In addition, handling heterogeneous datasets, cyberattacks, or using 5G/6G in communications have not yet been robustly applied, opening avenues for future research.

As future work, it is planned to design and implement scalable solutions capable of generating heterogeneous scenarios. To achieve this, it is necessary to create solutions using agnostic data types employing performance data from the devices acting as nodes. These solutions will be integrated with ML/DL techniques, communication optimization, and data aggregation techniques that preserve security and privacy while guaranteeing network and node performance capabilities. Furthermore, these solutions must offer a range of configurable options based on various federation architectures, including CFL, DFL, or SDFL, as well as decentralized technologies such as Blockchain. Finally, it is considered to define and build sets of metrics to complement the functionality of the solution and the evaluation of the system.


\bibliography{references}
\bibliographystyle{IEEEtran}




\vfill

\end{document}